\documentclass[times]{elsarticle}
\usepackage[top=2.cm,bottom=2.cm,left=2.5cm,right=2.5cm]{geometry}
\usepackage[utf8]{inputenc}
\usepackage{lineno,hyperref}
\usepackage{lscape}
\usepackage{multirow}
\usepackage{amsmath}
\modulolinenumbers[5]
\graphicspath{{figs/}}

\begin{document}

\begin{frontmatter}

\title{Are metaheuristics worth it? A computational comparison between nature-inspired and deterministic techniques on black-box optimization problems}

 \author{Jakub Kudela}
 \address{Institute of Automation and Computer Science, Brno University of Technology, Czech Republic\\ Jakub.Kudela@vutbr.cz}
%\author{Anonymous authors}
%\address{Hidden for the purpose \\of the review}

\journal{...}


\begin{abstract}
In the field of derivative-free optimization, both of its main branches, the deterministic and nature-inspired techniques, experienced in recent years substantial advancement. In this paper, we provide an extensive computational comparison of selected methods from each of these branches. The chosen representatives were either standard and well-utilized methods, or the best-performing methods from recent numerical comparisons. The computational comparison was performed on five different benchmark sets and the results were analyzed in terms of performance, time complexity, and convergence properties of the selected methods. The results showed that, when dealing with situations where the objective function evaluations are relatively cheap, the nature-inspired methods have a significantly better performance than their deterministic counterparts. However, in situations when the function evaluations are costly or otherwise prohibited, the deterministic methods might provide more consistent and overall better results.
\end{abstract}


\end{frontmatter}

%\linenumbers

\section{Introduction}
Derivative-free (or black-box) optimization is an area of long history, which experiences current rapid growth, fueled by a growing number of applications in science, engineering, and medical fields \cite{rios2013derivative}. As the analytic properties of the methods used in derivative-free optimization are difficult to study, benchmarking of the methods has an essential role in both the comparison of the different approaches and the development of new methods \cite{hellwig2019benchmarking}. Another use of benchmarking is in the qualification of the theoretical predictions of the behaviour of various methods \cite{rardin2001experimental}. The two main branches of derivative-free methods are deterministic and stochastic techniques.

Among the deterministic derivative-free global optimization techniques, the Lipschitz optimization methods \cite{pinter2013global} are one of the most classical ones. The most well-known and widely utilized extension to the Lipschitz optimization methods was the DIRECT (DIviding RECTangles) algorithm \cite{jones1993lipschitzian}. DIRECT has been found to be effective on low-dimensional, real-world problems from a wide range of areas such as surgery \cite{xiao2019evaluation}, genetics \cite{ljungberg2004simultaneous}, ship design \cite{campana2016derivative}, photovoltaic systems \cite{nguyen2010global}, and many more. However, the main drawback of DIRECT seemed to be its lack of exploitation capabilities \cite{sergeyev2006global}. While the original DIRECT was designed only for box-constrained optimization problems, various DIRECT-type extensions and modifications have been proposed \cite{jones2021direct}. It was also found that the performance of the DIRECT-type algorithms is one of the best among all tested state-of-the-art derivative-free global optimization approaches \cite{rios2013derivative}, outperforming even some of the standard nature-inspired methods. 

Arguably, the most currently active area of research in the stochastic derivative-free techniques (also called metaheuristics) are the nature-inspired (or bio-inspired) algorithms. These methods simulate biological processes such as natural selection, or evolution, where solutions are represented by individuals that reproduce and mutate to generate new, potentially improved candidate solutions for the given problem \cite{molina2020comprehensive}. Other nature-inspired methods try to mimick a collective behavior of simple agents, giving rise to the concept of swarm intelligence \cite{yang2013swarm}. These methods were successfully used in the optimization of various complex problems such as the hyperparameter optimization in deep learning \cite{young2015optimizing} or the design of quantum operators \cite{vzufan2021advances}. Although there has recently been a huge increase in the number of various bio-inspired methods, many of them were shown to be of questionable quality \cite{molina2020comprehensive, bujok2019comparison, campelo2021sharks, kudela2022commentary, kudelanature}. This does not mean that the progress in the development of the nature-inspired methods has stalled. Currently, the best place for finding the most competitive nature-inspired methods are the IEEE Congress on Evolutionary Computation (CEC) competitions \cite{cec22}, that have been running in the current form since 2013. 

Benchmarking different derivate-free techniques on black-box optimization problems has a rich history. In \cite{ali2005numerical}, a comparison of five stochastic algorithms (variants of simulated annealing, genetic algorithms, and differential evolution) were compared on a testbed of 56 functions (containing most of the now ubiquitous test functions such as Ackley's, Branin, Hartman, Levy, McCormick, or Rosenbrock) with dimensions $D$ ranging between 2 and 20, and maximum number of function evaluations $100{\cdot}D^2$ (i.e., 40{,}000 for the highest dimension). Deterministic techniques (pattern-search and model-based methods) were compared in \cite{more2009benchmarking}, where modified problems from the CUTEr \cite{gould2003cuter} collection were used. This benchmark set had 53 problems in dimensions between 2 and 12. The authors focused on the short-term behavior of the studied algorithms and set the maximum number of function evaluations to 1{,}300. A large computational study of 22 different implementations of both stochastic and deterministic techniques was performed in \cite{rios2013derivative}. The testbed comprised of 502 problems mainly from the globallib \cite{globallib} and princetonlib libraries, and a collection of nonsmooth problems \cite{lukvsan2000test}, in dimensions ranging from 1 to 300. However, most of the test problems were smooth, almost a half of them were convex, and their experimental setup used for the tests only allowed for 2{,}500 function evaluations, which heavily favoured the deterministic techniques. The most contemporary comparison of deterministic and stochastic techniques was performed in \cite{sergeyev2018efficiency}. In this study, 800 test problems were used to compare three standard stochastic algorithms (genetic algorithm, artificial bee colony, and firefly algorithm) and three deterministic algorithms (DIRECT, DIRECT-L, and diagonal algorithm) with a budget of $10^6$ function evaluations. Although the scale of the comparison was impressive, the authors used only a single source for the construction of the benchmark set (the GKLS generator \cite{gaviano2003algorithm}), the dimensions were only between 2 and 5, and the selected techniques can hardly be though of as the ``state-of-the-art'' in either of the two categories.

In this paper, we provide an extensive computational comparison between selected DIRECT-type and nature-inspired methods. In both of these classes, we chose the representatives to be either standard and well-utilized methods, or the best performing methods from recent numerical comparisons. These representatives were evaluated on five different benchmark sets and the results were analyzed in terms of performance, time complexity, and convergence properties of the selected methods. 

The rest of the paper is structured as follows. Section 2 provides the details on the algorithms selected for the computational comparison. In section 3 are given the details of the chosen benchmark sets and the experimental setup. Section 4 summarizes the results of the computations and provides discussion on the implications. Finally, conclusions are drawn in Section 5.

\section{Algorithms Selected for Computational Comparison}
All of the considered algorithms are designed to minimize a ``black-box'' objective function with lower and upper bounds on the variables, i.e.
\begin{align} \label{eq1}
\begin{split}
     \underset{x_1,\dots,x_n}{\text{minimize}} \,\,\, & f(x_1,\dots,x_n) \\
        \text{subject to} \,\,\, & l_k \leq x_k \leq u_k, \quad k=1,\dots,n.
        \end{split}
\end{align}
In this paper, only problems defined in a real-value, continuous search area are considered. For problems with a discrete search space, the use of other optimization algorithms or some modifications to the presented methods are needed. For both the nature-inspired and deterministic methods, we selected the representative algorithms as either being considered standard (and extensively utilized over a long time period) or being shown to be competitive in recent computational comparisons or competitions \cite{latorre2021prescription}.

\subsection{Nature-Inspired Metaheuristics}
There is an enormous variety of nature-inspired algorithms that are used in optimization \cite{campelo2021sharks}. However, as was shown in recent analytical \cite{camacho2019intelligent, camacho2020grey, villalon2021cuckoo} and numerical  \cite{bujok2019comparison,del2021more} comparisons, a large portion of the more recently proposed methods have various issues concerning their novelty and performance. For this reason, we chose for the computational comparison two standard and extensively utilized methods, as well as three well-performing methods from the recent CEC competitions on bound-constrained numerical optimization. 

The first chosen nature-inspired algorithm was the particle swarm optimization (PSO) \cite{kennedy1995particle}, which is still one of the most popular and utilized swarm-based methods with a wide range of applications such as cloud computing \cite{pradhan2020novel, pradhan2021survey}, path planning \cite{song2021improved, kanagaraj2022meta}, airfoil design \cite{muller2022improving}, or Internet of Things \cite{sennan2021t2fl}. As was found in \cite{bujok2019comparison}, other swarm-based methods, such as the Firefly algorithm  \cite{lukasik2009firefly} or the Artificial Bee Colony algorithm \cite{karaboga2007powerful} do not significantly outperform PSO, and as such, were not included in the comparison. In the PSO, with a population of $POP$ individuals, at  iteration $i$, each individual is accelerated by velocity $v$ toward both the best position it has seen ($x^{(i)}_{\text{best}}$) and the best position found thus far by any individual ($x_{\text{best}}$). This acceleration is weighted by a random term, with separate random numbers being generated for each acceleration. The update equations are:
    \begin{align*}
        x^{(i)} &= x^{(i)} + v^{(i)}, \\
        v^{(i)} &= wv^{(i)} + c_1r_1 (x^{(i)}_{\text{best}}- x^{(i)})+ c_2 r_2 (x_{\text{best}} - x^{(i)}),
    \end{align*}
where $w, c_1$, and $c_2$ are the parameters, and $r_1$ and $r_2$ are random numbers drawn from a uniform distribution $\mathcal{U}(0, 1)$. The slightly updated by the variation of PSO was used in the computational comparison \cite{shi1998modified}, in which the control parameter of variation $w$ is set for each generation as a linear interpolation from maximal value $w_{\max} = 0.9$ to $w_{\min} = 0.4$. The parameters controlling a local and a global part of the updated velocity were set to $c_1 = 2, c_2 = 2$.

The second nature-inspired algorithm was the differential evolution (DE) \cite{storn1997differential}, which is also still widely used in areas such as structural optimization \cite{huynh2021q}, photovoltaics \cite{li2020enhanced}, or in the optimization of parameters of a deep belief network \cite{deng2020improved}. This algorithm also served as a baseline method for many of the best-performing methods in the CEC competitions, such as SHADE \cite{tanabe2013success}, LSHADE \cite{tanabe2014improving}, or jSO \cite{brest2017single}. In the DE, with population $POP$, at each iteration, the algorithm constructs new individuals in the population by recombining other individuals in the population based on a simple procedure \cite{kochenderfer2019algorithms}, parametrized by a differential weight $F$ and crossover probability $CR$. For each individual $x$:
\begin{itemize}
    \item[1.] Choose three random distinct individuals $x^a$, $x^b$, and $x^c$ from the population.
    \item[2.] Construct an interim design $x^z=x^a+F\cdot(x^b-x^c)$
    \item[3.] Choose a random dimension $j \in [1, \dots, n]$.
    \item[4.] Construct the candidate individual $x'$ using binary crossover.\\
    $x'_i = \begin{cases} x^z_i & \text{if } i=j \text{ or with probability }CR  \\ x_i & \text{otherwise} \end{cases}$
    \item[5.] Insert the better design between $x$ and $x'$ into the next generation.
\end{itemize}
In our implementation, the parameters were set to $F=0.5$ and $CR=0.9$.

The third algorithm was the aforementioned Linear Population Size Reduction SHADE (LSHADE) \cite{tanabe2014improving} which is an improved version of the SHADE \cite{tanabe2013success} algorithm. In LSHADE, the population size is decreased linearly in order to set exploration in the early stages and exploitation in the latter stages of the search. The DE control parameters $F$ and $CR$ are adjusted based on the success of the previous generations. It also uses an archive of outperformed old solutions for a current-to-$p$best mutation. LSHADE was the winner of the CEC 2014 competition and was used as a basis for many of the best performing algorithms in the CEC Competitions in past few years. It was also one of the best performing algorithms on the ambiguous benchmark set \cite{kudela2022new}. 

The fourth selected algorithm was the Adaptive Gaining-Sharing Knowledge (AGSK) \cite{Wagdy2020}, which was the the runner-up of the CEC 2020 competition. The algorithm is an enhancement of the original GSK \cite{Wagdy2020b} algorithm with added adaptive settings to two main control parameters: the knowledge factor and the knowledge ratio, that together control junior and senior gaining and sharing phases between the solutions during the process of optimization.

The last nature-inspired algorithm was the Hybrid Sampling Evolution Strategy (HSES) \cite{8477908}, which is a combination of the traditional the covariance matrix adaptation-evolution strategy (CMAES) \cite{hansen2001completely} and the univariate sampling method (USM) \cite{hauschild2011introduction}. It relies on the CMAES to solve unimodal problems and USM for multimodal problems. HSES was the winner of the CEC 2018 competition and was also one of the best performing algorithms on the ambiguous benchmark set \cite{kudela2022new}.

We did not perform any hyperparameter tuning, as we want to compare the methods as if we only get one shot for the optimization. For the two standard methods (PSO and DE), we use the recommended settings. For the three high performing algorithms (LSHADE, AGSK, and HSES), we utilized the same parameter settings that was used in the corresponding CEC competitions \cite{kazikova2020tuning}.

\subsection{DIRECT-type Methods}
The DIRECT-type methods are motivated by Lipschitzian optimization. In this approach, one assumes that the objective function $f$ is Lipschitz continuous, i.e., that there exists a constant $K<\infty$ such that for any two variables $x$ and $x'$, the following equation holds:
\begin{equation} \label{eq2}
  |f(x) - f(x')| \leq K || x-x'||.  
\end{equation}
In (\ref{eq2}), the norm on the r.h.s. can be either the standard Euclidean 2-norm (most DIRECT variant) or other non-Euclidean norms (such as the infinity norm used in DIRECT-L). The DIRECT-type methods avoid the drawbacks of using a single, large Lipschitz constant \cite{jones2021direct} -- they search  the design space in areas that could have the lowest Lipschitzian lower bound for some Lipschitz constant $K > 0$, subject to the constraint that the resulting lower bound be non-trivially better than the best solution found so far.

Because the optimization problem has only lower and upper bounds on the variables, we can normalize the variables to $[0, 1]$ transforming the search space into a unit hypercube. The original DIRECT algorithm partitions this unit hypercube into subrectangles in such a way that the objective function has been evaluated at each rectangle’s center point. The individual iterations then consist of:
\begin{itemize}
    \item Splitting a selected subrectangle into smaller ones:
\begin{itemize}
    \item Sampling the points $c \pm \delta e_k $, $k = 1,\dots , n,$ where $c$ is the center point of the subrectangles, $\delta$ is one-third the side length of the subrectangles, and $e_k$ is the $k$th unit vector.
    \item Computing 
    $$
    w_k = \min\{ f (c + \delta e_k ), f (c -\delta e_k )\}
    $$
    and splitting the subrectangles along the dimension with the smallest $w$ value. Afterward, splitting the subrectangles containing $c$ into thirds along the dimension with the next smallest $w$ value, etc., until the split has been performed on all dimensions.
    \end{itemize}
    \item Identifying the subrectangles $j$ in which there is a potential for the improvement of the objective function, i.e., where there exists some $K>0$ such that 
    \begin{align*}
        &f(c_j) - Kd_j \leq f(c_i) - Kd_i, \quad \text{for all } i = 1,\dots,m, \\
        &f(c_j) - Kd_j \leq f_{\text{min}} - \varepsilon|f_{\text{min}}|,
    \end{align*}
    where $c_i$ is the center points of the $i$th subrectangles, $d_i$ the distance from the center point to its vertices, and $\varepsilon >0$ a small positive constant (accuracy).
\end{itemize}
The DIRECT-type methods subdivide one of the largest hyperrectangles during each iteration, meaning that as number of iterations increases, the size of the largest rectangle approaches zero. This accounts to one of the weaknesses of these methods, as they tend to spend numerous iterations ``refining'' subregions, instead of focusing on finding the global minimum. 

Another weakness of the original DIRECT method is the inflexibility of the partitioning scheme - it samples only points that are centers of hyperrectangles in the partition, meaning that some points (such as the boundary points of the feasible space) cannot be sampled exactly. It also is difficult for DIRECT to accept a predefined starting point \cite{matousek2022start} (even with good objective function values), or use techniques such as Latin Hypercube Sampling. 

The strengths of the DIRECT-type methods lie in its deterministic nature (there is no need to perform multiple runs). They are guaranteed to converge to the global minimum, provided that the objective function is continuous (even though the required number of iteration might be enormous). Also, the original DIRECT has no hyperparameters that set the balance between local and global search (the only parameter is the accuracy $\varepsilon$).

There is a large variety of the DIRECT-type algorithms. For the computational comparison, four variations of the DIRECT algorithm were selected. The selection covers both the older methods and newer methods that performed well in numerical comparisons \cite{stripinis2021dgo}:
\begin{itemize}
    \item The original DIRECT method \cite{jones1993lipschitzian} denoted as DIR.
    \item Locally-biased DIRECT method \cite{gablonsky2001locally} denoted as DIR-L. This variant introduced certain changes that reduce global drag on refining local solutions. 
    \item Hybrid DIRECT method with a separate local optimizer \cite{liuzzi2010direct} denoted as DIRMIN. In this variant, a local search (utilizing a truncated Newton's method) starting from the centerpoint of every selected rectangle is performed. 
    \item Two-step global-local DIRECT method \cite{stripinis2018improved} denoted as DIR-GL. This variant performs a two-step selection process in every iteration. The global step selects rectangles that are nondominated with respect to centerpoint function value versus rectangle size. The local step selects rectangles that are nondominated with respect to distance to the current best point and rectangle size.
\end{itemize}
The particular implementations of the chosen methods can be found in the DIRECTGO toolbox in MATLAB \cite{stripinis2021dgo}.

\section{Benchmark Sets and Experimental Setup}

The selection of the benchmark sets for the computational comparison was based on newly proposed guidelines for comparing nature-inspired optimization algorithms \cite{latorre2021prescription}. The benchmark sets will be introduced in a chronological order (based on the date of the corresponding publication). As the first benchmark set we chose the test suite for the CEC 2017 \cite{wu2017problem}. Originally, this set included 29 benchmark functions but it was found the second function in the set had numerical issues and was removed (leaving 28 benchmark functions). The dimension of the problems used in our study was $D=10$.

The remaining four benchmark sets were all proposed in the past few years. The second set is based on the DIRECTLib \cite{dirlib} library that is mainly used in benchmarking Lipschitz-based deterministic techniques. From this set, we chose 18 box-constrained problems with variable dimension, with $D=[5,10,15,20]$. The third set was the one used in the Black-box Optimization Benchmarking (BBOB) workshop series \cite{hansen2021coco}, from which we chose 24 noiseless functions in dimensions $D=[5,10,20]$. As the fourth benchmark set, we selected the test suite for the CEC 2022 competition \cite{cec22}. This benchmark set contains 12 functions of various computational complexity in dimensions $D = [10, 20]$. As the final benchmark set, we selected the recently proposed ambiguous benchmark set \cite{kudela2022new}. This benchmark set uses 8 functions that are based on a zigzag pattern \cite{9504720} and are non-differentiable and highly multimodal, all of them in dimensions $D=[5,10,15,20]$. In total, 228 function-dimension instances were used in the computational experiments.

For the experimental setting, we use similar rules to the recent CEC competitions. The search space for all the considered functions from all benchmarks was scaled to $[-100,100]^D$. The main termination criterion was reaching the maximum number of function evaluations (MaxFES), which was set to 50{,}000, 200{,}000, 500{,}000, and 1{,}000{,}000 evaluations for problems with $D=[5,10,15,20]$, respectively. All nature-inspired algorithms were run 30 times to get representative results (the deterministic methods were run only once). The true function value at the global minimum point of each of the considered functions $f(x^*)$ was known. This means that the function error could be computed as $f(x_{\min}) - f(x^*)$, where $f(x_{\min})$ was the minimal function value found by an algorithm within the search. These function error values achieved by the individual algorithms are presented in tables and figures in the upcoming sections. For each run, if the function error value of the resulting solution was less than or equal to 1E-08, it was considered as zero. All algorithms were run in a single thread environment in a MATLAB R2021b, on a PC with 3.2 GHz Core I5 processor, 16 GB RAM, and Windows 10.

\section{Results and Discussion}
The results of the computational comparison are presented for each benchmark set separately but the style of the presentation and the applied methods of statistical analysis are the same. Detailed convergence plots for all problems in the five benchmark sets can be found in Supplement A, while more detailed statistics about the individual runs of the algorithms can be found in Supplement B.

\subsection{CEC 2017 Results}
\begin{figure}[!h]
    \centering
        \includegraphics[width = 0.4\linewidth]{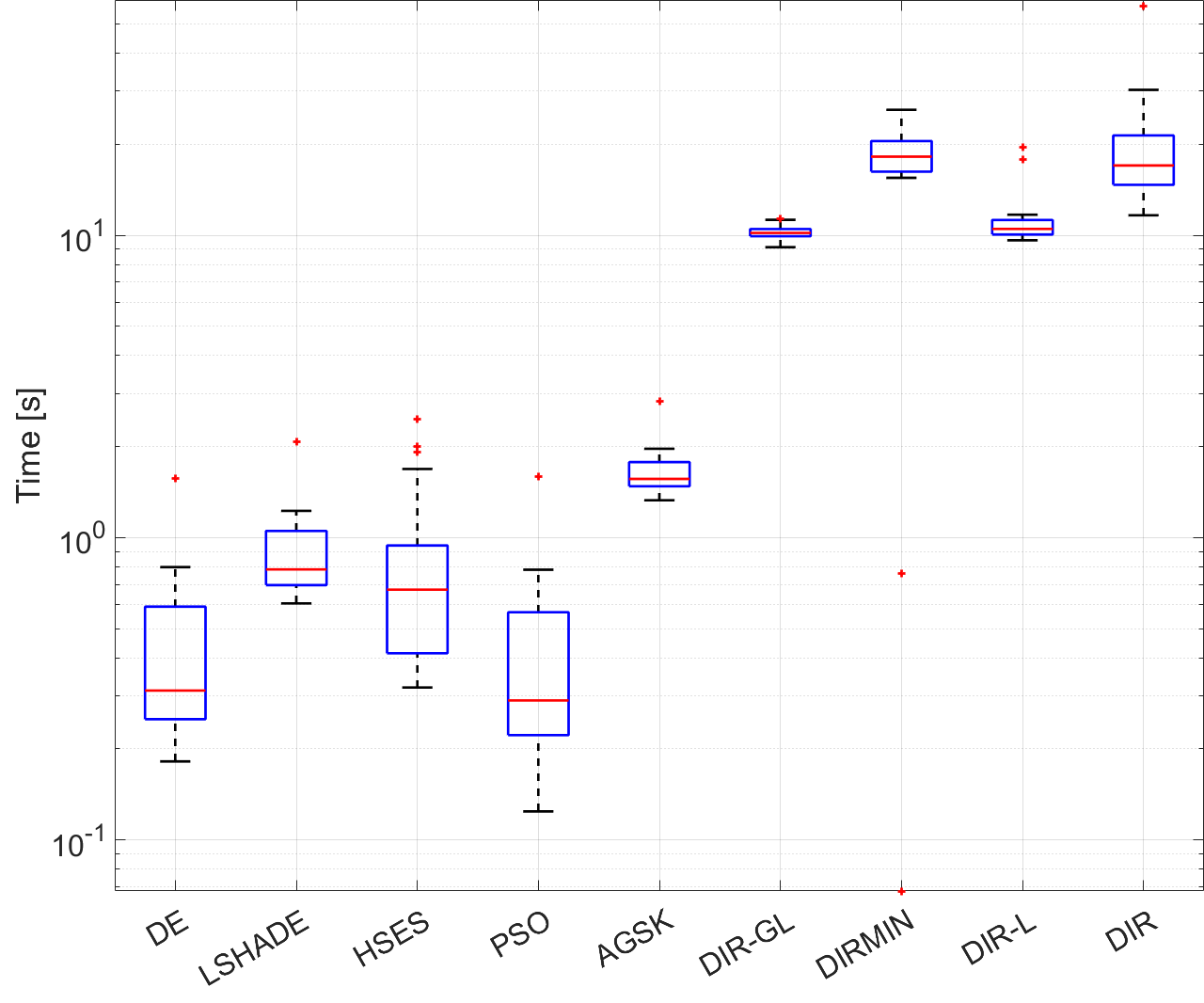} 
    \caption{Time complexity of the different algorithms on the CEC 2017 benchmark set, $D=10$.} \label{fig_cec17_time}
\end{figure}

First, we focus on the results on the CEC 2017 benchmark set, where we also describe in detail the different metrics used for the comparison of the algorithms. In Figure \ref{fig_cec17_time} are
shown the boxplots of time required for one run of the different methods (either reaching the MaxFES function evaluations or the 1E-08 error threshold). Even though there is some variability within the methods, the most easily perceptible difference is between the groups of nature-inspired and DIRECT-type methods. This difference is roughly between one and two orders of magnitude. Also, within the nature-inspired methods, the ``simpler ones'' (DE and PSO) tend to have a slightly smaller time complexity than their more involved counterparts from the CEC competitions. 

This difference in time complexity between the nature-inspired and DIRECT-type methods has direct effect on the way we compare the performance of the two groups of algorithms. If we can consider the evaluations of the objective function to be ``cheap'' (i.e., having negligible time), we could compare the results of the DIRECT-type methods against the best results of the nature-inspired ones over the 30 runs, as the time spent by these methods would be roughly the same. If, on the other hand, we consider the objective function evaluations as “costly” (i.e., we are really given only the MaxFES evaluations as they are computationally expensive), it would make more sense to compare the results of the DIRECT-type methods against the median (risk-neural) or the worst (risk-averse) results of the nature-inspired ones.

In Table \ref{cec17_res} are reported detailed statistics of the results of the computations on the CEC 2022 benchmark set for all nine considered algorithms (after reaching the MaxFES). When comparing the results of the DIRECT-type methods against the best results of the nature-inspired ones over the 30 runs, a clear verdict can be made. On most problems, the nature-inspired methods, in particular AGSK, DE, and LSHADE, outperform the DIRECT-type ones by a large margin. When comparing the median of the runs, situation changes slightly. While AGSK, DE, and LSHADE remain as the overall best methods, one of the DIRECT-type methods, DIRMIN, was the best performing one on F23, F24, F25, and F27. Also, both DIRMIN and DIR-GL were more successful than PSO. This shift towards relative improvement of the DIRECT-type methods persists into the comparison against the worst (max) results. Here, the best results are obtained by LSHADE, followed by AGSK and DE, while DIRMIN and DIR-GL are on par with HSES, and DIR and DIR-L are comparable to PSO.

Additional interesting results can gain from analyzing the converges of the different methods through the iterations. This can be done in several manners. The first option would be investigating the convergence plots (all of which can be found in Supplement A). The second option is to perform a statistical analysis of the ranks of the methods in different stages of the search. The third option is to set up additional precision targets and find the empirical cumulative distribution (ECD) function of the fraction of targets that were achieved throughout the evaluations.

\begin{figure}[!h]
    \centering
    \begin{tabular}{cc}
        \includegraphics[width = 0.45\linewidth]{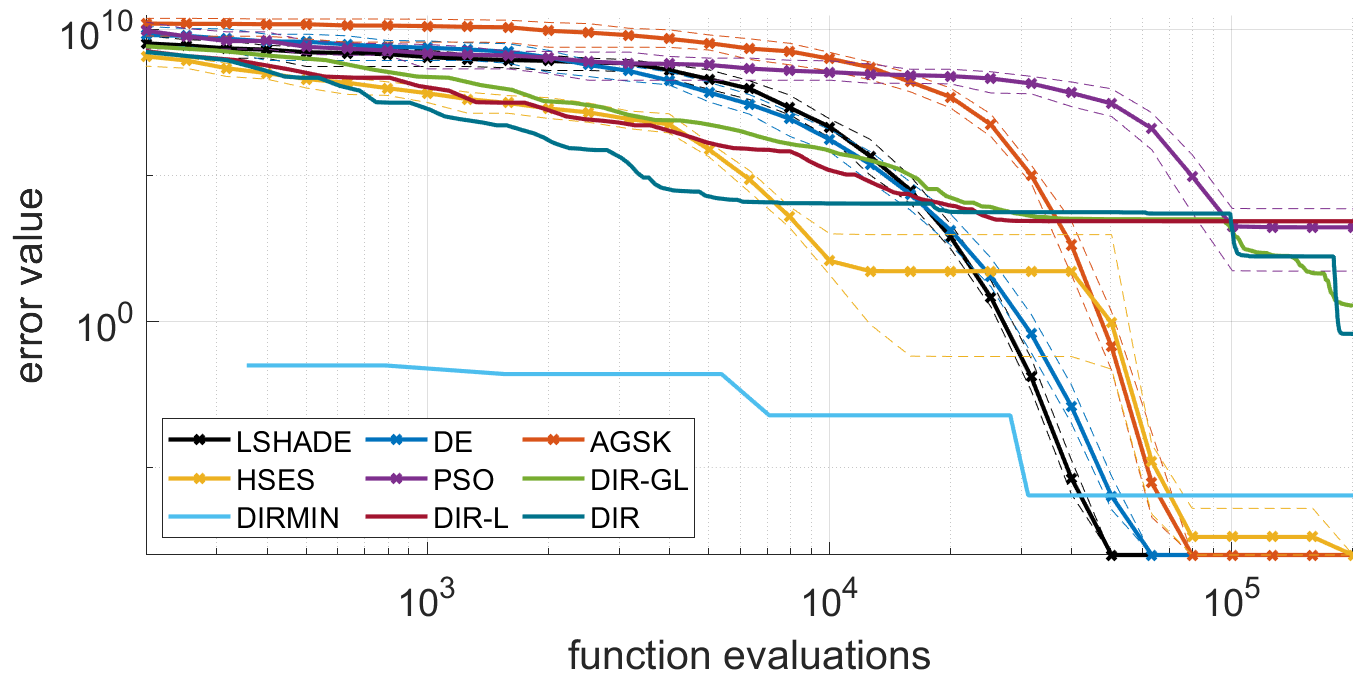} & \includegraphics[width = 0.45\linewidth]{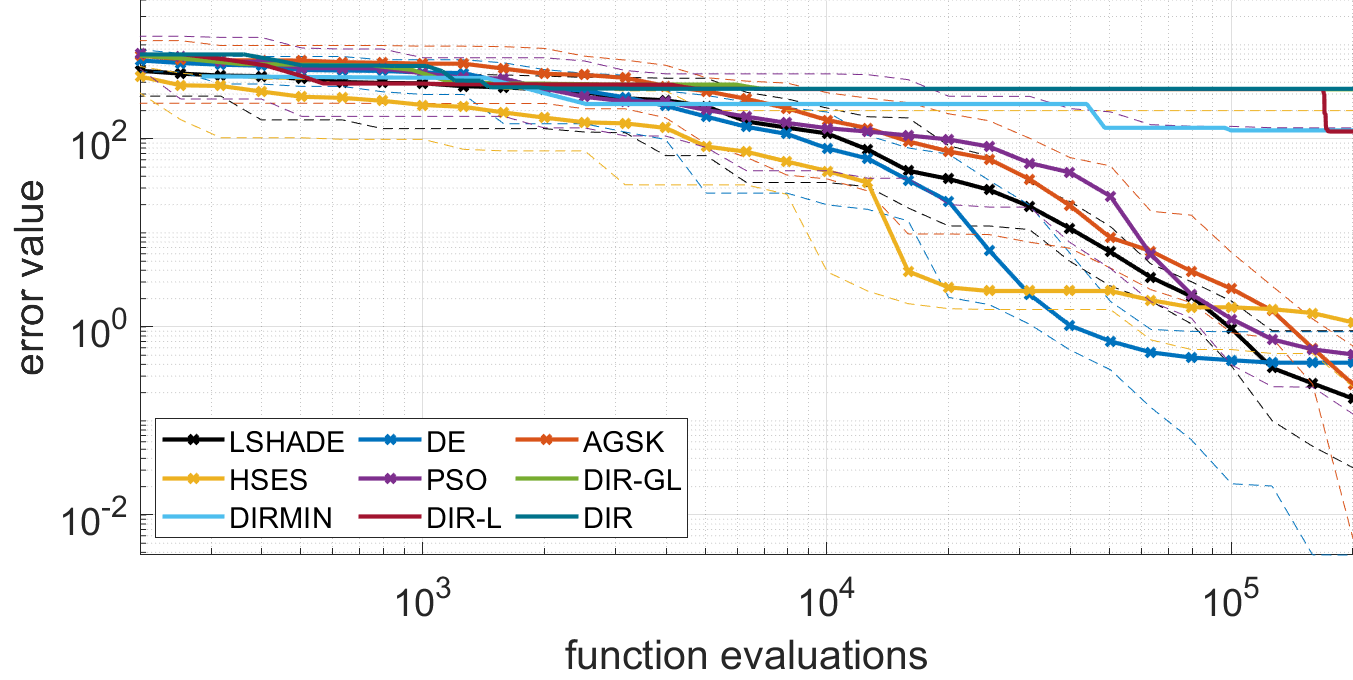} \\
        a) & b)
    \end{tabular}
    \caption{Convergence plots from the CEC 2017 benchmark set, a) F1 in $D=10$, b) F15 in $D=10$.} \label{conv_cec17}
\end{figure}

Two representative convergence plot are shown in Figure \ref{conv_cec17}, where for the nature-inspired methods the solid lines show the progression of the median values, while the dashed lines show the worst and best values. These convergence plots show a common theme in the performance of the algorithms. The DIRECT-type methods are generally well-equipped to find a reasonable solution in relatively small number of iterations (in the case on the left, it was DIRMIN that found a really good solution early on). However, they lack a good exploitation mechanism that would help them improve it, and after approximately $10^4$ function evaluations, the most of the DIRECT-type methods effectively stalled. Another interesting, an quite common, observation is that HSES was also able to find a reasonable solution relatively quickly, then stalled for a large portion of the iterations, but was able to eventually find an improvement.

The statistical analysis of the ranking of the nine methods at eight different stages of the search was performed by the use of the Friedman rank tests. The null hypothesis on equivalent efficiency of the algorithms was rejected at all stages of the search. The results of this analysis are reported in Table \ref{cec2017_ranks}, where the best three ranking methods for each stage of the search are highlighted in bold. It shows a similar theme to the convergence plot above. In the early stages of
the search, HSES and some the of the DIRECT-type methods (in this case DIR and DIRMIN) are able to find good solutions quickly. However, as the iterations progress, they get dominated by DE and LSHADE, and in the later stages also by AGSK. Overall, the best of the DIRECT-type methods is the DIRMIN, but its performance was worse than that of all the nature-inspired methods with the exception of PSO.

\begin{table}[!h]
\caption{Mean ranks from Friedman tests at different stages of the search, ambiguous benchmark set.}
\centering
\resizebox{0.6\columnwidth}{!}{%
\setlength{\tabcolsep}{0.3em}
\bgroup
\def\arraystretch{1.1}
\begin{tabular}{l|ccccccccc}
FES            & \multicolumn{1}{c}{DE} & \multicolumn{1}{c}{LSHADE} & \multicolumn{1}{c}{HSES} & \multicolumn{1}{c}{PSO} &\multicolumn{1}{c}{AGSK} & \multicolumn{1}{c}{DIR-GL} & \multicolumn{1}{c}{DIRMIN} & \multicolumn{1}{c}{DIR-L} & \multicolumn{1}{c}{DIR} \\ \hline 
(1/256)*MaxFES & 5.7 & 4.5 & \textbf{2.7} & 5.8 & 7.7 & 5.7 & \textbf{3.3} & 5.1 & \textbf{4.5} \\
(1/128)*MaxFES & 5.9 & 5.0 & \textbf{3.0} & 5.9 & 7.8 & 5.1 & \textbf{3.4} & 4.7 & \textbf{4.2} \\
(1/32)*MaxFES  & 4.2 & 5.8 & \textbf{4.1} & 6.8 & 7.4 & \textbf{4.1} & \textbf{2.9} & 4.8 & 4.7 \\
(1/8)*MaxFES   & \textbf{3.7} & \textbf{3.9} & 4.3 & 7.6 & 5.9 & 4.1 & \textbf{3.6} & 6.0 & 5.9 \\
(1/4)*MaxFES   & \textbf{3.7} & \textbf{3.3} & 4.8 & 7.5 & 4.8 & 4.2 & \textbf{4.0} & 6.6 & 6.1 \\
(1/2)*MaxFES   & \textbf{3.8} & \textbf{2.8} & \textbf{3.9} & 6.9 & 4.2 & 4.8 & 4.8 & 7.4 & 6.5 \\
(3/4)*MaxFES   & \textbf{3.7} & \textbf{2.8} & 4.1 & 6.2 & \textbf{3.7} & 5.2 & 5.0 & 7.6 & 6.7 \\
MaxFES         & \textbf{3.7} & \textbf{3.1} & 4.2 & 5.9 & \textbf{3.1} & 5.5 & 5.3 & 7.7 & 6.8
\end{tabular} \label{cec2017_ranks}
\egroup
}
\end{table}

\begin{figure}[!h]
    \centering
        \includegraphics[width = 0.4\linewidth]{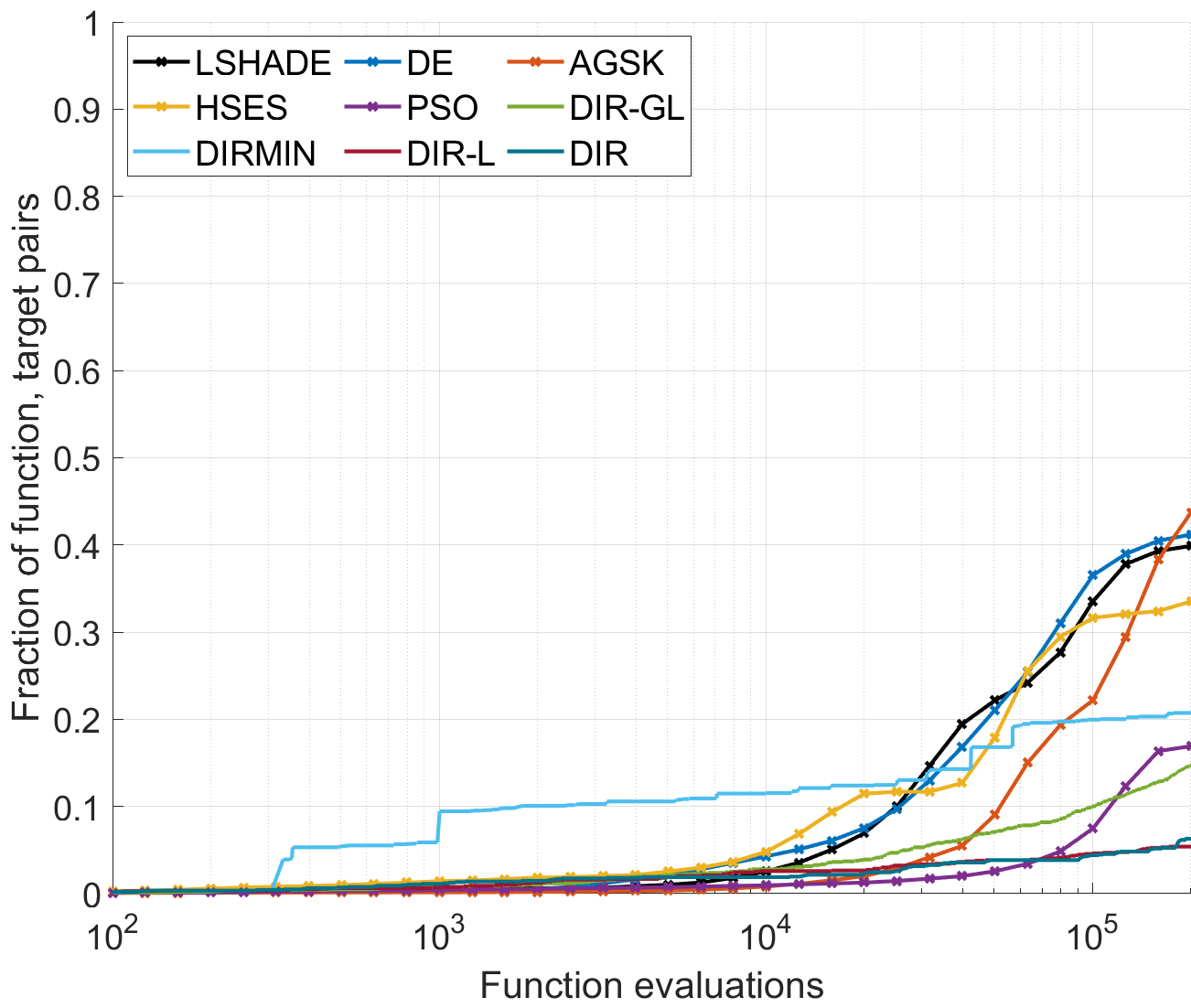} 
    \caption{ECD of the number of objective function evaluations for different target precisions on the CEC 2017 benchmark set, $D=10$.} \label{fig_cec17_ecd}
\end{figure}

The last comparison of the algorithms is done on the ECD of the number of objective function evaluations for 51 targets with target precision in $10^{[-8..2]}$ (this is the same setup used in the COCO platform for comparing continuous optimizers in a black-box setting \cite{hansen2021coco}). The results of this comparison are depicted in Figure \ref{fig_cec17_ecd} and show very similar results to the Friedman tests above. While DIRMIN was able to find good solutions (i.e., hitting many of the targets) relatively quickly, its convergence towards better solutions was very slow and was eventually overtaken by AGSK, DE, LSHADE, and HSES. On the other hand, we can see that even the best methods were only able to hit about 40\% of the precision targets, which signals that the benchmark set is relatively difficult and the methods would undoubtedly benefit from an increase in the maximum available function evaluations.

\begin{landscape}
\begin{table}[!h]
\caption{Detailed statistics of the 30 runs of the selected algorithms on the CEC 2017 benchmark set.}
\centering
%\scriptsize	
\resizebox{1.00\columnwidth}{!}{%
\setlength{\tabcolsep}{0.3em}
\bgroup
\def\arraystretch{1.3}
\begin{tabular}{cc|cccc|ccc|ccc|ccc|ccc|ccc}
\multicolumn{2}{l}{}& \multicolumn{4}{c}{Direct methods} & \multicolumn{3}{c}{DE} & \multicolumn{3}{c}{LSHADE} & \multicolumn{3}{c}{HSES} & \multicolumn{3}{c}{PSO} & \multicolumn{3}{c}{AGSK} \\
\multicolumn{1}{l}{} & D                    & DIR-GL               & DIRMIN               & DIR-L                & DIR                  & min                  & median   & max                            & min                  & median   & max                                  & min                  & median   & max                               & min                  & median   & max     & min                  & median   & max                               \\ \hline
F1  & 10                    & 2.53E+00                   & 1.09E-06                   & 2.65E+03                  & 3.69E-01                & 0.00E+00                & 0.00E+00                   & 0.00E+00                & 0.00E+00                & 0.00E+00                   & 0.00E+00                & 0.00E+00                & 0.00E+00                   & 0.00E+00                & 5.17E+01                & 1.64E+03                   & 7.09E+03                & 0.00E+00                & 0.00E+00                   & 0.00E+00                \\
F2  & 10                    & 7.61E-03                   & 0.00E+00                   & 2.93E-06                  & 2.33E+03                & 0.00E+00                & 0.00E+00                   & 0.00E+00                & 0.00E+00                & 0.00E+00                   & 0.00E+00                & 0.00E+00                & 0.00E+00                   & 0.00E+00                & 0.00E+00                & 0.00E+00                   & 0.00E+00                & 0.00E+00                & 0.00E+00                   & 0.00E+00                \\
F3  & 10                    & 1.69E-01                   & 0.00E+00                   & 5.59E+00                  & 1.59E+00                & 0.00E+00                & 0.00E+00                   & 0.00E+00                & 0.00E+00                & 0.00E+00                   & 0.00E+00                & 0.00E+00                & 0.00E+00                   & 0.00E+00                & 6.84E-01                & 3.11E+00                   & 4.06E+00                & 0.00E+00                & 0.00E+00                   & 0.00E+00                \\
F4  & 10                    & 3.98E+00                   & 7.96E+00                   & 1.99E+01                  & 7.96E+00                & 1.99E+00                & 1.59E+01                   & 2.26E+01                & 9.95E-01                & 1.99E+00                   & 3.98E+00                & 0.00E+00                & 9.95E-01                   & 1.99E+00                & 2.98E+00                & 6.47E+00                   & 1.71E+01                & 9.95E-01                & 4.97E+00                   & 6.96E+00                \\
F5  & 10                    & 1.03E-03                   & 8.00E+00                   & 6.27E+00                  & 1.07E+00                & 0.00E+00                & 0.00E+00                   & 0.00E+00                & 0.00E+00                & 0.00E+00                   & 0.00E+00                & 0.00E+00                & 0.00E+00                   & 0.00E+00                & 0.00E+00                & 0.00E+00                   & 0.00E+00                & 0.00E+00                & 0.00E+00                   & 3.55E-07                \\
F6  & 10                    & 1.53E+01                   & 3.65E+01                   & 5.26E+01                  & 2.49E+01                & 2.20E+01                & 3.16E+01                   & 3.85E+01                & 1.07E+01                & 1.22E+01                   & 1.34E+01                & 1.04E+01                & 1.11E+01                   & 1.18E+01                & 3.98E+00                & 1.79E+01                   & 2.46E+01                & 1.05E+01                & 1.61E+01                   & 1.97E+01                \\
F7  & 10                    & 3.98E+00                   & 3.98E+00                   & 3.38E+01                  & 5.97E+00                & 1.99E+00                & 1.54E+01                   & 2.96E+01                & 9.00E-04                & 1.99E+00                   & 3.98E+00                & 0.00E+00                & 0.00E+00                   & 2.98E+00                & 1.99E+00                & 6.47E+00                   & 1.19E+01                & 2.98E+00                & 5.97E+00                   & 1.29E+01                \\
F8  & 10                    & 8.54E-08                   & 8.00E+00                   & 1.30E+01                  & 2.35E-05                & 0.00E+00                & 0.00E+00                   & 0.00E+00                & 0.00E+00                & 0.00E+00                   & 0.00E+00                & 0.00E+00                & 0.00E+00                   & 0.00E+00                & 0.00E+00                & 0.00E+00                   & 0.00E+00                & 0.00E+00                & 0.00E+00                   & 0.00E+00                \\
F9  & 10                    & 1.29E+02                   & 2.44E+02                   & 5.37E+02                  & 1.29E+02                & 2.07E+00                & 7.47E+02                   & 1.15E+03                & 3.54E+00                & 8.62E+00                   & 1.29E+02                & 3.12E-01                & 3.63E+00                   & 3.98E+02                & 4.01E+01                & 2.70E+02                   & 7.15E+02                & 7.28E+01                & 2.53E+02                   & 3.79E+02                \\
F10 & 10                    & 1.14E+01                   & 4.97E+00                   & 1.05E+01                  & 1.06E+04                & 0.00E+00                & 0.00E+00                   & 9.95E-01                & 0.00E+00                & 0.00E+00                   & 1.25E+00                & 0.00E+00                & 0.00E+00                   & 9.95E-01                & 9.98E-01                & 3.00E+00                   & 8.68E+00                & 0.00E+00                & 0.00E+00                   & 0.00E+00                \\
F11 & 10                    & 1.09E+03                   & 1.15E+01                   & 2.10E+05                  & 1.62E+06                & 0.00E+00                & 2.08E-01                   & 1.20E+02                & 0.00E+00                & 3.12E-01                   & 1.20E+02                & 2.47E-06                & 2.09E-01                   & 1.30E+02                & 4.29E+02                & 2.34E+04                   & 4.79E+04                & 3.46E-06                & 4.20E-01                   & 1.20E+02                \\
F12 & 10                    & 7.33E+02                   & 3.02E+00                   & 7.40E+02                  & 6.38E+07                & 0.00E+00                & 0.00E+00                   & 5.39E+00                & 0.00E+00                & 4.84E+00                   & 5.39E+00                & 0.00E+00                & 5.18E+00                   & 6.51E+00                & 1.39E+01                & 1.83E+03                   & 1.61E+04                & 0.00E+00                & 4.51E-02                   & 4.84E+00                \\
F13 & 10                    & 3.91E+01                   & 2.73E+01                   & 3.22E+04                  & 5.30E+06                & 0.00E+00                & 0.00E+00                   & 9.95E-01                & 0.00E+00                & 0.00E+00                   & 9.95E-01                & 0.00E+00                & 9.88E-04                   & 6.53E+03                & 1.80E+00                & 2.22E+01                   & 3.52E+01                & 0.00E+00                & 0.00E+00                   & 2.88E-05                \\
F14 & 10                    & 3.01E+02                   & 9.99E-01                   & 3.07E+02                  & 1.40E+07                & 3.46E-06                & 8.68E-04                   & 1.48E-01                & 6.15E-06                & 1.21E-02                   & 5.00E-01                & 3.26E-04                & 4.99E-01                   & 4.19E+03                & 1.43E+00                & 1.52E+01                   & 6.00E+01                & 5.50E-06                & 1.10E-03                   & 3.45E-02                \\
F15 & 10                    & 3.37E+02                   & 1.23E+02                   & 1.20E+02                  & 3.41E+02                & 3.75E-03                & 4.16E-01                   & 8.89E-01                & 3.18E-02                & 1.73E-01                   & 9.10E-01                & 2.30E-01                & 1.12E+00                   & 2.19E+02                & 1.19E-01                & 5.09E-01                   & 1.31E+02                & 5.58E-03                & 2.45E-01                   & 6.26E-01                \\
F16 & 10                    & 4.63E+00                   & 6.83E+01                   & 6.68E+00                  & 3.60E+02                & 0.00E+00                & 3.12E-01                   & 9.95E-01                & 2.52E-04                & 2.42E-02                   & 3.12E-01                & 3.45E-01                & 7.64E+00                   & 3.82E+01                & 3.11E-07                & 4.46E+00                   & 3.92E+01                & 0.00E+00                & 7.14E-02                   & 7.18E-01                \\
F17 & 10                    & 2.53E+04                   & 1.54E+00                   & 6.56E+03                  & 1.85E+07                & 1.49E-07                & 3.52E-03                   & 4.26E-01                & 3.04E-05                & 3.01E-02                   & 5.00E-01                & 2.91E-01                & 2.27E+03                   & 7.19E+03                & 9.37E+00                & 1.41E+03                   & 1.67E+04                & 5.48E-06                & 2.61E-03                   & 4.53E-01                \\
F18 & 10                    & 1.03E+04                   & 4.93E+00                   & 2.29E+03                  & 2.41E+07                & 0.00E+00                & 0.00E+00                   & 1.94E-02                & 0.00E+00                & 2.53E-08                   & 1.97E-02                & 3.21E-03                & 1.54E-01                   & 6.57E+00                & 1.02E+00                & 9.53E+00                   & 2.63E+01                & 0.00E+00                & 1.94E-02                   & 5.11E-02                \\
F19 & 10                    & 4.19E+01                   & 1.04E+02                   & 1.98E+02                  & 5.21E+02                & 0.00E+00                & 3.12E-01                   & 1.31E+00                & 0.00E+00                & 0.00E+00                   & 0.00E+00                & 3.12E-01                & 2.08E+01                   & 5.36E+01                & 0.00E+00                & 9.66E-01                   & 3.68E+01                & 0.00E+00                & 0.00E+00                   & 0.00E+00                \\
F20 & 10                    & 1.00E+02                   & 1.00E+02                   & 1.00E+02                  & 1.01E+02                & 0.00E+00                & 1.00E+02                   & 2.25E+02                & 1.00E+02                & 1.51E+02                   & 2.06E+02                & 2.00E+02                & 2.02E+02                   & 2.03E+02                & 1.00E+02                & 2.10E+02                   & 2.22E+02                & 0.00E+00                & 1.00E+02                   & 1.00E+02                \\
F21 & 10                    & 1.01E+02                   & 1.00E+02                   & 1.05E+02                  & 2.30E+01                & 0.00E+00                & 1.00E+02                   & 1.01E+02                & 1.00E+02                & 1.00E+02                   & 1.00E+02                & 1.00E+02                & 1.00E+02                   & 1.00E+02                & 1.16E+01                & 1.02E+02                   & 1.04E+02                & 0.00E+00                & 1.00E+02                   & 1.00E+02                \\
F22 & 10                    & 3.03E+02                   & 3.08E+02                   & 3.19E+02                  & 3.16E+02                & 3.00E+02                & 3.04E+02                   & 3.08E+02                & 3.00E+02                & 3.03E+02                   & 3.05E+02                & 3.00E+02                & 3.03E+02                   & 3.05E+02                & 3.00E+02                & 3.07E+02                   & 3.16E+02                & 0.00E+00                & 3.08E+02                   & 3.13E+02                \\
F23 & 10                    & 1.00E+02                   & 2.32E+01                   & 1.00E+02                  & 1.20E+02                & 1.00E+02                & 3.31E+02                   & 3.35E+02                & 1.00E+02                & 3.30E+02                   & 3.32E+02                & 1.00E+02                & 3.28E+02                   & 3.29E+02                & 1.00E+02                & 3.39E+02                   & 3.53E+02                & 0.00E+00                & 1.00E+02                   & 1.00E+02                \\
F24 & 10                    & 3.98E+02                   & 2.00E+02                   & 4.44E+02                  & 3.98E+02                & 3.98E+02                & 3.98E+02                   & 4.43E+02                & 3.98E+02                & 3.98E+02                   & 4.46E+02                & 4.44E+02                & 4.46E+02                   & 4.48E+02                & 3.98E+02                & 3.99E+02                   & 4.50E+02                & 1.00E+02                & 3.98E+02                   & 3.98E+02                \\
F25 & 10                    & 1.31E-05                   & 7.64E-06                   & 2.00E+02                  & 8.50E+01                & 3.00E+02                & 3.00E+02                   & 3.00E+02                & 3.00E+02                & 3.00E+02                   & 3.00E+02                & 3.00E+02                & 3.00E+02                   & 3.00E+02                & 2.00E+02                & 3.00E+02                   & 1.25E+03                & 0.00E+00                & 2.00E+02                   & 3.00E+02                \\
F26 & 10                    & 3.91E+02                   & 4.01E+02                   & 4.28E+02                  & 3.90E+02                & 3.87E+02                & 3.90E+02                   & 3.94E+02                & 3.89E+02                & 3.90E+02                   & 3.90E+02                & 3.94E+02                & 3.98E+02                   & 4.00E+02                & 3.87E+02                & 3.94E+02                   & 4.01E+02                & 3.87E+02                & 3.89E+02                   & 3.89E+02                \\
F27 & 10                    & 1.34E-03                   & 2.66E-05                   & 6.47E+02                  & 4.85E+01                & 3.00E+02                & 3.00E+02                   & 5.84E+02                & 3.00E+02                & 3.00E+02                   & 6.12E+02                & 5.84E+02                & 5.84E+02                   & 6.72E+02                & 3.00E+02                & 3.00E+02                   & 6.27E+02                & 0.00E+00                & 3.00E+02                   & 3.00E+02                \\
F28 & 10                    & 2.34E+02                   & 2.91E+02                   & 3.71E+02                  & 2.38E+02                & 2.26E+02                & 2.27E+02                   & 2.30E+02                & 2.28E+02                & 2.33E+02                   & 2.37E+02                & 2.47E+02                & 2.65E+02                   & 2.75E+02                & 2.29E+02                & 2.43E+02                   & 2.98E+02                & 2.28E+02                & 2.35E+02                   & 2.41E+02               

\end{tabular} \label{cec17_res}
\egroup
}
\end{table}
\end{landscape}

% \begin{landscape}
% \begin{table}[!h]
% \caption{Detailed statistics of the 30 runs of the selected algorithms on the CEC 2017 benchmark set.}
% \centering
% %\scriptsize	
% \label{cec17_res}
% \includegraphics[width = \linewidth]{tab1.png}
% \end{table}
% \end{landscape}

\subsection{DIRECTLib Results}
\begin{figure}[!h]
    \centering
    \begin{tabular}{cc}
        \includegraphics[width = 0.4\linewidth]{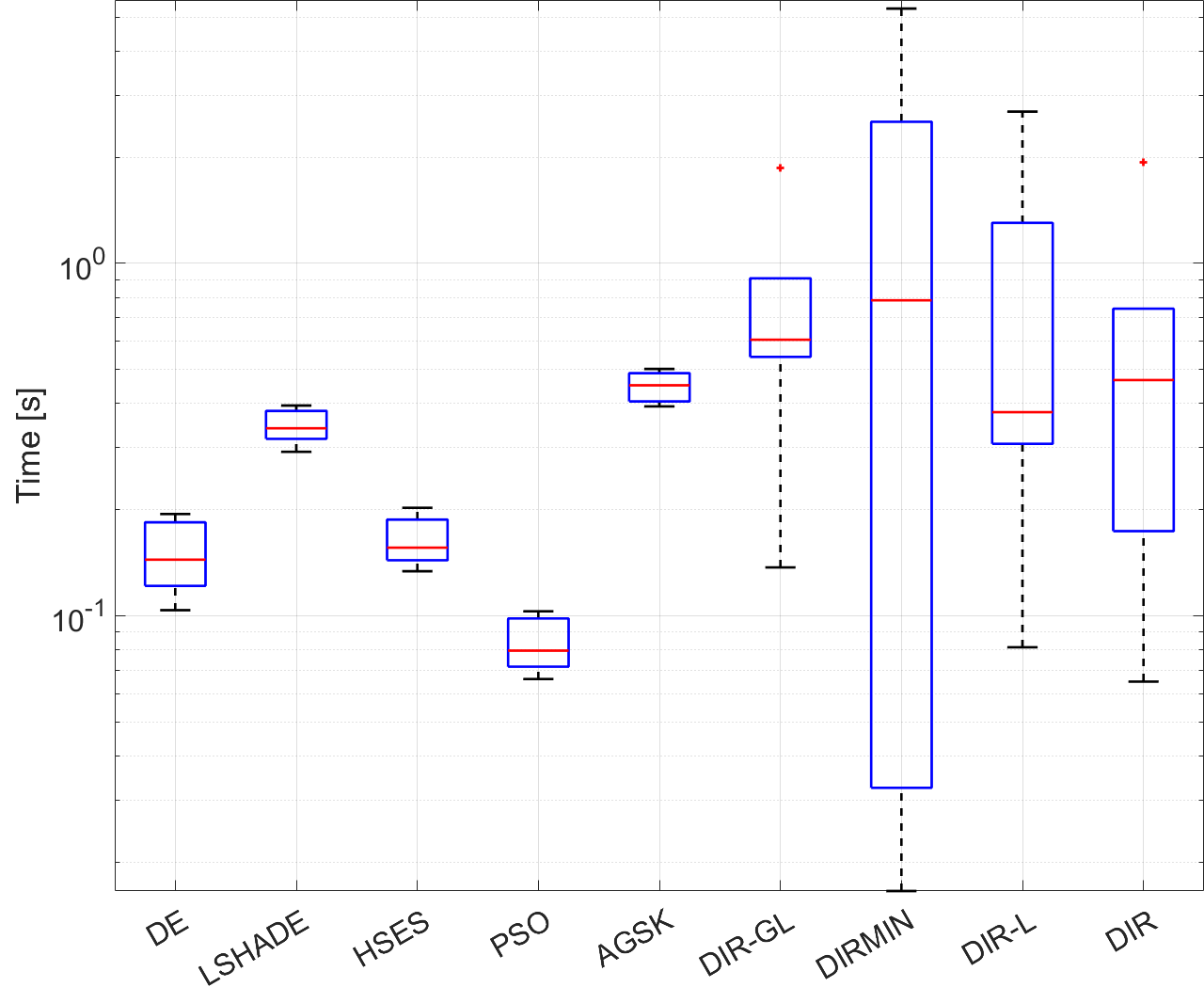} & \includegraphics[width = 0.4\linewidth]{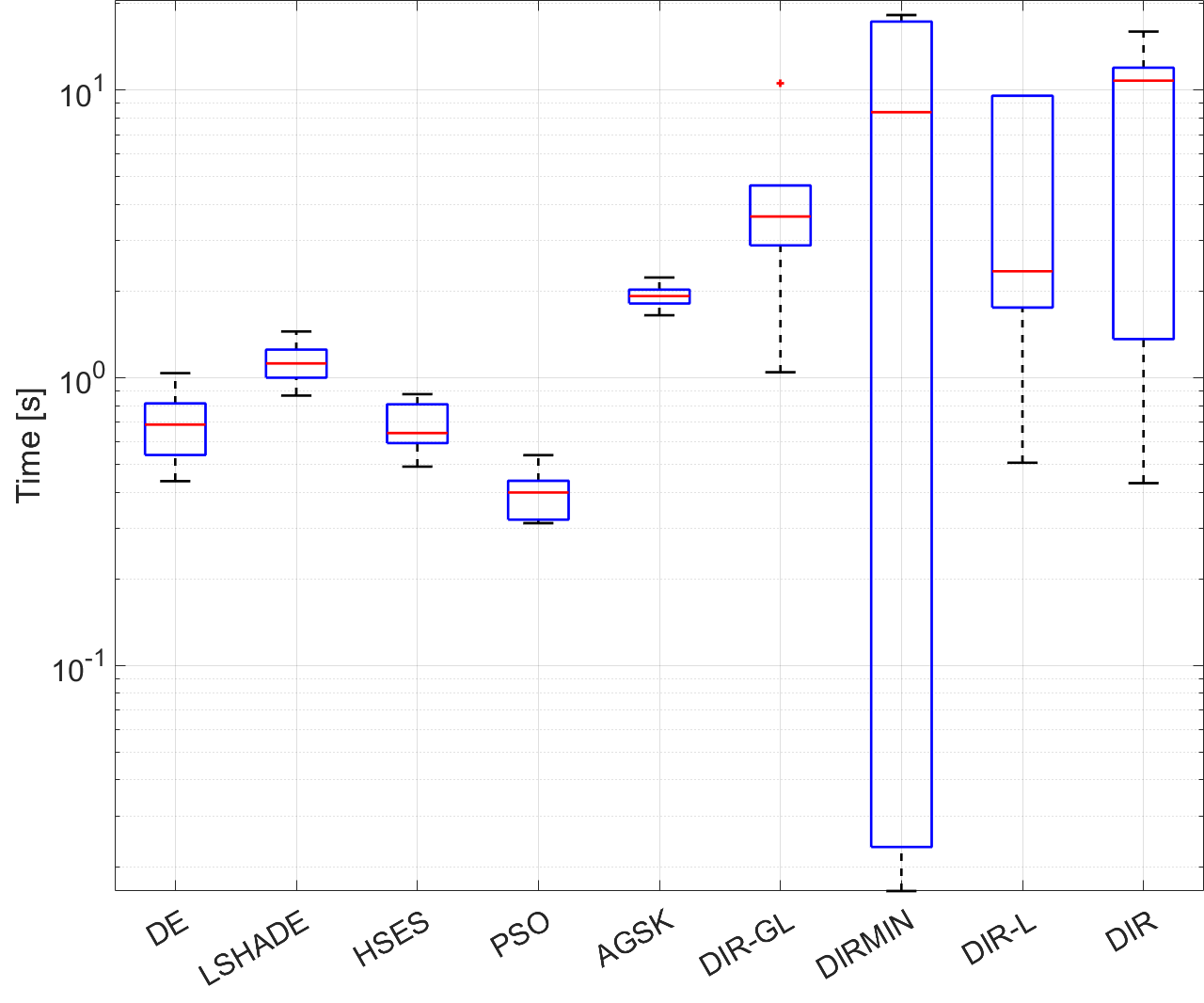} \\
        a) & b) \vspace{3mm}\\
        \includegraphics[width = 0.4\linewidth]{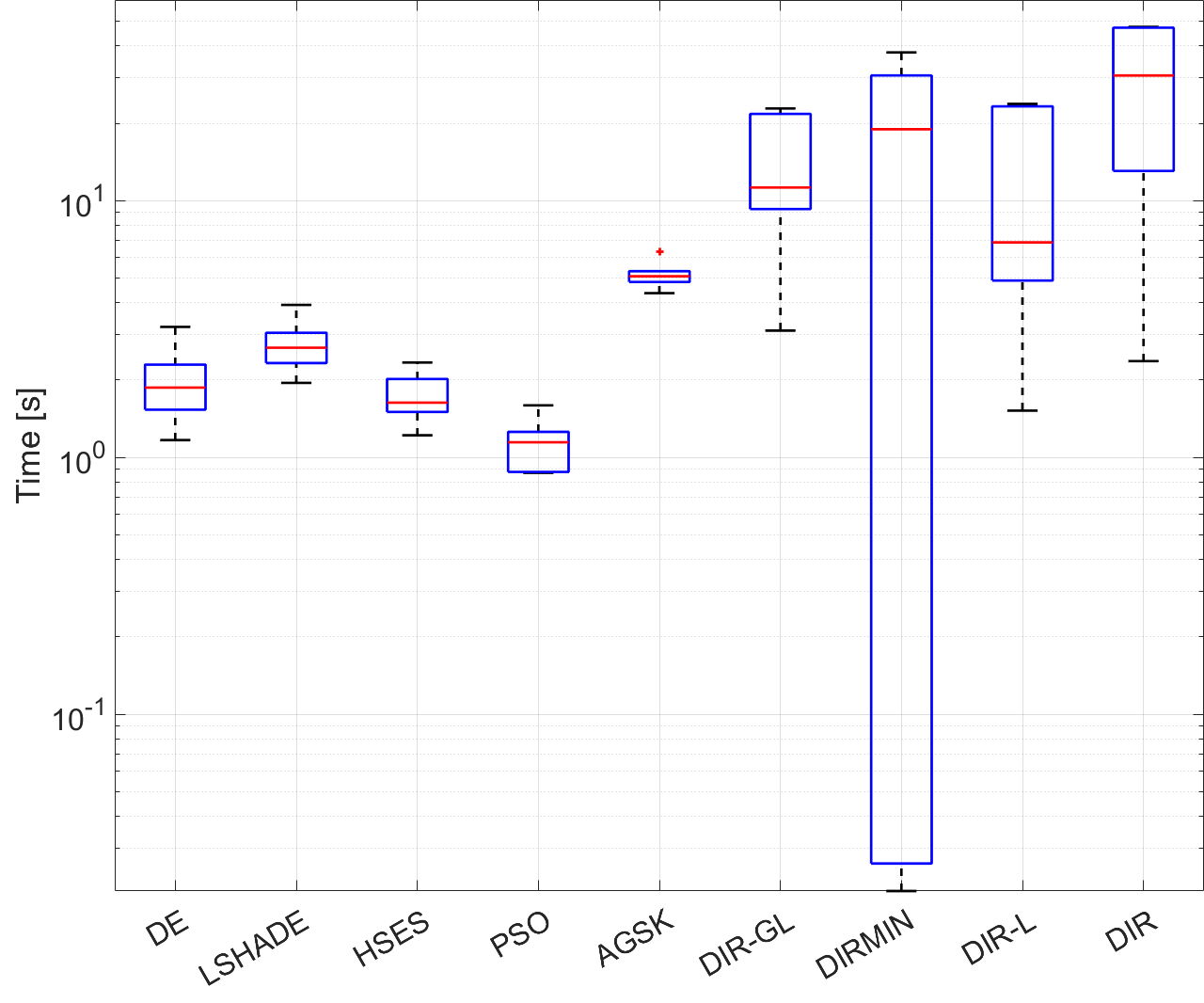} & \includegraphics[width = 0.4\linewidth]{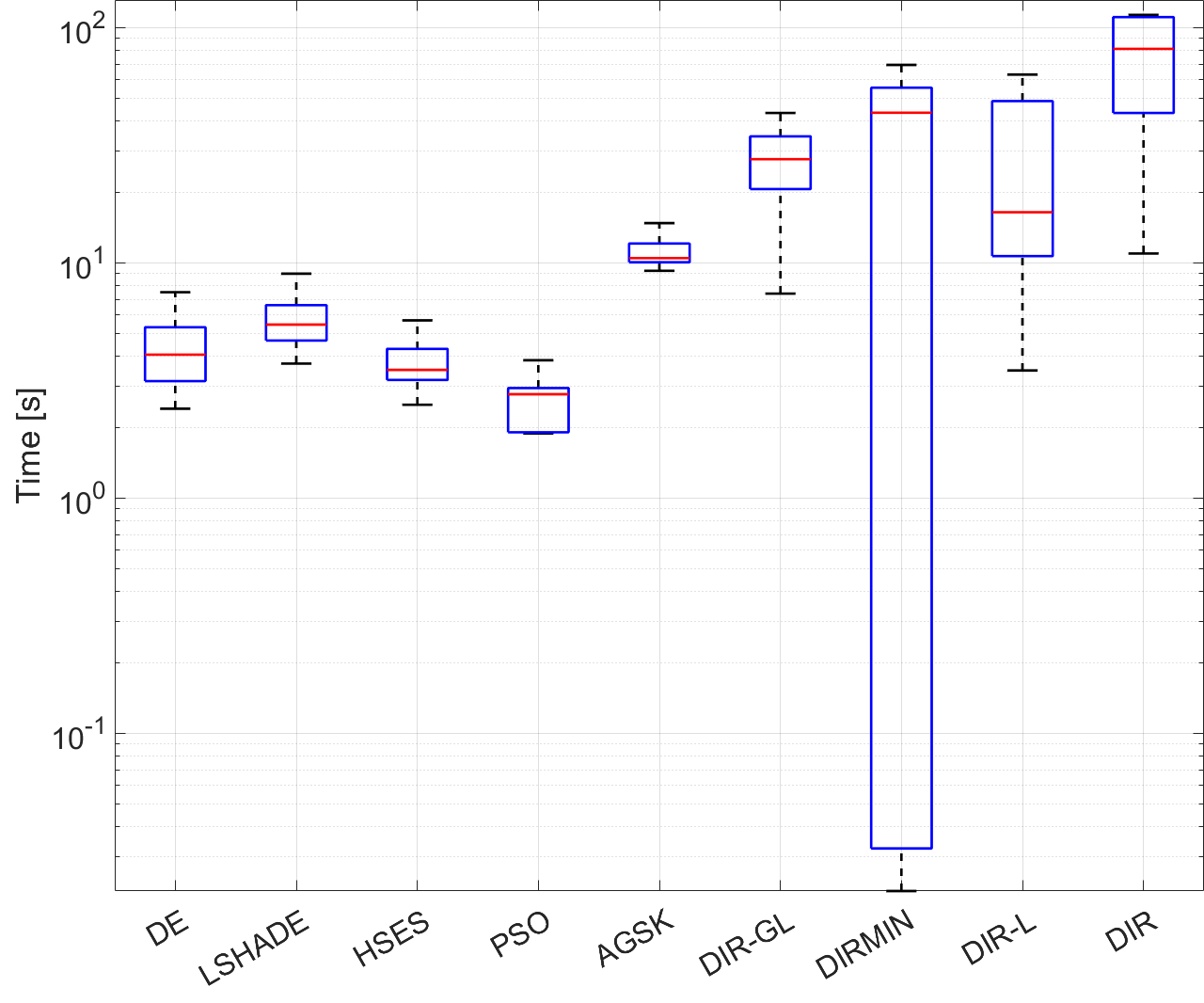} \\
        c) & d)
    \end{tabular}
    \caption{Time complexity of the different algorithms on the DIRECTLib benchmark set, a) $D=5$, b) $D=10$, c) $D=15$, d) $D=20$.} \label{fig_dirlib_time}
\end{figure}

Next, we investigate the results on the DIRECTLib benchmark set. In Figure \ref{fig_dirlib_time} are shown the computational times, where we can observe quite large variability for the DIRECT-type methods (in particular for DIRMIN). This variability is caused by their ability to reach the desired error threshold on many problems much sooner than approaching the MaxFES. However, they still remain roughly an order of magnitude slower than their nature-inspired counterparts.

\begin{figure}[!h]
    \centering
    \begin{tabular}{cc}
        \includegraphics[width = 0.45\linewidth]{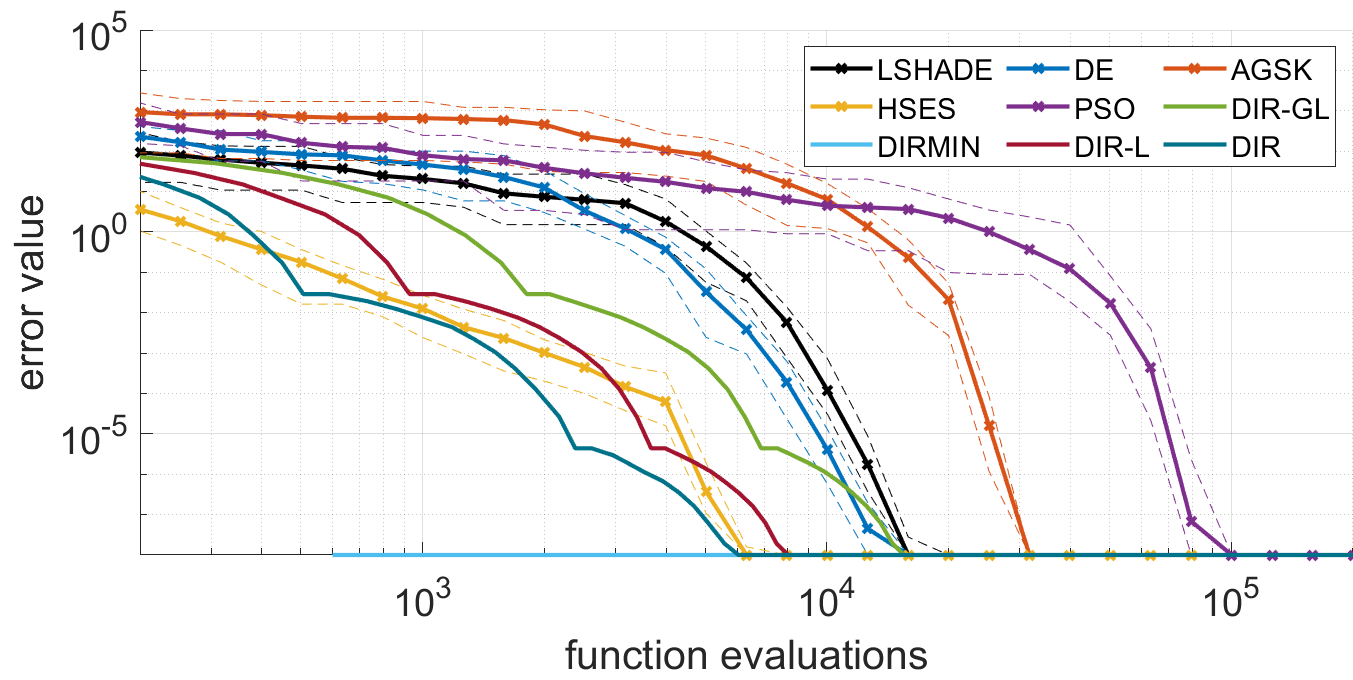} & \includegraphics[width = 0.45\linewidth]{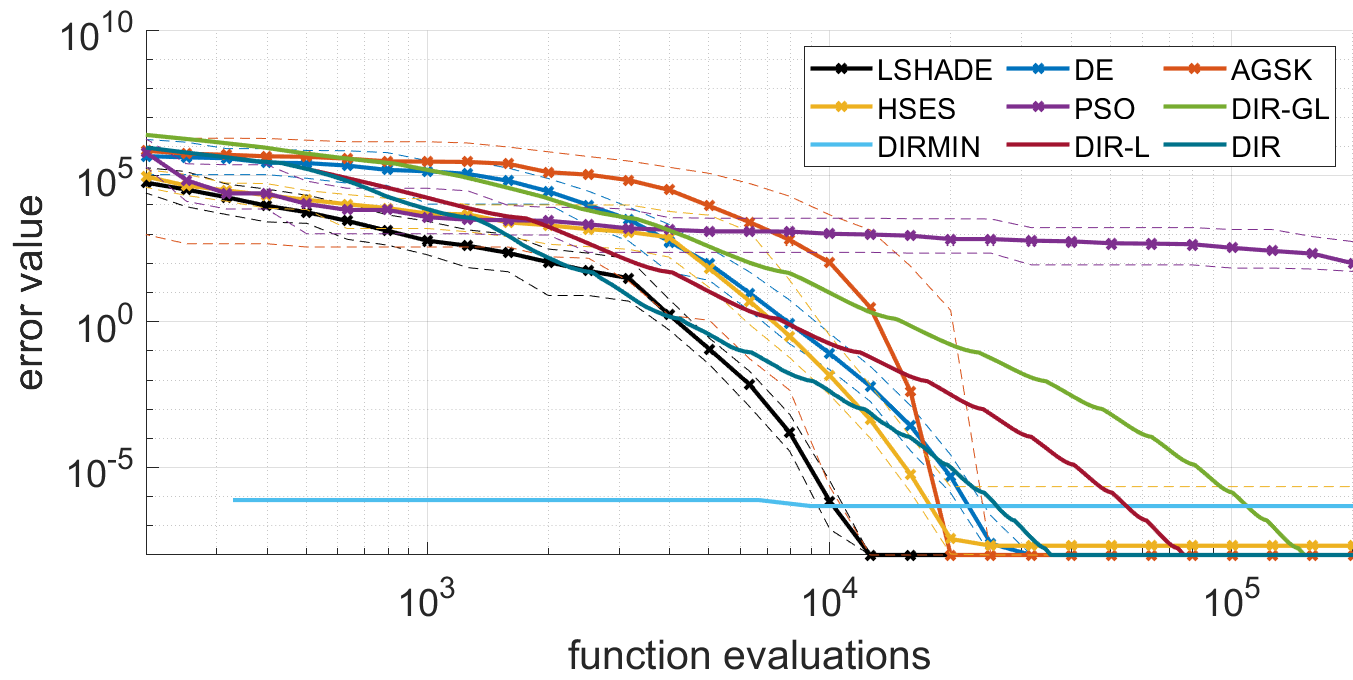} \\
        a) & b)
    \end{tabular}
    \caption{Convergences plot from the DIRECTLib benchmark set, a) F6 in $D=10$, b) F7 in $D=10$.} \label{conv_dirlib}
\end{figure}

The detailed statistics of the results, reported in Tables \ref{dirlib_res1} and \ref{dirlib_res2}, also show that for almost all problems (except for F16 in $D=20$), there was always at least one algorithm that could find a solution within the 1E-8 threshold. The two convergence plots in Figure \ref{conv_dirlib} exemplify the behaviour of the compared algorithms on the DIRECTLib benchmark set, where either all, or almost all methods were able to eventually find the optimal solution.

\begin{figure}[!h]
    \centering
    \begin{tabular}{cc}
        \includegraphics[width = 0.4\linewidth]{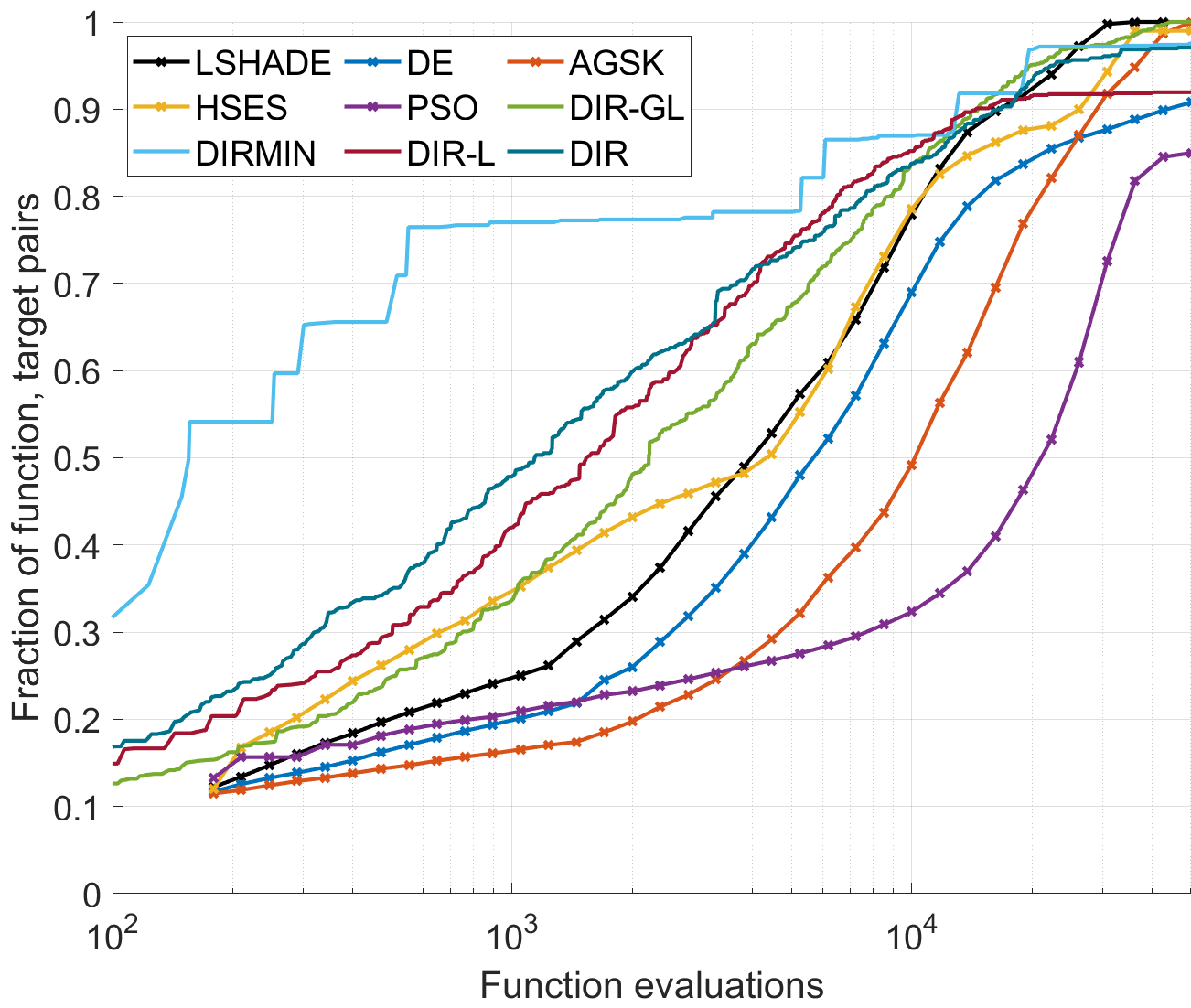} & \includegraphics[width = 0.4\linewidth]{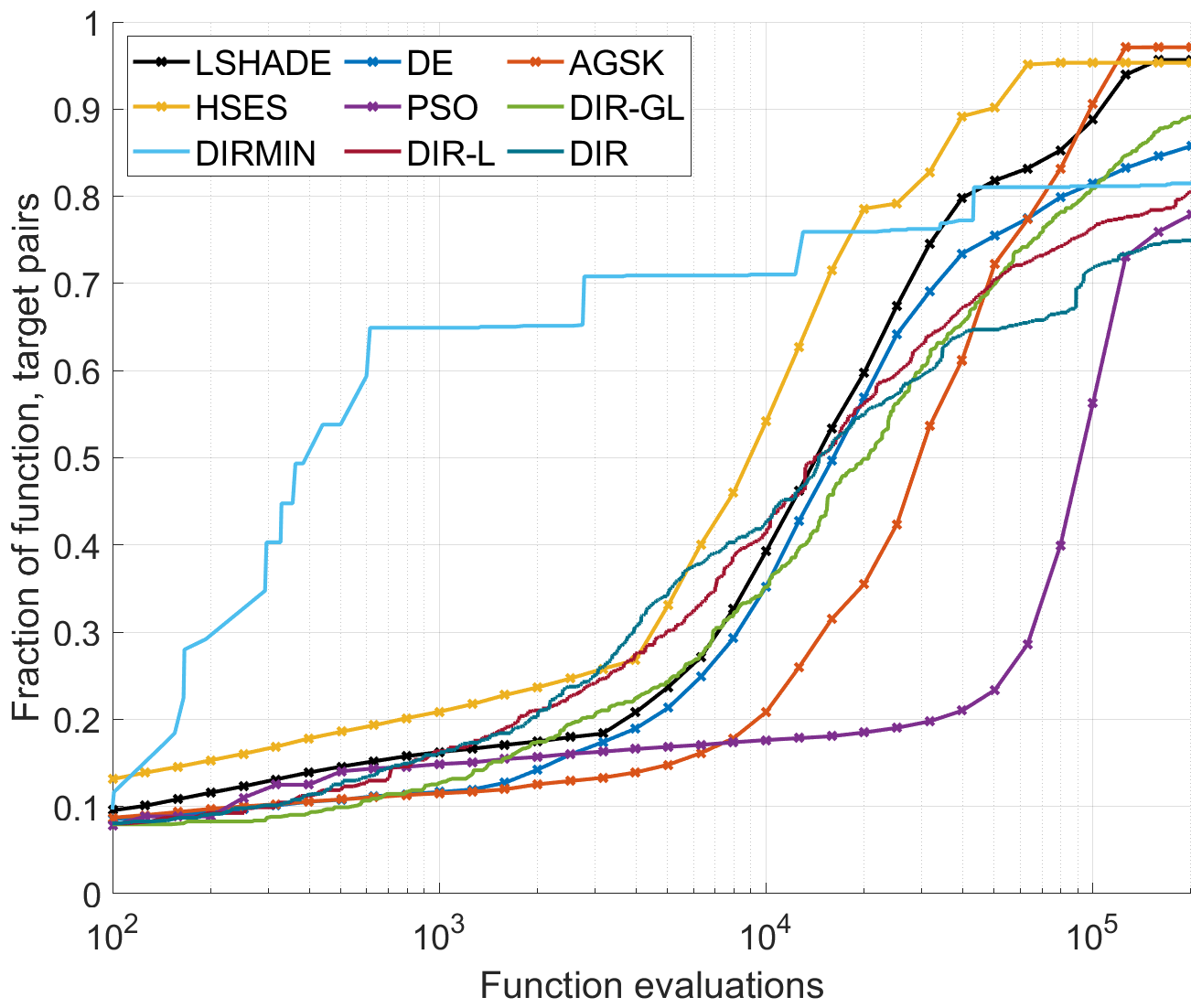} \\
        a) & b) \vspace{3mm}\\
        \includegraphics[width = 0.4\linewidth]{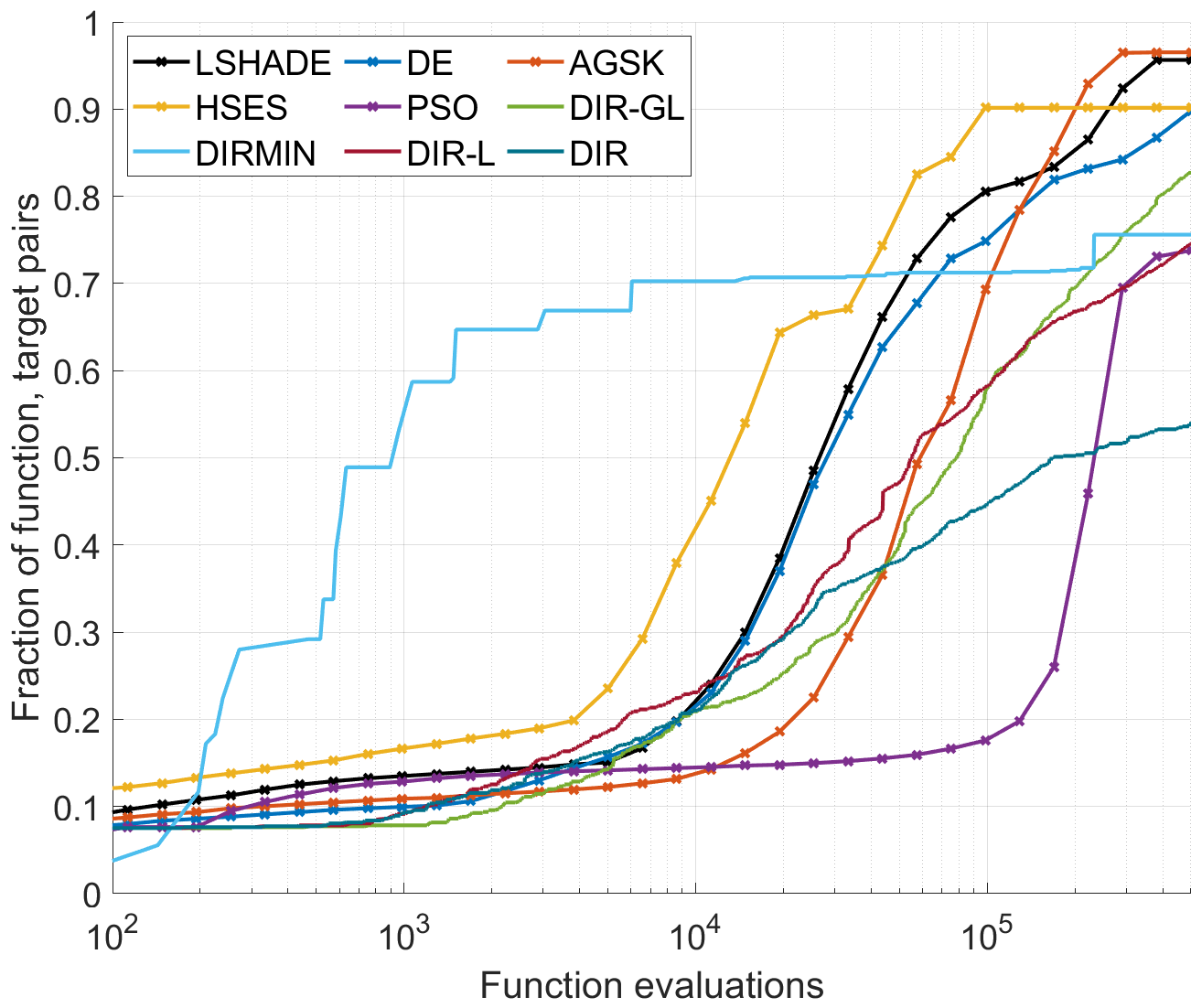} & \includegraphics[width = 0.4\linewidth]{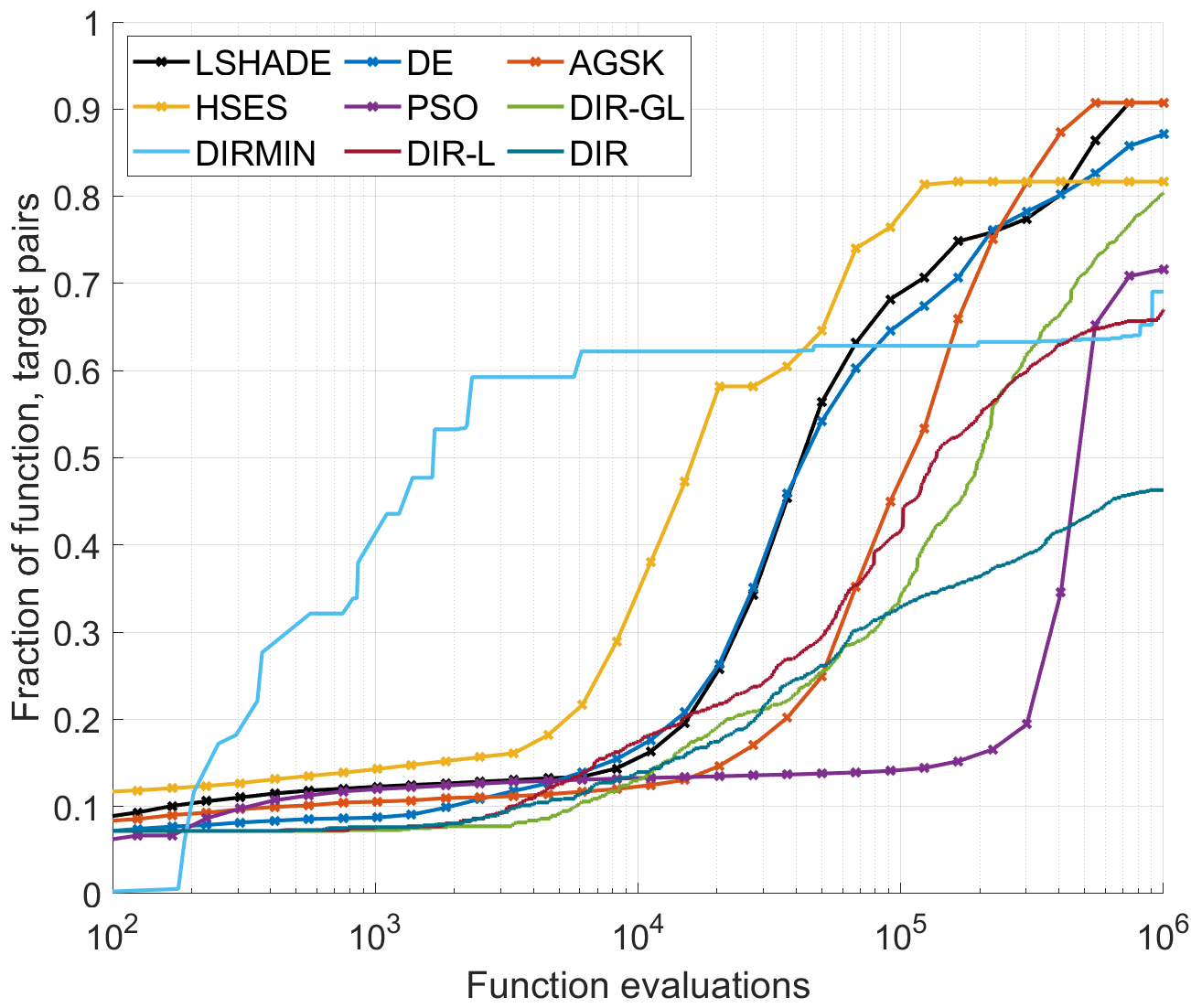} \\
        c) & d)
    \end{tabular}
    \caption{ECD of the number of objective function evaluations for different target precisions on the DIRECTLib benchmark set, a) $D=5$, b) $D=10$, c) $D=15$, d) $D=20$.} \label{fig_dirlib_ecd}
\end{figure}

\newpage
In these situations, the rank tests (reported in Table \ref{dirlib_ranks}) and the ECD plots (shown in Figure \ref{fig_dirlib_ecd}) are especially valuable. They reveal that at the start, the DIRECT-type methods (and especially DIRMIN) produce the best results, but as the number of function evaluations rises, they get outperformed by advanced nature-inspired algorithms (LSHADE, HSES, and AGSK). An interesting observation can be made in the lowest dimension ($D=5$), where the DIRECT-type methods had comparatively the best performance. Another notable observation is that the relative performance of DE improved as the dimensions increased. We can see also that the best methods were able to hit more than 90\% of the precision targets.

\begin{table}[!h]
\caption{Mean ranks from Friedman tests at different stages of the search, DIRECTLib set.}
\centering
\resizebox{0.6\columnwidth}{!}{%
\setlength{\tabcolsep}{0.3em}
\bgroup
\def\arraystretch{1.1}
\begin{tabular}{l|ccccccccc}
FES            & \multicolumn{1}{c}{DE} & \multicolumn{1}{c}{LSHADE} & \multicolumn{1}{c}{HSES} & \multicolumn{1}{c}{PSO} &\multicolumn{1}{c}{AGSK} & \multicolumn{1}{c}{DIR-GL} & \multicolumn{1}{c}{DIRMIN} & \multicolumn{1}{c}{DIR-L} & \multicolumn{1}{c}{DIR} \\ \hline 
(1/256)*MaxFES & 6.4 & 4.9 & \textbf{4.3} & 5.6 & 7.7 & 5.7 & \textbf{2.2} & \textbf{4.3} & \textbf{3.9} \\
(1/128)*MaxFES & 6.1 & 5.3 & \textbf{4.0} & 6.2 & 7.9 & 5.4 & \textbf{2.3} & \textbf{4.0} & \textbf{3.8} \\
(1/32)*MaxFES  & 5.3 & 5.0 & \textbf{3.8} & 7.7 & 7.5 & 4.6 & \textbf{2.7} & \textbf{3.9} & 4.4 \\
(1/8)*MaxFES   & 4.8 & \textbf{3.9} & \textbf{3.7} & 8.3 & 5.7 & 4.8 & \textbf{3.9} & 4.6 & 5.4 \\
(1/4)*MaxFES   & \textbf{4.4} & \textbf{3.7} & \textbf{3.6} & 8.3 & 4.8 & 4.9 & 4.7 & 4.9 & 5.6 \\
(1/2)*MaxFES   & 4.4 & \textbf{4.0} & \textbf{4.0} & 7.7 & \textbf{4.2} & 4.7 & 5.2 & 4.9 & 5.9 \\
(3/4)*MaxFES   & 4.6 & \textbf{4.0} & \textbf{4.3} & 5.8 & \textbf{4.1} & 5.0 & 5.6 & 5.3 & 6.1 \\
MaxFES         & 4.6 & \textbf{4.0} & \textbf{4.5} & 5.4 & \textbf{4.1} & 4.9 & 5.8 & 5.5 & 6.2
\end{tabular} \label{dirlib_ranks}
\egroup
}
\end{table}

\begin{landscape}
\begin{table}[!h]
\caption{Detailed statistics of the 30 runs of the selected algorithms on the DIRECTLib benchmark set, F1-F9.}
\centering
%\scriptsize	
\resizebox{1.00\columnwidth}{!}{%
\setlength{\tabcolsep}{0.3em}
\bgroup
\def\arraystretch{1.3}
\begin{tabular}{cc|cccc|ccc|ccc|ccc|ccc|ccc}
\multicolumn{2}{l}{}& \multicolumn{4}{c}{Direct methods} & \multicolumn{3}{c}{DE} & \multicolumn{3}{c}{LSHADE} & \multicolumn{3}{c}{HSES} & \multicolumn{3}{c}{PSO} & \multicolumn{3}{c}{AGSK} \\
\multicolumn{1}{l}{} & D                    & DIR-GL               & DIRMIN               & DIR-L                & DIR                  & min                  & median   & max                            & min                  & median   & max                                  & min                  & median   & max                               & min                  & median   & max     & min                  & median   & max                               \\ \hline
\multirow{4}{*}{F1}                      & 5                    & 0.00E+00                & 0.00E+00                & 0.00E+00                & 0.00E+00                & 1.94E-04 & 1.20E-03 & 2.87E-03 & 0.00E+00 & 0.00E+00 & 0.00E+00 & 0.00E+00             & 0.00E+00 & 0.00E+00             & 0.00E+00             & 0.00E+00 & 0.00E+00             & 0.00E+00             & 0.00E+00 & 0.00E+00             \\
                      & 10                   & 0.00E+00                & 0.00E+00                & 0.00E+00                & 0.00E+00                & 2.06E-02 & 5.15E-02 & 8.36E-02 & 0.00E+00 & 0.00E+00 & 0.00E+00 & 0.00E+00             & 0.00E+00 & 0.00E+00             & 0.00E+00             & 0.00E+00 & 0.00E+00             & 0.00E+00             & 0.00E+00 & 0.00E+00             \\
                      & 15                   & 0.00E+00                & 0.00E+00                & 0.00E+00                & 3.52E-03                & 1.04E-02 & 1.14E-01 & 1.81E-01 & 0.00E+00 & 0.00E+00 & 0.00E+00 & 0.00E+00             & 0.00E+00 & 0.00E+00             & 0.00E+00             & 0.00E+00 & 0.00E+00             & 0.00E+00             & 0.00E+00 & 0.00E+00             \\
                      & 20                   & 0.00E+00                & 0.00E+00                & 0.00E+00                & 2.17E-02                & 2.21E-03 & 1.68E-02 & 1.31E-01 & 0.00E+00 & 0.00E+00 & 0.00E+00 & 0.00E+00             & 0.00E+00 & 0.00E+00             & 0.00E+00             & 0.00E+00 & 0.00E+00             & 0.00E+00             & 0.00E+00 & 0.00E+00             \\ \hline
\multirow{4}{*}{F2}                      & 5                    & 0.00E+00                & 0.00E+00                & 0.00E+00                & 0.00E+00                & 4.42E-04 & 2.58E-03 & 5.30E-03 & 0.00E+00 & 0.00E+00 & 0.00E+00 & 0.00E+00             & 0.00E+00 & 0.00E+00             & 0.00E+00             & 0.00E+00 & 0.00E+00             & 0.00E+00             & 0.00E+00 & 9.91E-08             \\
                      & 10                   & 0.00E+00                & 0.00E+00                & 0.00E+00                & 0.00E+00                & 2.43E-04 & 2.21E-02 & 8.65E-02 & 0.00E+00 & 0.00E+00 & 0.00E+00 & 0.00E+00             & 0.00E+00 & 0.00E+00             & 0.00E+00             & 0.00E+00 & 0.00E+00             & 0.00E+00             & 0.00E+00 & 0.00E+00             \\
                      & 15                   & 0.00E+00                & 0.00E+00                & 0.00E+00                & 3.25E-07                & 0.00E+00 & 6.89E-04 & 3.75E-02 & 0.00E+00 & 0.00E+00 & 0.00E+00 & 0.00E+00             & 0.00E+00 & 0.00E+00             & 0.00E+00             & 0.00E+00 & 0.00E+00             & 0.00E+00             & 0.00E+00 & 0.00E+00             \\
                      & 20                   & 0.00E+00                & 0.00E+00                & 0.00E+00                & 1.84E-04                & 0.00E+00 & 0.00E+00 & 1.49E-03 & 0.00E+00 & 0.00E+00 & 0.00E+00 & 0.00E+00             & 0.00E+00 & 0.00E+00             & 0.00E+00             & 0.00E+00 & 0.00E+00             & 0.00E+00             & 0.00E+00 & 0.00E+00             \\ \hline
\multirow{4}{*}{F3}                       & 5                    & 0.00E+00                & 0.00E+00                & 6.67E-01                & 0.00E+00                & 0.00E+00 & 0.00E+00 & 0.00E+00 & 0.00E+00 & 0.00E+00 & 0.00E+00 & 0.00E+00             & 0.00E+00 & 6.67E-01             & 0.00E+00             & 0.00E+00 & 0.00E+00             & 0.00E+00             & 0.00E+00 & 0.00E+00             \\
                      & 10                   & 6.67E-01                & 6.67E-01                & 6.67E-01                & 6.67E-01                & 6.67E-01 & 6.67E-01 & 6.67E-01 & 6.67E-01 & 6.67E-01 & 6.67E-01 & 6.67E-01             & 6.67E-01 & 6.67E-01             & 0.00E+00             & 6.67E-01 & 6.67E-01             & 0.00E+00             & 6.67E-01 & 6.67E-01             \\
                      & 15                   & 6.67E-01                & 6.67E-01                & 6.67E-01                & 6.67E-01                & 6.67E-01 & 6.67E-01 & 6.67E-01 & 6.67E-01 & 6.67E-01 & 6.67E-01 & 6.67E-01             & 6.67E-01 & 6.67E-01             & 0.00E+00             & 6.67E-01 & 6.67E-01             & 0.00E+00             & 6.67E-01 & 6.67E-01             \\
                      & 20                   & 6.67E-01                & 6.67E-01                & 6.67E-01                & 6.67E-01                & 6.67E-01 & 6.67E-01 & 6.67E-01 & 6.67E-01 & 6.67E-01 & 6.67E-01 & 6.67E-01             & 6.67E-01 & 6.67E-01             & 0.00E+00             & 6.67E-01 & 6.67E-01             & 6.67E-01             & 6.67E-01 & 6.67E-01             \\ \hline
\multirow{4}{*}{F4}                       & 5                    & 0.00E+00                & 0.00E+00                & 0.00E+00                & 0.00E+00                & 0.00E+00 & 0.00E+00 & 0.00E+00 & 0.00E+00 & 0.00E+00 & 0.00E+00 & 0.00E+00             & 0.00E+00 & 0.00E+00             & 0.00E+00             & 0.00E+00 & 0.00E+00             & 0.00E+00             & 0.00E+00 & 0.00E+00             \\
                      & 10                   & 0.00E+00                & 0.00E+00                & 0.00E+00                & 0.00E+00                & 0.00E+00 & 0.00E+00 & 0.00E+00 & 0.00E+00 & 0.00E+00 & 0.00E+00 & 0.00E+00             & 0.00E+00 & 0.00E+00             & 0.00E+00             & 0.00E+00 & 0.00E+00             & 0.00E+00             & 0.00E+00 & 0.00E+00             \\
                      & 15                   & 0.00E+00                & 0.00E+00                & 0.00E+00                & 0.00E+00                & 0.00E+00 & 0.00E+00 & 0.00E+00 & 0.00E+00 & 0.00E+00 & 0.00E+00 & 0.00E+00             & 0.00E+00 & 0.00E+00             & 0.00E+00             & 0.00E+00 & 0.00E+00             & 0.00E+00             & 0.00E+00 & 0.00E+00             \\
                      & 20                   & 0.00E+00                & 0.00E+00                & 0.00E+00                & 1.11E-08                & 0.00E+00 & 0.00E+00 & 0.00E+00 & 0.00E+00 & 0.00E+00 & 0.00E+00 & 0.00E+00             & 0.00E+00 & 0.00E+00             & 0.00E+00             & 0.00E+00 & 0.00E+00             & 0.00E+00             & 0.00E+00 & 0.00E+00             \\ \hline
\multirow{4}{*}{F5}                       & 5                    & 0.00E+00                & 0.00E+00                & 0.00E+00                & 0.00E+00                & 0.00E+00 & 0.00E+00 & 0.00E+00 & 0.00E+00 & 0.00E+00 & 0.00E+00 & 0.00E+00             & 0.00E+00 & 0.00E+00             & 0.00E+00             & 0.00E+00 & 1.54E+01             & 0.00E+00             & 0.00E+00 & 0.00E+00             \\
                      & 10                   & 0.00E+00                & 8.93E+01                & 1.58E-08                & 0.00E+00                & 0.00E+00 & 0.00E+00 & 0.00E+00 & 0.00E+00 & 0.00E+00 & 0.00E+00 & 0.00E+00             & 0.00E+00 & 0.00E+00             & 0.00E+00             & 0.00E+00 & 5.95E+01             & 0.00E+00             & 0.00E+00 & 0.00E+00             \\
                      & 15                   & 1.68E-08                & 1.62E+02                & 4.28E-02                & 2.88E+02                & 0.00E+00 & 0.00E+00 & 0.00E+00 & 0.00E+00 & 0.00E+00 & 0.00E+00 & 0.00E+00             & 0.00E+00 & 0.00E+00             & 0.00E+00             & 0.00E+00 & 0.00E+00             & 0.00E+00             & 0.00E+00 & 0.00E+00             \\
                      & 20                   & 0.00E+00                & 7.66E+02                & 6.96E-01                & 7.40E+02                & 0.00E+00 & 0.00E+00 & 0.00E+00 & 0.00E+00 & 0.00E+00 & 0.00E+00 & 0.00E+00             & 0.00E+00 & 8.06E+01             & 0.00E+00             & 0.00E+00 & 3.29E+02             & 0.00E+00             & 0.00E+00 & 0.00E+00             \\ \hline
\multirow{4}{*}{F6}                       & 5                    & 0.00E+00                & 0.00E+00                & 0.00E+00                & 0.00E+00                & 0.00E+00 & 0.00E+00 & 0.00E+00 & 0.00E+00 & 0.00E+00 & 0.00E+00 & 0.00E+00             & 0.00E+00 & 0.00E+00             & 0.00E+00             & 0.00E+00 & 0.00E+00             & 0.00E+00             & 0.00E+00 & 0.00E+00             \\
                      & 10                   & 0.00E+00                & 0.00E+00                & 0.00E+00                & 0.00E+00                & 0.00E+00 & 0.00E+00 & 0.00E+00 & 0.00E+00 & 0.00E+00 & 0.00E+00 & 0.00E+00             & 0.00E+00 & 0.00E+00             & 0.00E+00             & 0.00E+00 & 0.00E+00             & 0.00E+00             & 0.00E+00 & 0.00E+00             \\
                      & 15                   & 0.00E+00                & 0.00E+00                & 0.00E+00                & 0.00E+00                & 0.00E+00 & 0.00E+00 & 0.00E+00 & 0.00E+00 & 0.00E+00 & 0.00E+00 & 0.00E+00             & 0.00E+00 & 0.00E+00             & 0.00E+00             & 0.00E+00 & 0.00E+00             & 0.00E+00             & 0.00E+00 & 0.00E+00             \\
                      & 20                   & 0.00E+00                & 0.00E+00                & 0.00E+00                & 0.00E+00                & 0.00E+00 & 0.00E+00 & 0.00E+00 & 0.00E+00 & 0.00E+00 & 0.00E+00 & 0.00E+00             & 0.00E+00 & 0.00E+00             & 0.00E+00             & 0.00E+00 & 0.00E+00             & 0.00E+00             & 0.00E+00 & 0.00E+00             \\ \hline
\multirow{4}{*}{F7}                       & 5                    & 0.00E+00                & 2.38E-07                & 0.00E+00                & 0.00E+00                & 0.00E+00 & 0.00E+00 & 0.00E+00 & 0.00E+00 & 0.00E+00 & 0.00E+00 & 0.00E+00             & 0.00E+00 & 0.00E+00             & 5.53E-02             & 2.04E-01 & 5.11E-01             & 0.00E+00             & 0.00E+00 & 0.00E+00             \\
                      & 10                   & 0.00E+00                & 4.76E-07                & 0.00E+00                & 0.00E+00                & 0.00E+00 & 0.00E+00 & 0.00E+00 & 0.00E+00 & 0.00E+00 & 0.00E+00 & 0.00E+00             & 2.08E-08 & 3.94E-04             & 5.27E+01             & 1.00E+02 & 5.56E+02             & 0.00E+00             & 0.00E+00 & 0.00E+00             \\
                      & 15                   & 5.38E-08                & 7.14E-07                & 0.00E+00                & 0.00E+00                & 0.00E+00 & 0.00E+00 & 0.00E+00 & 0.00E+00 & 0.00E+00 & 0.00E+00 & 1.34E-02             & 1.64E+02 & 6.85E+05             & 3.21E+03             & 6.05E+03 & 1.47E+04             & 0.00E+00             & 0.00E+00 & 0.00E+00             \\
                      & 20                   & 1.31E-06                & 3.49E-06                & 0.00E+00                & 0.00E+00                & 0.00E+00 & 0.00E+00 & 0.00E+00 & 0.00E+00 & 0.00E+00 & 0.00E+00 & 1.83E+04             & 6.12E+06 & 1.43E+07             & 3.09E+04             & 1.08E+05 & 2.66E+05             & 0.00E+00             & 0.00E+00 & 0.00E+00             \\ \hline
\multirow{4}{*}{F8}                       & 5                    & 0.00E+00                & 0.00E+00                & 4.72E-02                & 0.00E+00                & 0.00E+00 & 0.00E+00 & 0.00E+00 & 0.00E+00 & 0.00E+00 & 0.00E+00 & 0.00E+00             & 0.00E+00 & 0.00E+00             & 4.19E-02             & 2.05E-01 & 4.01E+00             & 0.00E+00             & 0.00E+00 & 0.00E+00             \\
                      & 10                   & 4.60E-02                & 0.00E+00                & 5.16E+00                & 4.89E-02                & 0.00E+00 & 0.00E+00 & 0.00E+00 & 0.00E+00 & 0.00E+00 & 0.00E+00 & 0.00E+00             & 0.00E+00 & 0.00E+00             & 1.83E+00             & 2.61E+00 & 7.20E+00             & 0.00E+00             & 0.00E+00 & 0.00E+00             \\
                      & 15                   & 5.59E-02                & 0.00E+00                & 1.07E+01                & 3.33E-01                & 0.00E+00 & 0.00E+00 & 0.00E+00 & 0.00E+00 & 0.00E+00 & 0.00E+00 & 0.00E+00             & 0.00E+00 & 0.00E+00             & 9.82E-03             & 6.47E+00 & 1.22E+01             & 0.00E+00             & 0.00E+00 & 0.00E+00             \\
                      & 20                   & 4.97E-02                & 0.00E+00                & 1.64E+01                & 1.62E+00                & 0.00E+00 & 0.00E+00 & 0.00E+00 & 0.00E+00 & 0.00E+00 & 0.00E+00 & 0.00E+00             & 0.00E+00 & 0.00E+00             & 9.67E+00             & 1.12E+01 & 1.73E+01             & 0.00E+00             & 0.00E+00 & 0.00E+00             \\ \hline
\multirow{4}{*}{F9}   & 5                    & 0.00E+00                & 2.00E-08                & 0.00E+00                & 0.00E+00                & 0.00E+00 & 0.00E+00 & 0.00E+00 & 0.00E+00 & 0.00E+00 & 0.00E+00 & 0.00E+00             & 1.22E-04 & 4.84E-04             & 0.00E+00             & 0.00E+00 & 2.55E-06             & 0.00E+00             & 0.00E+00 & 0.00E+00             \\
  & 10                   & 0.00E+00                & 2.01E-08                & 0.00E+00                & 0.00E+00                & 0.00E+00 & 0.00E+00 & 0.00E+00 & 0.00E+00 & 0.00E+00 & 0.00E+00 & 1.44E-07             & 5.18E-06 & 2.29E-05             & 0.00E+00             & 0.00E+00 & 0.00E+00             & 0.00E+00             & 0.00E+00 & 0.00E+00             \\
  & 15                   & 0.00E+00                & 2.00E-08                & 0.00E+00                & 0.00E+00                & 0.00E+00 & 0.00E+00 & 0.00E+00 & 0.00E+00 & 0.00E+00 & 0.00E+00 & 1.01E-08             & 5.87E-07 & 9.69E-06             & 0.00E+00             & 0.00E+00 & 0.00E+00             & 0.00E+00             & 0.00E+00 & 0.00E+00             \\
  & 20                   & 0.00E+00                & 2.00E-08                & 0.00E+00                & 0.00E+00                & 0.00E+00 & 0.00E+00 & 0.00E+00 & 0.00E+00 & 0.00E+00 & 0.00E+00 & 0.00E+00             & 5.74E-08 & 1.42E-05             & 0.00E+00             & 0.00E+00 & 0.00E+00             & 0.00E+00             & 0.00E+00 & 0.00E+00             \\
\end{tabular} \label{dirlib_res1}
\egroup
}
\end{table}
\end{landscape}

% \begin{landscape}
% \begin{table}[!h]
% \caption{Detailed statistics of the 30 runs of the selected algorithms on the DIRECTLib benchmark set, F1-F9.}
% \centering
% %\scriptsize	
% \includegraphics[width = \linewidth]{tab2.png}
% \label{dirlib_res1}
% \end{table}
% \end{landscape}

\begin{landscape}
\begin{table}[!h]
\caption{Detailed statistics of the 30 runs of the selected algorithms on the DIRECTLib benchmark set, F10-F18.}
\centering
%\scriptsize	
\resizebox{1.00\columnwidth}{!}{%
\setlength{\tabcolsep}{0.3em}
\bgroup
\def\arraystretch{1.3}
\begin{tabular}{cc|cccc|ccc|ccc|ccc|ccc|ccc}
\multicolumn{2}{l}{}& \multicolumn{4}{c}{Direct methods} & \multicolumn{3}{c}{DE} & \multicolumn{3}{c}{LSHADE} & \multicolumn{3}{c}{HSES} & \multicolumn{3}{c}{PSO} & \multicolumn{3}{c}{AGSK} \\
\multicolumn{1}{l}{} & D                    & DIR-GL               & DIRMIN               & DIR-L                & DIR                  & min                  & median   & max                            & min                  & median   & max                                  & min                  & median   & max                               & min                  & median   & max     & min                  & median   & max                               \\ \hline
\multirow{4}{*}{F10} & 5                    & 0.00E+00                & 0.00E+00                & 0.00E+00                & 0.00E+00                & 0.00E+00 & 0.00E+00 & 0.00E+00 & 0.00E+00 & 0.00E+00 & 0.00E+00 & 0.00E+00             & 0.00E+00 & 0.00E+00             & 0.00E+00             & 0.00E+00 & 0.00E+00             & 0.00E+00             & 0.00E+00 & 0.00E+00             \\
 & 10                   & 0.00E+00                & 0.00E+00                & 0.00E+00                & 0.00E+00                & 0.00E+00 & 0.00E+00 & 0.00E+00 & 0.00E+00 & 0.00E+00 & 0.00E+00 & 0.00E+00             & 0.00E+00 & 0.00E+00             & 0.00E+00             & 0.00E+00 & 0.00E+00             & 0.00E+00             & 0.00E+00 & 0.00E+00             \\
& 15                   & 0.00E+00                & 0.00E+00                & 0.00E+00                & 0.00E+00                & 0.00E+00 & 0.00E+00 & 0.00E+00 & 0.00E+00 & 0.00E+00 & 0.00E+00 & 0.00E+00             & 0.00E+00 & 0.00E+00             & 0.00E+00             & 0.00E+00 & 0.00E+00             & 0.00E+00             & 0.00E+00 & 0.00E+00             \\
 & 20                   & 0.00E+00                & 4.42E-07                & 0.00E+00                & 0.00E+00                & 0.00E+00 & 0.00E+00 & 0.00E+00 & 0.00E+00 & 0.00E+00 & 0.00E+00 & 0.00E+00             & 0.00E+00 & 0.00E+00             & 0.00E+00             & 0.00E+00 & 0.00E+00             & 0.00E+00             & 0.00E+00 & 0.00E+00             \\ \hline
\multirow{4}{*}{F11}                     & 5                    & 0.00E+00                & 2.38E-07                & 0.00E+00                & 0.00E+00                & 0.00E+00 & 0.00E+00 & 0.00E+00 & 0.00E+00 & 0.00E+00 & 0.00E+00 & 0.00E+00             & 0.00E+00 & 0.00E+00             & 3.76E-03             & 2.42E-02 & 4.87E-02             & 0.00E+00             & 0.00E+00 & 0.00E+00             \\
                     & 10                   & 0.00E+00                & 4.76E-07                & 0.00E+00                & 0.00E+00                & 0.00E+00 & 0.00E+00 & 0.00E+00 & 0.00E+00 & 0.00E+00 & 0.00E+00 & 0.00E+00             & 0.00E+00 & 2.28E-05             & 2.00E-01             & 3.14E-01 & 5.24E-01             & 0.00E+00             & 0.00E+00 & 0.00E+00             \\
                     & 15                   & 0.00E+00                & 1.92E-06                & 0.00E+00                & 0.00E+00                & 0.00E+00 & 0.00E+00 & 0.00E+00 & 0.00E+00 & 0.00E+00 & 0.00E+00 & 2.12E-08             & 3.54E-08 & 5.60E-08             & 6.22E-01             & 1.03E+00 & 1.87E+00             & 0.00E+00             & 0.00E+00 & 0.00E+00             \\
                     & 20                   & 0.00E+00                & 2.56E-06                & 0.00E+00                & 0.00E+00                & 0.00E+00 & 0.00E+00 & 0.00E+00 & 0.00E+00 & 0.00E+00 & 0.00E+00 & 2.29E-06             & 3.55E-06 & 5.61E-06             & 1.84E+00             & 2.34E+00 & 3.93E+00             & 0.00E+00             & 0.00E+00 & 0.00E+00             \\ \hline
\multirow{4}{*}{F12}                     & 5                    & 0.00E+00                & 1.98E-07                & 0.00E+00                & 0.00E+00                & 0.00E+00 & 0.00E+00 & 0.00E+00 & 0.00E+00 & 0.00E+00 & 0.00E+00 & 0.00E+00             & 0.00E+00 & 0.00E+00             & 3.57E-04             & 8.34E-04 & 1.32E-03             & 0.00E+00             & 0.00E+00 & 0.00E+00             \\
                     & 10                   & 0.00E+00                & 1.60E-06                & 0.00E+00                & 2.84E-05                & 0.00E+00 & 0.00E+00 & 0.00E+00 & 0.00E+00 & 0.00E+00 & 0.00E+00 & 0.00E+00             & 0.00E+00 & 0.00E+00             & 3.36E-03             & 6.08E-03 & 1.11E-02             & 0.00E+00             & 0.00E+00 & 0.00E+00             \\
                     & 15                   & 0.00E+00                & 3.00E-06                & 0.00E+00                & 3.84E-04                & 0.00E+00 & 0.00E+00 & 0.00E+00 & 0.00E+00 & 0.00E+00 & 0.00E+00 & 0.00E+00             & 0.00E+00 & 0.00E+00             & 7.87E-03             & 1.23E-02 & 1.57E-02             & 0.00E+00             & 0.00E+00 & 0.00E+00             \\
                     & 20                   & 0.00E+00                & 3.99E-06                & 0.00E+00                & 5.39E-04                & 0.00E+00 & 0.00E+00 & 0.00E+00 & 0.00E+00 & 0.00E+00 & 0.00E+00 & 0.00E+00             & 6.28E-08 & 7.81E-02             & 1.32E-02             & 2.03E-02 & 2.80E-02             & 0.00E+00             & 0.00E+00 & 0.00E+00             \\ \hline
\multirow{4}{*}{F13}                     & 5                    & 0.00E+00                & 0.00E+00                & 0.00E+00                & 0.00E+00                & 0.00E+00 & 0.00E+00 & 0.00E+00 & 0.00E+00 & 0.00E+00 & 0.00E+00 & 0.00E+00             & 0.00E+00 & 0.00E+00             & 0.00E+00             & 0.00E+00 & 0.00E+00             & 0.00E+00             & 0.00E+00 & 0.00E+00             \\
                     & 10                   & 0.00E+00                & 0.00E+00                & 9.00E+00                & 0.00E+00                & 0.00E+00 & 0.00E+00 & 0.00E+00 & 0.00E+00 & 0.00E+00 & 0.00E+00 & 0.00E+00             & 0.00E+00 & 0.00E+00             & 0.00E+00             & 0.00E+00 & 0.00E+00             & 0.00E+00             & 0.00E+00 & 0.00E+00             \\
                     & 15                   & 0.00E+00                & 7.06E+00                & 9.00E+00                & 2.40E-05                & 0.00E+00 & 0.00E+00 & 0.00E+00 & 0.00E+00 & 0.00E+00 & 0.00E+00 & 0.00E+00             & 0.00E+00 & 0.00E+00             & 0.00E+00             & 0.00E+00 & 0.00E+00             & 0.00E+00             & 0.00E+00 & 0.00E+00             \\
                     & 20                   & 0.00E+00                & 9.82E+00                & 9.00E+00                & 9.04E+00                & 0.00E+00 & 0.00E+00 & 0.00E+00 & 0.00E+00 & 0.00E+00 & 0.00E+00 & 0.00E+00             & 0.00E+00 & 0.00E+00             & 0.00E+00             & 0.00E+00 & 0.00E+00             & 0.00E+00             & 0.00E+00 & 0.00E+00             \\ \hline
\multirow{4}{*}{F14}                     & 5                    & 0.00E+00                & 0.00E+00                & 0.00E+00                & 0.00E+00                & 0.00E+00 & 0.00E+00 & 0.00E+00 & 0.00E+00 & 0.00E+00 & 0.00E+00 & 0.00E+00             & 0.00E+00 & 0.00E+00             & 0.00E+00             & 0.00E+00 & 0.00E+00             & 0.00E+00             & 0.00E+00 & 0.00E+00             \\
                     & 10                   & 0.00E+00                & 0.00E+00                & 0.00E+00                & 3.39E-03                & 0.00E+00 & 0.00E+00 & 0.00E+00 & 0.00E+00 & 0.00E+00 & 0.00E+00 & 0.00E+00             & 0.00E+00 & 0.00E+00             & 0.00E+00             & 0.00E+00 & 0.00E+00             & 0.00E+00             & 0.00E+00 & 0.00E+00             \\
                     & 15                   & 0.00E+00                & 0.00E+00                & 0.00E+00                & 2.65E-01                & 0.00E+00 & 0.00E+00 & 0.00E+00 & 0.00E+00 & 0.00E+00 & 0.00E+00 & 0.00E+00             & 0.00E+00 & 0.00E+00             & 0.00E+00             & 0.00E+00 & 0.00E+00             & 0.00E+00             & 0.00E+00 & 0.00E+00             \\
                     & 20                   & 0.00E+00                & 0.00E+00                & 0.00E+00                & 1.14E+01                & 0.00E+00 & 0.00E+00 & 0.00E+00 & 0.00E+00 & 0.00E+00 & 0.00E+00 & 0.00E+00             & 0.00E+00 & 0.00E+00             & 0.00E+00             & 0.00E+00 & 0.00E+00             & 0.00E+00             & 0.00E+00 & 0.00E+00             \\ \hline
\multirow{4}{*}{F15}                     & 5                    & 0.00E+00                & 0.00E+00                & 0.00E+00                & 0.00E+00                & 0.00E+00 & 0.00E+00 & 1.77E-05 & 0.00E+00 & 0.00E+00 & 0.00E+00 & 0.00E+00             & 0.00E+00 & 0.00E+00             & 0.00E+00             & 0.00E+00 & 0.00E+00             & 0.00E+00             & 0.00E+00 & 0.00E+00             \\
                     & 10                   & 0.00E+00                & 3.60E+01                & 0.00E+00                & 3.73E+00                & 0.00E+00 & 0.00E+00 & 0.00E+00 & 0.00E+00 & 0.00E+00 & 0.00E+00 & 0.00E+00             & 0.00E+00 & 0.00E+00             & 0.00E+00             & 0.00E+00 & 0.00E+00             & 0.00E+00             & 0.00E+00 & 0.00E+00             \\
                     & 15                   & 1.11E-06                & 1.40E+02                & 1.21E-08                & 1.65E+02                & 0.00E+00 & 0.00E+00 & 0.00E+00 & 0.00E+00 & 0.00E+00 & 0.00E+00 & 0.00E+00             & 0.00E+00 & 0.00E+00             & 0.00E+00             & 0.00E+00 & 0.00E+00             & 0.00E+00             & 0.00E+00 & 0.00E+00             \\
                     & 20                   & 4.38E-06                & 3.85E+02                & 1.14E+01                & 2.66E+02                & 0.00E+00 & 0.00E+00 & 0.00E+00 & 0.00E+00 & 0.00E+00 & 0.00E+00 & 0.00E+00             & 0.00E+00 & 0.00E+00             & 0.00E+00             & 0.00E+00 & 0.00E+00             & 0.00E+00             & 0.00E+00 & 0.00E+00             \\ \hline
\multirow{4}{*}{F16}                     & 5                    & 0.00E+00                & 0.00E+00                & 0.00E+00                & 0.00E+00                & 0.00E+00 & 0.00E+00 & 0.00E+00 & 0.00E+00 & 0.00E+00 & 0.00E+00 & 0.00E+00             & 0.00E+00 & 0.00E+00             & 0.00E+00             & 0.00E+00 & 0.00E+00             & 0.00E+00             & 0.00E+00 & 0.00E+00             \\
                     & 10                   & 6.75E-07                & 0.00E+00                & 1.33E-01                & 1.10E-03                & 0.00E+00 & 0.00E+00 & 0.00E+00 & 0.00E+00 & 0.00E+00 & 0.00E+00 & 0.00E+00             & 0.00E+00 & 0.00E+00             & 0.00E+00             & 7.95E-08 & 4.87E-06             & 0.00E+00             & 0.00E+00 & 0.00E+00             \\
                     & 15                   & 4.93E-02                & 0.00E+00                & 4.29E+02                & 6.60E+01                & 0.00E+00 & 0.00E+00 & 0.00E+00 & 0.00E+00 & 0.00E+00 & 0.00E+00 & 0.00E+00             & 0.00E+00 & 0.00E+00             & 2.21E-05             & 1.74E-01 & 4.89E+01             & 0.00E+00             & 0.00E+00 & 0.00E+00             \\
                     & 20                   & 1.26E+01                & 1.00E+01                & 1.40E+03                & 1.04E+02                & 1.00E+01 & 1.00E+01 & 1.00E+01 & 1.00E+01 & 1.00E+01 & 1.00E+01 & 8.57E+02             & 9.81E+02 & 1.12E+03             & 1.41E+01             & 2.20E+01 & 4.75E+02             & 1.00E+01             & 1.00E+01 & 1.00E+01             \\ \hline
\multirow{4}{*}{F17}                     & 5                    & 0.00E+00                & 0.00E+00                & 0.00E+00                & 0.00E+00                & 0.00E+00 & 6.75E-03 & 2.08E-02 & 0.00E+00 & 0.00E+00 & 0.00E+00 & 0.00E+00             & 0.00E+00 & 0.00E+00             & 0.00E+00             & 0.00E+00 & 0.00E+00             & 0.00E+00             & 0.00E+00 & 0.00E+00             \\
                     & 10                   & 0.00E+00                & 0.00E+00                & 0.00E+00                & 0.00E+00                & 0.00E+00 & 2.76E-02 & 5.34E-01 & 0.00E+00 & 0.00E+00 & 0.00E+00 & 0.00E+00             & 0.00E+00 & 0.00E+00             & 0.00E+00             & 0.00E+00 & 0.00E+00             & 0.00E+00             & 0.00E+00 & 0.00E+00             \\
                     & 15                   & 0.00E+00                & 0.00E+00                & 0.00E+00                & 2.50E-02                & 0.00E+00 & 0.00E+00 & 3.38E-05 & 0.00E+00 & 0.00E+00 & 0.00E+00 & 0.00E+00             & 0.00E+00 & 0.00E+00             & 0.00E+00             & 0.00E+00 & 0.00E+00             & 0.00E+00             & 0.00E+00 & 0.00E+00             \\
                     & 20                   & 0.00E+00                & 0.00E+00                & 0.00E+00                & 2.54E+00                & 0.00E+00 & 0.00E+00 & 0.00E+00 & 0.00E+00 & 0.00E+00 & 0.00E+00 & 0.00E+00             & 0.00E+00 & 0.00E+00             & 0.00E+00             & 0.00E+00 & 0.00E+00             & 0.00E+00             & 0.00E+00 & 0.00E+00             \\ \hline
\multirow{4}{*}{F18}                     & 5                    & 0.00E+00                & 0.00E+00                & 0.00E+00                & 2.17E-03                & 0.00E+00 & 0.00E+00 & 0.00E+00 & 0.00E+00 & 0.00E+00 & 0.00E+00 & 0.00E+00             & 0.00E+00 & 0.00E+00             & 0.00E+00             & 0.00E+00 & 0.00E+00             & 0.00E+00             & 0.00E+00 & 0.00E+00             \\
                     & 10                   & 9.12E-06                & 0.00E+00                & 1.42E-06                & 1.48E+00                & 0.00E+00 & 0.00E+00 & 0.00E+00 & 0.00E+00 & 0.00E+00 & 0.00E+00 & 0.00E+00             & 0.00E+00 & 0.00E+00             & 0.00E+00             & 0.00E+00 & 0.00E+00             & 0.00E+00             & 0.00E+00 & 0.00E+00             \\
                     & 15                   & 1.33E-02                & 0.00E+00                & 1.38E-05                & 4.54E+00                & 0.00E+00 & 0.00E+00 & 0.00E+00 & 0.00E+00 & 0.00E+00 & 0.00E+00 & 0.00E+00             & 0.00E+00 & 0.00E+00             & 0.00E+00             & 0.00E+00 & 0.00E+00             & 0.00E+00             & 0.00E+00 & 0.00E+00             \\
                     & 20                   & 4.85E-02                & 0.00E+00                & 5.46E-02                & 9.54E+00                & 0.00E+00 & 0.00E+00 & 0.00E+00 & 0.00E+00 & 0.00E+00 & 0.00E+00 & 0.00E+00             & 0.00E+00 & 0.00E+00             & 0.00E+00             & 0.00E+00 & 0.00E+00             & 0.00E+00             & 0.00E+00 & 0.00E+00            
\end{tabular} \label{dirlib_res2}
\egroup
}
\end{table}
\end{landscape}

% \begin{landscape}
% \begin{table}[!h]
% \caption{Detailed statistics of the 30 runs of the selected algorithms on the DIRECTLib benchmark set, F10-F18.}
% \centering
% %\scriptsize	
% \includegraphics[width=\linewidth]{tab3.png}
% \label{dirlib_res2}
% \end{table}
% \end{landscape}

\subsection{BBOB Results}
\begin{figure}[!h]
    \centering
    \begin{tabular}{cc}
        \includegraphics[width = 0.4\linewidth]{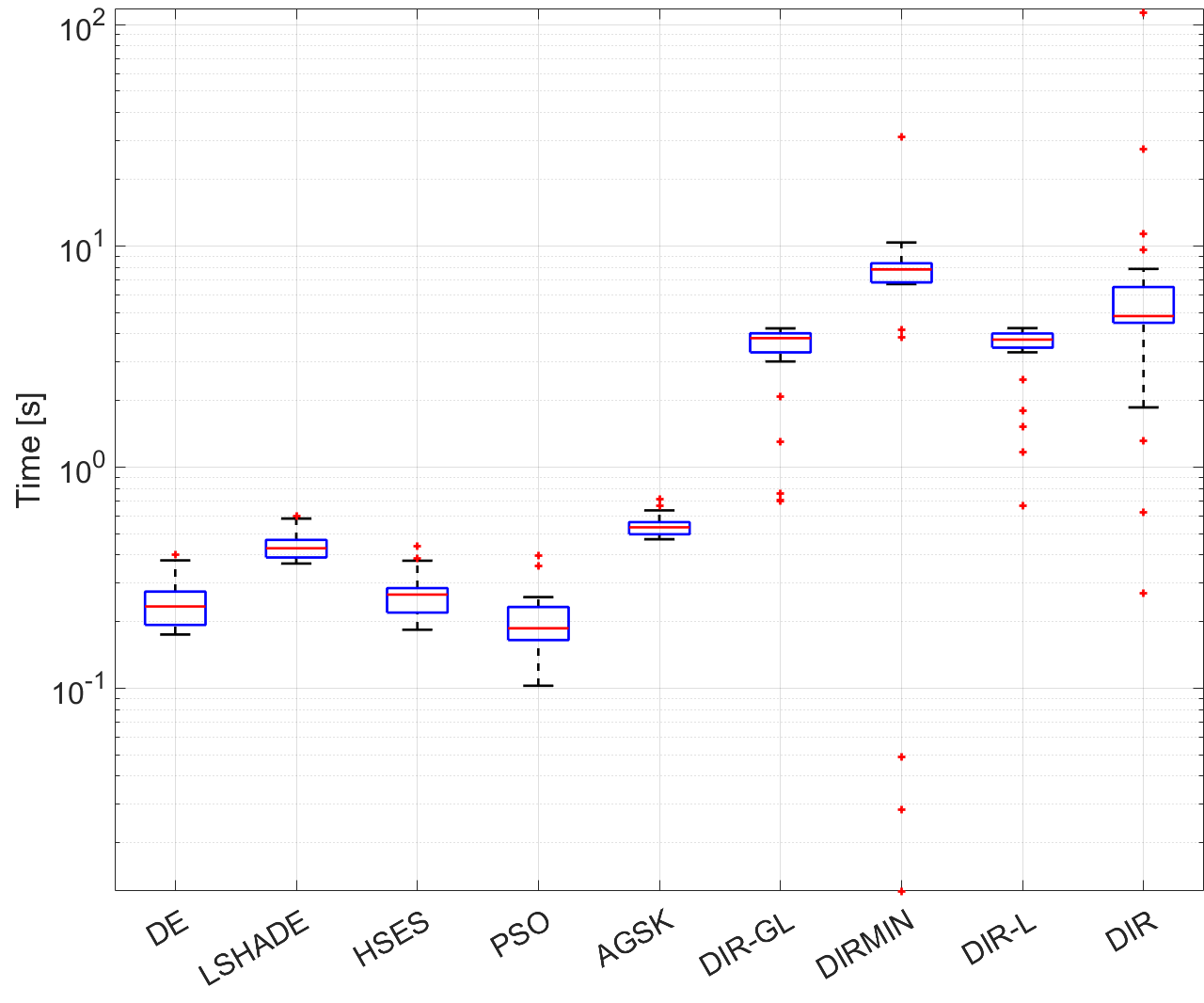} & \includegraphics[width = 0.4\linewidth]{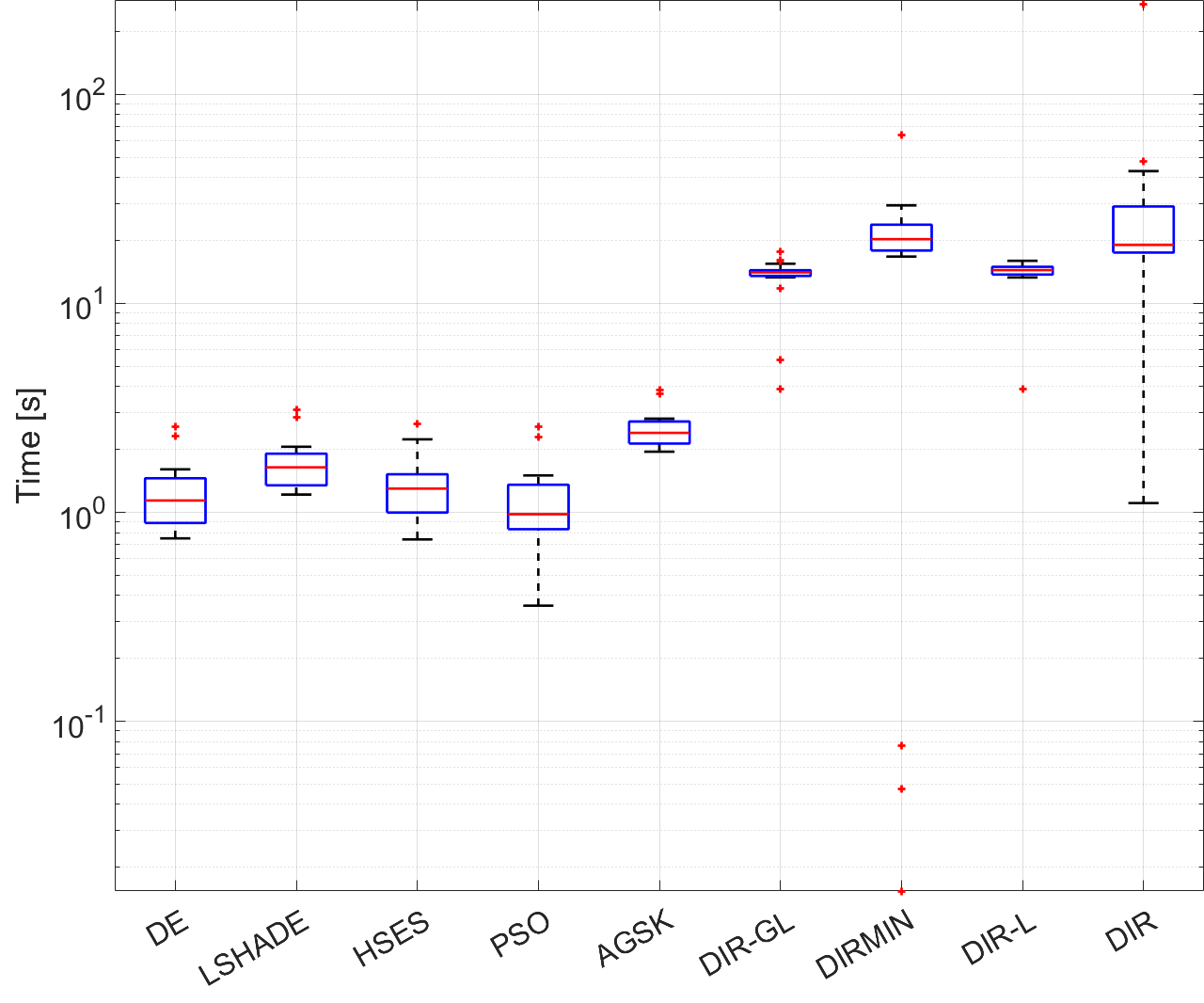} \\
        a) & b) \vspace{3mm}\\
        \multicolumn{2}{c}{\includegraphics[width = 0.4\linewidth]{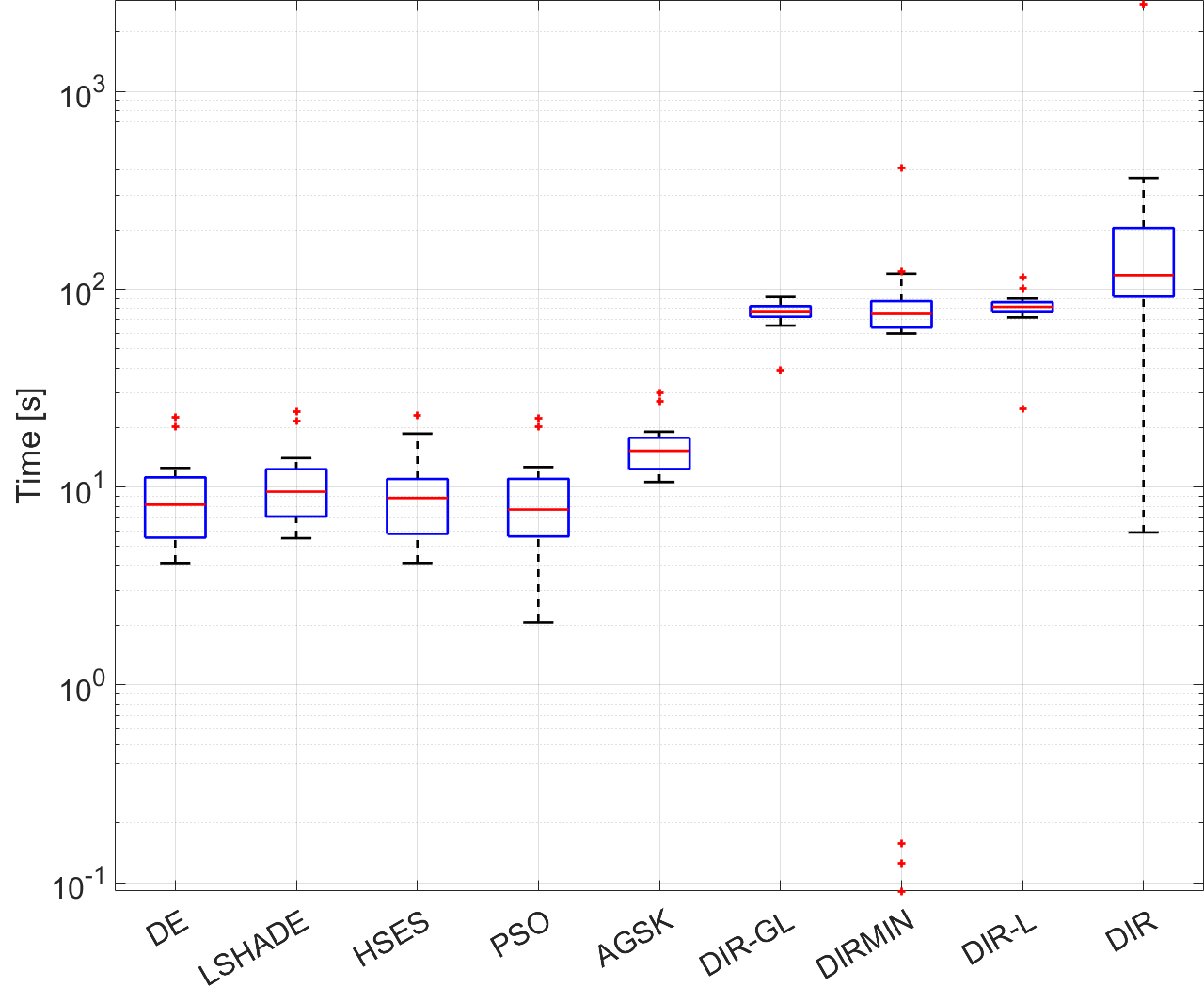}} \\
        \multicolumn{2}{c}{c)}
    \end{tabular}
    \caption{Time complexity of the different algorithms on the BBOB benchmark set, a) $D=5$, b) $D=10$, c) $D=20$.} \label{fig_bbob_time}
\end{figure}

Now, we take a close look at the results on the BBOB benchmark set. In Figure \ref{fig_bbob_time}, we can again observe that the time complexity of the DIRECT-type methods is roughly an order of magnitude higher than that of their nature-inspired counterparts. Compared to the results of the DIRECTLib set, the variability in is much lower as the functions in the BBOB set are much more challenging.

\begin{figure}[!h]
    \centering
    \begin{tabular}{cc}
        \includegraphics[width = 0.45\linewidth]{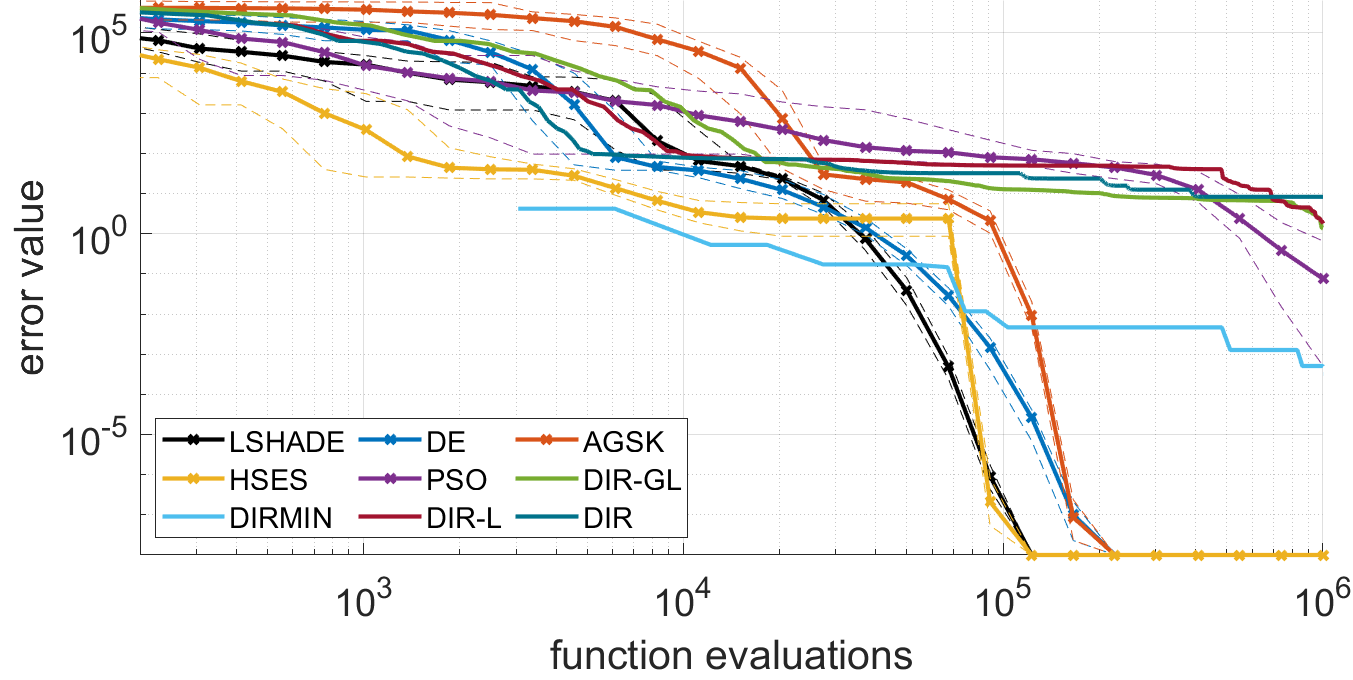} & \includegraphics[width = 0.45\linewidth]{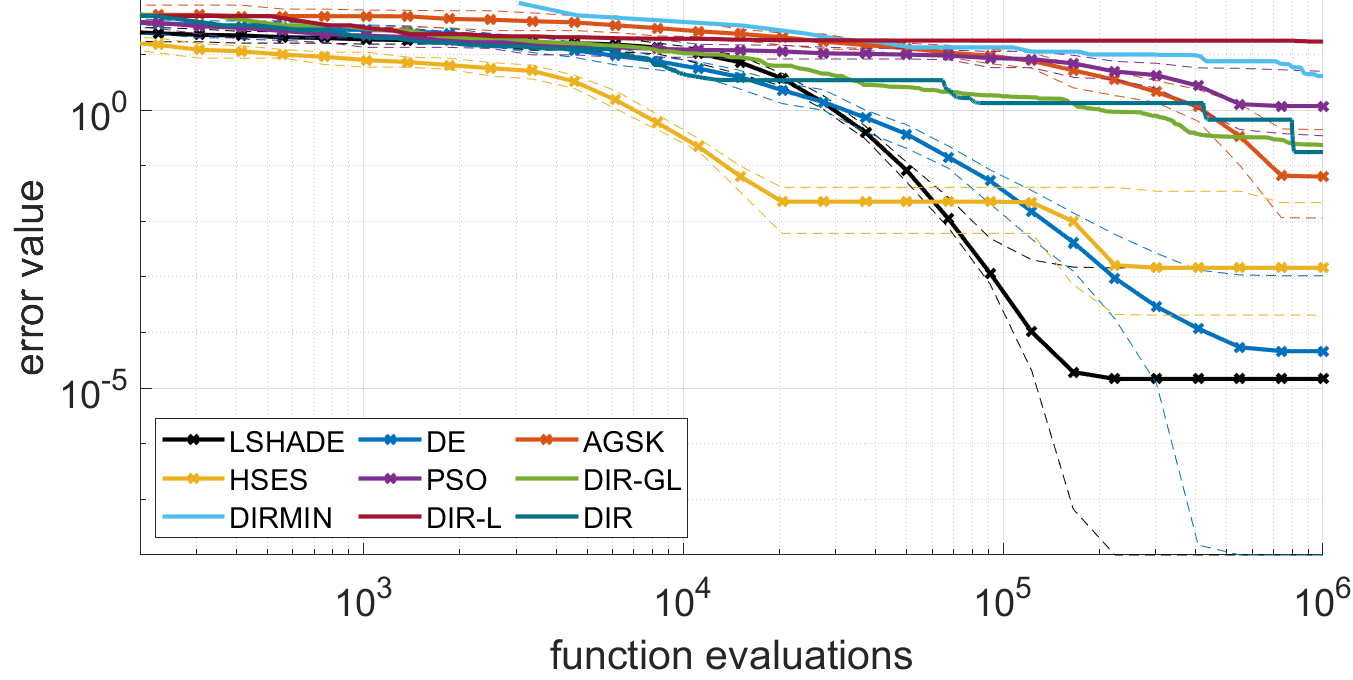} \\
        a) & b)
    \end{tabular}
    \caption{Convergence plots from the BBOB benchmark set, a) F6 in $D=20$, b) F18 in $D=20$.} \label{conv_bbob}
\end{figure}

Detailed statistics of the results of the computations are reported in Tables \ref{bbob_res1} and \ref{bbob_res2}, and a typical convergence plot is shown in Figure \ref{conv_bbob}. When comparing the best (``min'') results, we find that LSHADE, AGSK, HSES, and DE all dominate the DIRECT-type methods. However, there were two instances (F24 in $D=[5,10]$), where the best result was obtained by DIR-GL. When looking at the ``median'' results, although LSHADE, AGSK, HSES, and DE are still be overall best methods, DIR-GL found best solution in 6 instances and was tied for the best in 16 instances. And, finally, when comparing against the worst (``max'') results, only LSHADE was consistently better than the best DIRECT-type methods (DIR-GL and DIRMIN).

\begin{figure}[!h]
    \centering
    \begin{tabular}{cc}
        \includegraphics[width = 0.4\linewidth]{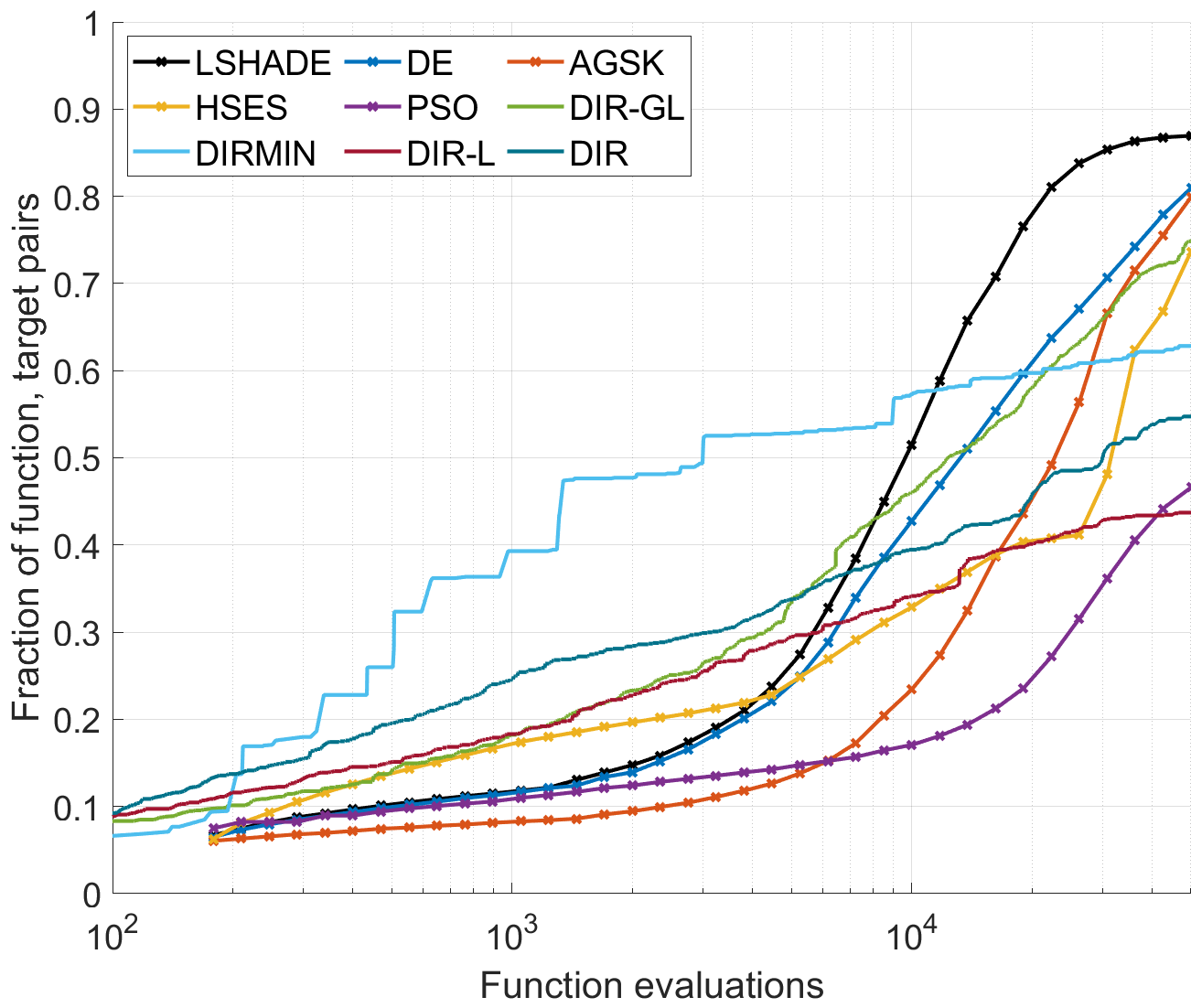} & \includegraphics[width = 0.4\linewidth]{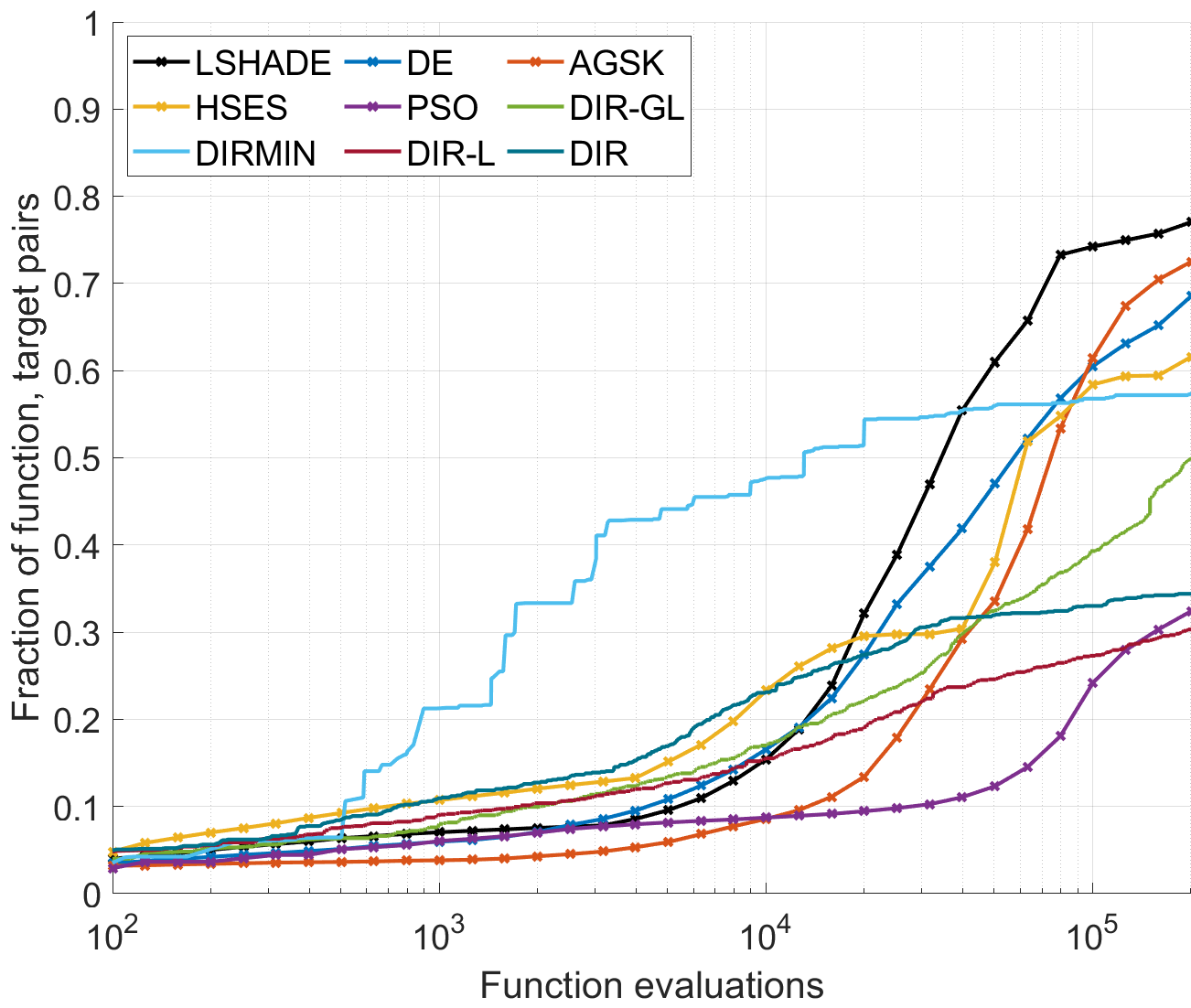} \\
        a) & b) \vspace{3mm}\\
        \multicolumn{2}{c}{\includegraphics[width = 0.4\linewidth]{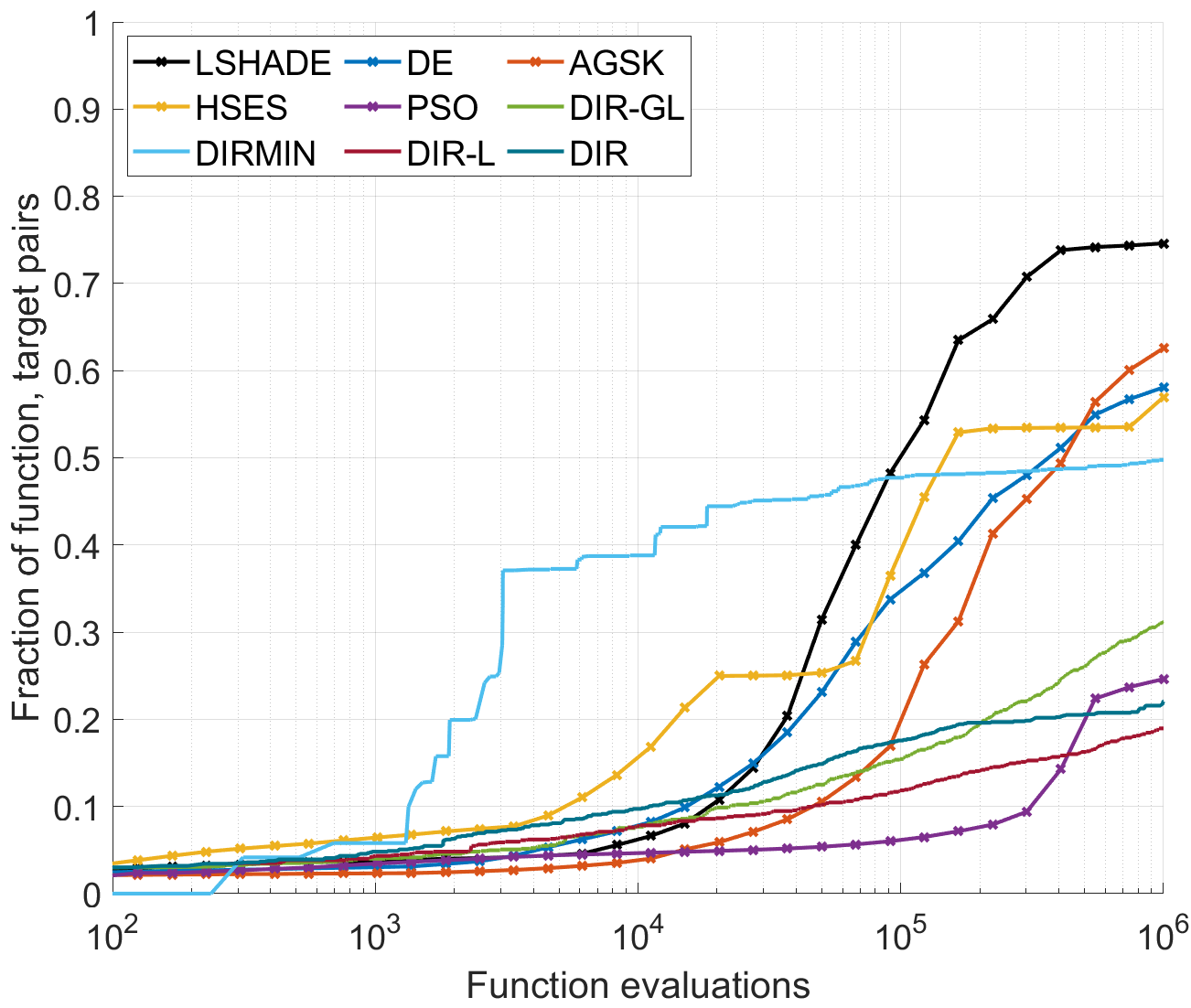}} \\
        \multicolumn{2}{c}{c)}
    \end{tabular}
     \caption{ECD of the number of objective function evaluations for different target precisions on the BBOB benchmark set, a) $D=5$, b) $D=10$, c) $D=20$.} \label{fig_bbob_ecd}
\end{figure}

The rank tests (reported in Table \ref{bbob_ranks}) and ECD plots (shown in Figure \ref{fig_bbob_ecd}) reveal that when the number of function evaluations is small, the DIRECT-type methods (in particular DIR, DIRMIN, and DIR-L) have superior performance (especially in lower dimensions). As the iterations progress, DIR-GL and LSHADE start to take over. And after reaching the maximum number of available function evaluations, the top performing methods on this set were the complex nature-inspired ones: LSHADE, AGSK, and HSES.

\begin{table}[!h]
\caption{Mean ranks from Friedman tests at different stages of the search, BBOB benchmark set.}
\centering
\resizebox{0.6\columnwidth}{!}{%
\setlength{\tabcolsep}{0.3em}
\bgroup
\def\arraystretch{1.1}
\begin{tabular}{l|ccccccccc}
FES            & \multicolumn{1}{c}{DE} & \multicolumn{1}{c}{LSHADE} & \multicolumn{1}{c}{HSES} & \multicolumn{1}{c}{PSO} &\multicolumn{1}{c}{AGSK} & \multicolumn{1}{c}{DIR-GL} & \multicolumn{1}{c}{DIRMIN} & \multicolumn{1}{c}{DIR-L} & \multicolumn{1}{c}{DIR} \\ \hline 
(1/256)*MaxFES & 6.5 & 6.0 & 4.0 & 6.3 & 8.0 & 5.0 & \textbf{2.9} & \textbf{3.8} & \textbf{2.6} \\
(1/128)*MaxFES & 6.2 & 6.3 & \textbf{3.3} & 6.8 & 8.3 & 4.5 & \textbf{3.2} & 3.9 & \textbf{2.5} \\
(1/32)*MaxFES  & 5.4 & 5.6 & \textbf{3.6} & 7.7 & 8.0 & 3.9 & \textbf{3.1} & 4.5 & \textbf{3.2} \\
(1/8)*MaxFES   & 4.8 & \textbf{3.6} & 4.1 & 8.1 & 6.5 & \textbf{3.8} & \textbf{3.6} & 5.8 & 4.6 \\
(1/4)*MaxFES   & 4.5 & \textbf{2.7} & 4.5 & 8.2 & 5.2 & \textbf{4.1} & \textbf{4.1} & 6.5 & 5.3 \\
(1/2)*MaxFES   & \textbf{4.1} & \textbf{2.4} & \textbf{4.3} & 7.5 & 4.5 & 4.4 & 4.8 & 7.1 & 5.9 \\
(3/4)*MaxFES   & \textbf{4.0} & \textbf{2.5} & 4.1 & 6.5 & \textbf{3.8} & 4.8 & 5.4 & 7.3 & 6.5 \\
MaxFES         & 4.2 & \textbf{3.1} & \textbf{3.9} & 6.5 & \textbf{3.8} & 4.7 & 5.3 & 7.1 & 6.4
\end{tabular} \label{bbob_ranks}
\egroup
}
\end{table}

\begin{landscape}
\begin{table}[!h]
\caption{Detailed statistics of the 30 runs of the selected algorithms on the BBOB benchmark set, F1-12.}
\centering
%\scriptsize	
\resizebox{1.00\columnwidth}{!}{%
\setlength{\tabcolsep}{0.3em}
\bgroup
\def\arraystretch{1.3}
\begin{tabular}{cc|cccc|ccc|ccc|ccc|ccc|ccc}
\multicolumn{2}{l}{}& \multicolumn{4}{c}{Direct methods} & \multicolumn{3}{c}{DE} & \multicolumn{3}{c}{LSHADE} & \multicolumn{3}{c}{HSES} & \multicolumn{3}{c}{PSO} & \multicolumn{3}{c}{AGSK} \\
\multicolumn{1}{l}{} & D                    & DIR-GL               & DIRMIN               & DIR-L                & DIR                  & min                  & median   & max                            & min                  & median   & max                                  & min                  & median   & max                               & min                  & median   & max     & min                  & median   & max                               \\ \hline
\multirow{3}{*}{F1}  & 5  & 0.00E+00 & 0.00E+00 & 0.00E+00 & 0.00E+00 & 0.00E+00 & 0.00E+00 & 0.00E+00 & 0.00E+00 & 0.00E+00 & 0.00E+00 & 0.00E+00 & 0.00E+00 & 0.00E+00 & 0.00E+00 & 0.00E+00 & 0.00E+00 & 0.00E+00 & 0.00E+00 & 0.00E+00 \\
  & 10 & 0.00E+00 & 0.00E+00 & 0.00E+00 & 0.00E+00 & 0.00E+00 & 0.00E+00 & 0.00E+00 & 0.00E+00 & 0.00E+00 & 0.00E+00 & 0.00E+00 & 0.00E+00 & 0.00E+00 & 0.00E+00 & 0.00E+00 & 0.00E+00 & 0.00E+00 & 0.00E+00 & 0.00E+00 \\
  & 20 & 0.00E+00 & 0.00E+00 & 0.00E+00 & 0.00E+00 & 0.00E+00 & 0.00E+00 & 0.00E+00 & 0.00E+00 & 0.00E+00 & 0.00E+00 & 0.00E+00 & 0.00E+00 & 0.00E+00 & 0.00E+00 & 0.00E+00 & 0.00E+00 & 0.00E+00 & 0.00E+00 & 0.00E+00 \\ \hline
\multirow{3}{*}{F2}  & 5  & 1.45E-08 & 0.00E+00 & 0.00E+00 & 2.79E-03 & 0.00E+00 & 0.00E+00 & 0.00E+00 & 0.00E+00 & 0.00E+00 & 0.00E+00 & 0.00E+00 & 0.00E+00 & 0.00E+00 & 0.00E+00 & 0.00E+00 & 0.00E+00 & 0.00E+00 & 0.00E+00 & 0.00E+00 \\
  & 10 & 7.68E-08 & 0.00E+00 & 6.74E-08 & 1.30E-06 & 0.00E+00 & 0.00E+00 & 0.00E+00 & 0.00E+00 & 0.00E+00 & 0.00E+00 & 0.00E+00 & 0.00E+00 & 0.00E+00 & 0.00E+00 & 0.00E+00 & 0.00E+00 & 0.00E+00 & 0.00E+00 & 0.00E+00 \\
  & 20 & 2.86E-05 & 0.00E+00 & 3.11E+00 & 4.05E+01 & 0.00E+00 & 0.00E+00 & 0.00E+00 & 0.00E+00 & 0.00E+00 & 0.00E+00 & 0.00E+00 & 0.00E+00 & 0.00E+00 & 0.00E+00 & 0.00E+00 & 0.00E+00 & 0.00E+00 & 0.00E+00 & 0.00E+00 \\ \hline
\multirow{3}{*}{F3}  & 5  & 0.00E+00 & 9.95E-01 & 2.98E+00 & 0.00E+00 & 0.00E+00 & 0.00E+00 & 1.97E-01 & 0.00E+00 & 0.00E+00 & 0.00E+00 & 0.00E+00 & 9.95E-01 & 2.98E+00 & 0.00E+00 & 0.00E+00 & 9.95E-01 & 0.00E+00 & 0.00E+00 & 0.00E+00 \\
  & 10 & 9.95E-01 & 1.89E+01 & 1.69E+01 & 2.98E+00 & 0.00E+00 & 4.17E+00 & 1.93E+01 & 0.00E+00 & 0.00E+00 & 0.00E+00 & 9.95E-01 & 2.98E+00 & 5.97E+00 & 0.00E+00 & 9.95E-01 & 3.98E+00 & 0.00E+00 & 0.00E+00 & 0.00E+00 \\
  & 20 & 1.39E+01 & 5.87E+01 & 4.08E+01 & 4.08E+01 & 0.00E+00 & 2.98E+00 & 5.97E+00 & 0.00E+00 & 0.00E+00 & 0.00E+00 & 2.98E+00 & 6.96E+00 & 1.49E+01 & 9.95E-01 & 4.97E+00 & 9.95E+00 & 0.00E+00 & 0.00E+00 & 0.00E+00 \\ \hline
\multirow{3}{*}{F4}  & 5  & 0.00E+00 & 7.96E+00 & 1.39E+01 & 1.52E+00 & 0.00E+00 & 5.34E-09 & 9.95E-01 & 0.00E+00 & 0.00E+00 & 0.00E+00 & 1.99E+00 & 3.98E+00 & 5.97E+00 & 0.00E+00 & 9.95E-01 & 1.99E+00 & 0.00E+00 & 0.00E+00 & 0.00E+00 \\
  & 10 & 1.99E+00 & 1.99E+01 & 3.68E+01 & 2.79E+01 & 9.95E-01 & 3.62E+00 & 1.29E+01 & 0.00E+00 & 0.00E+00 & 0.00E+00 & 4.97E+00 & 8.95E+00 & 1.09E+01 & 0.00E+00 & 2.49E+00 & 6.96E+00 & 0.00E+00 & 0.00E+00 & 0.00E+00 \\
  & 20 & 2.80E+01 & 1.50E+02 & 5.57E+01 & 6.27E+01 & 3.98E+00 & 7.96E+00 & 1.29E+01 & 0.00E+00 & 0.00E+00 & 0.00E+00 & 1.09E+01 & 1.69E+01 & 2.29E+01 & 3.98E+00 & 7.96E+00 & 1.29E+01 & 0.00E+00 & 0.00E+00 & 0.00E+00 \\ \hline
\multirow{3}{*}{F5}  & 5  & 8.80E-08 & 1.00E-07 & 0.00E+00 & 0.00E+00 & 0.00E+00 & 0.00E+00 & 0.00E+00 & 0.00E+00 & 0.00E+00 & 0.00E+00 & 0.00E+00 & 0.00E+00 & 1.24E-08 & 7.63E-01 & 2.09E+00 & 2.76E+00 & 0.00E+00 & 0.00E+00 & 0.00E+00 \\
  & 10 & 3.86E-05 & 3.98E-06 & 7.00E-08 & 0.00E+00 & 0.00E+00 & 0.00E+00 & 0.00E+00 & 0.00E+00 & 0.00E+00 & 0.00E+00 & 1.31E-04 & 3.24E-04 & 6.07E-04 & 6.50E+00 & 8.80E+00 & 1.08E+01 & 0.00E+00 & 0.00E+00 & 0.00E+00 \\
  & 20 & 1.68E-03 & 7.96E-06 & 1.08E-05 & 0.00E+00 & 0.00E+00 & 0.00E+00 & 0.00E+00 & 0.00E+00 & 0.00E+00 & 0.00E+00 & 8.74E+00 & 3.89E+01 & 8.24E+01 & 2.26E+01 & 2.58E+01 & 4.03E+01 & 0.00E+00 & 0.00E+00 & 0.00E+00 \\ \hline
\multirow{3}{*}{F6}  & 5  & 0.00E+00 & 0.00E+00 & 0.00E+00 & 2.05E-01 & 0.00E+00 & 0.00E+00 & 0.00E+00 & 0.00E+00 & 0.00E+00 & 0.00E+00 & 0.00E+00 & 0.00E+00 & 0.00E+00 & 0.00E+00 & 0.00E+00 & 2.80E-08 & 0.00E+00 & 0.00E+00 & 0.00E+00 \\
  & 10 & 2.75E-04 & 6.86E-08 & 7.85E-05 & 3.67E+00 & 0.00E+00 & 0.00E+00 & 0.00E+00 & 0.00E+00 & 0.00E+00 & 0.00E+00 & 0.00E+00 & 0.00E+00 & 0.00E+00 & 0.00E+00 & 5.91E-08 & 7.79E-07 & 0.00E+00 & 0.00E+00 & 0.00E+00 \\
  & 20 & 1.37E+00 & 3.22E-04 & 1.78E+00 & 8.31E+00 & 0.00E+00 & 0.00E+00 & 0.00E+00 & 0.00E+00 & 0.00E+00 & 0.00E+00 & 0.00E+00 & 0.00E+00 & 0.00E+00 & 5.11E-04 & 7.52E-02 & 6.59E-01 & 0.00E+00 & 0.00E+00 & 0.00E+00 \\ \hline
\multirow{3}{*}{F7}  & 5  & 0.00E+00 & 3.18E-01 & 0.00E+00 & 3.18E-01 & 0.00E+00 & 0.00E+00 & 0.00E+00 & 0.00E+00 & 0.00E+00 & 0.00E+00 & 3.00E-01 & 4.93E-01 & 1.83E+00 & 0.00E+00 & 0.00E+00 & 4.58E-01 & 0.00E+00 & 0.00E+00 & 0.00E+00 \\
  & 10 & 0.00E+00 & 3.55E-01 & 1.61E-01 & 3.55E-01 & 0.00E+00 & 0.00E+00 & 0.00E+00 & 0.00E+00 & 0.00E+00 & 0.00E+00 & 4.98E-01 & 1.59E+00 & 2.30E+00 & 0.00E+00 & 3.39E-01 & 8.46E-01 & 0.00E+00 & 0.00E+00 & 0.00E+00 \\
  & 20 & 3.88E-02 & 6.05E+00 & 2.83E+00 & 6.05E+00 & 0.00E+00 & 0.00E+00 & 0.00E+00 & 0.00E+00 & 0.00E+00 & 0.00E+00 & 4.10E-01 & 1.48E+00 & 3.05E+00 & 5.11E-02 & 2.00E+00 & 5.00E+00 & 0.00E+00 & 0.00E+00 & 3.50E-02 \\ \hline
\multirow{3}{*}{F8}  & 5  & 0.00E+00 & 0.00E+00 & 3.51E-02 & 0.00E+00 & 0.00E+00 & 0.00E+00 & 0.00E+00 & 0.00E+00 & 0.00E+00 & 0.00E+00 & 0.00E+00 & 0.00E+00 & 0.00E+00 & 3.34E-02 & 2.51E-01 & 1.11E+00 & 0.00E+00 & 0.00E+00 & 0.00E+00 \\
  & 10 & 3.35E-06 & 0.00E+00 & 1.10E-02 & 6.03E+00 & 0.00E+00 & 0.00E+00 & 0.00E+00 & 0.00E+00 & 0.00E+00 & 0.00E+00 & 0.00E+00 & 0.00E+00 & 0.00E+00 & 1.14E-02 & 2.37E+00 & 3.46E+00 & 0.00E+00 & 0.00E+00 & 0.00E+00 \\
  & 20 & 1.28E+01 & 0.00E+00 & 4.16E+01 & 1.86E+01 & 0.00E+00 & 0.00E+00 & 0.00E+00 & 0.00E+00 & 0.00E+00 & 0.00E+00 & 0.00E+00 & 0.00E+00 & 0.00E+00 & 8.82E-01 & 9.93E+00 & 1.18E+01 & 0.00E+00 & 0.00E+00 & 0.00E+00 \\ \hline
\multirow{3}{*}{F9}  & 5  & 2.05E-08 & 0.00E+00 & 4.29E-02 & 0.00E+00 & 0.00E+00 & 0.00E+00 & 1.75E-01 & 0.00E+00 & 0.00E+00 & 0.00E+00 & 0.00E+00 & 0.00E+00 & 0.00E+00 & 1.08E-01 & 1.77E-01 & 6.55E+00 & 0.00E+00 & 0.00E+00 & 0.00E+00 \\
  & 10 & 2.33E-02 & 0.00E+00 & 5.66E+00 & 3.01E+00 & 0.00E+00 & 0.00E+00 & 0.00E+00 & 0.00E+00 & 0.00E+00 & 0.00E+00 & 0.00E+00 & 0.00E+00 & 0.00E+00 & 2.22E-01 & 5.02E+00 & 1.11E+02 & 0.00E+00 & 0.00E+00 & 0.00E+00 \\
  & 20 & 7.22E+00 & 0.00E+00 & 1.13E+01 & 1.88E+01 & 0.00E+00 & 0.00E+00 & 0.00E+00 & 0.00E+00 & 0.00E+00 & 0.00E+00 & 0.00E+00 & 0.00E+00 & 0.00E+00 & 7.45E+00 & 1.61E+01 & 1.69E+02 & 0.00E+00 & 0.00E+00 & 0.00E+00 \\ \hline
\multirow{3}{*}{F10} & 5  & 5.80E-01 & 0.00E+00 & 7.81E+00 & 1.47E+00 & 0.00E+00 & 0.00E+00 & 0.00E+00 & 0.00E+00 & 0.00E+00 & 0.00E+00 & 0.00E+00 & 0.00E+00 & 0.00E+00 & 1.13E+00 & 1.15E+01 & 2.42E+02 & 0.00E+00 & 0.00E+00 & 0.00E+00 \\
 & 10 & 4.34E+01 & 4.98E-08 & 2.58E+02 & 1.18E+02 & 0.00E+00 & 0.00E+00 & 0.00E+00 & 0.00E+00 & 0.00E+00 & 0.00E+00 & 0.00E+00 & 0.00E+00 & 0.00E+00 & 3.58E+01 & 1.19E+03 & 4.19E+03 & 0.00E+00 & 0.00E+00 & 3.37E-05 \\
 & 20 & 2.51E+02 & 1.68E-07 & 1.42E+03 & 4.92E+03 & 3.78E-03 & 1.52E+00 & 2.81E+01 & 0.00E+00 & 0.00E+00 & 0.00E+00 & 0.00E+00 & 0.00E+00 & 0.00E+00 & 3.43E+02 & 2.73E+03 & 1.94E+04 & 7.63E-02 & 1.17E+00 & 1.47E+01 \\ \hline
\multirow{3}{*}{F11} & 5  & 3.57E-03 & 1.11E-08 & 2.32E-01 & 5.56E-01 & 0.00E+00 & 0.00E+00 & 0.00E+00 & 0.00E+00 & 0.00E+00 & 0.00E+00 & 0.00E+00 & 0.00E+00 & 0.00E+00 & 4.23E-02 & 6.89E-01 & 3.12E+00 & 0.00E+00 & 0.00E+00 & 0.00E+00 \\
 & 10 & 1.89E-01 & 2.55E-06 & 4.88E-02 & 1.60E+01 & 0.00E+00 & 0.00E+00 & 0.00E+00 & 0.00E+00 & 0.00E+00 & 0.00E+00 & 0.00E+00 & 0.00E+00 & 0.00E+00 & 3.28E-01 & 1.10E+00 & 2.72E+00 & 0.00E+00 & 0.00E+00 & 0.00E+00 \\
 & 20 & 1.84E-01 & 3.68E-07 & 2.52E+00 & 4.89E+01 & 0.00E+00 & 0.00E+00 & 0.00E+00 & 0.00E+00 & 0.00E+00 & 0.00E+00 & 0.00E+00 & 0.00E+00 & 0.00E+00 & 2.12E-01 & 5.76E-01 & 1.01E+00 & 0.00E+00 & 3.36E-08 & 2.60E-06 \\ \hline
\multirow{3}{*}{F12} & 5  & 4.37E-03 & 0.00E+00 & 1.73E+00 & 4.59E-04 & 0.00E+00 & 2.08E-06 & 4.75E-01 & 0.00E+00 & 0.00E+00 & 0.00E+00 & 0.00E+00 & 0.00E+00 & 0.00E+00 & 8.53E-05 & 7.67E+00 & 3.60E+01 & 0.00E+00 & 9.52E-07 & 6.85E-03 \\
 & 10 & 2.85E-04 & 0.00E+00 & 1.11E+00 & 2.99E-04 & 1.53E-08 & 1.20E-05 & 3.61E-04 & 0.00E+00 & 0.00E+00 & 0.00E+00 & 0.00E+00 & 0.00E+00 & 0.00E+00 & 1.62E-02 & 1.49E+00 & 5.11E+01 & 8.22E-07 & 1.44E-04 & 3.47E-02 \\
 & 20 & 1.87E-01 & 0.00E+00 & 3.09E+00 & 1.24E+03 & 3.14E-08 & 5.42E-06 & 4.21E-04 & 0.00E+00 & 0.00E+00 & 0.00E+00 & 0.00E+00 & 0.00E+00 & 0.00E+00 & 2.11E-03 & 1.66E+00 & 3.15E+01 & 9.81E-06 & 3.54E-04 & 5.48E-03
\end{tabular} \label{bbob_res1}
\egroup
}
\end{table}
\end{landscape}

% \begin{landscape}
% \begin{table}[!h]
% \caption{Detailed statistics of the 30 runs of the selected algorithms on the BBOB benchmark set, F1-12.}
% \centering
% %\scriptsize	
% \includegraphics[width = \linewidth]{tab4.png}
% \label{bbob_res1}
% \end{table}
% \end{landscape}

\begin{landscape}
\begin{table}[!h]
\caption{Detailed statistics of the 30 runs of the selected algorithms on the BBOB benchmark set, F13-24.}
\centering
%\scriptsize	
\resizebox{1.00\columnwidth}{!}{%
\setlength{\tabcolsep}{0.3em}
\bgroup
\def\arraystretch{1.3}
\begin{tabular}{cc|cccc|ccc|ccc|ccc|ccc|ccc}
\multicolumn{2}{l}{}& \multicolumn{4}{c}{Direct methods} & \multicolumn{3}{c}{DE} & \multicolumn{3}{c}{LSHADE} & \multicolumn{3}{c}{HSES} & \multicolumn{3}{c}{PSO} & \multicolumn{3}{c}{AGSK} \\
\multicolumn{1}{l}{} & D                    & DIR-GL               & DIRMIN               & DIR-L                & DIR                  & min                  & median   & max                            & min                  & median   & max                                  & min                  & median   & max                               & min                  & median   & max     & min                  & median   & max                               \\ \hline
\multirow{3}{*}{F13} & 5  & 5.20E-05 & 2.12E-05 & 3.75E+00 & 1.81E-04 & 0.00E+00 & 0.00E+00 & 0.00E+00 & 0.00E+00 & 0.00E+00 & 0.00E+00 & 0.00E+00 & 0.00E+00 & 0.00E+00 & 2.51E-03 & 2.35E+00 & 1.44E+01 & 0.00E+00 & 0.00E+00 & 0.00E+00 \\
 & 10 & 2.23E-04 & 1.19E-05 & 2.76E+01 & 1.49E-02 & 0.00E+00 & 0.00E+00 & 0.00E+00 & 0.00E+00 & 0.00E+00 & 0.00E+00 & 0.00E+00 & 0.00E+00 & 0.00E+00 & 8.08E-04 & 2.25E+00 & 2.72E+01 & 2.69E-08 & 4.65E-06 & 3.77E-05 \\
 & 20 & 3.27E+00 & 1.93E-05 & 1.10E+01 & 5.10E+01 & 1.22E-07 & 1.30E-06 & 5.15E-06 & 0.00E+00 & 0.00E+00 & 0.00E+00 & 0.00E+00 & 0.00E+00 & 1.65E-03 & 8.23E-03 & 3.93E+00 & 3.34E+01 & 4.22E-07 & 2.36E-05 & 5.28E-04 \\ \hline
\multirow{3}{*}{F14} & 5  & 8.57E-06 & 5.51E-08 & 2.90E-05 & 1.43E-05 & 0.00E+00 & 0.00E+00 & 0.00E+00 & 0.00E+00 & 0.00E+00 & 0.00E+00 & 0.00E+00 & 0.00E+00 & 0.00E+00 & 1.31E-05 & 3.65E-05 & 7.90E-05 & 0.00E+00 & 0.00E+00 & 0.00E+00 \\
 & 10 & 8.54E-05 & 9.45E-08 & 1.50E-04 & 6.25E-04 & 0.00E+00 & 0.00E+00 & 0.00E+00 & 0.00E+00 & 0.00E+00 & 0.00E+00 & 0.00E+00 & 0.00E+00 & 0.00E+00 & 7.69E-06 & 7.51E-05 & 1.31E-04 & 0.00E+00 & 0.00E+00 & 3.87E-08 \\
 & 20 & 5.16E-04 & 1.84E-06 & 5.15E-03 & 9.54E-03 & 1.67E-07 & 3.71E-06 & 1.10E-05 & 0.00E+00 & 0.00E+00 & 0.00E+00 & 0.00E+00 & 0.00E+00 & 0.00E+00 & 1.95E-04 & 2.42E-04 & 3.54E-04 & 3.69E-07 & 1.52E-06 & 3.03E-06 \\ \hline
\multirow{3}{*}{F15} & 5  & 0.00E+00 & 1.99E+00 & 1.99E+00 & 1.99E+00 & 0.00E+00 & 2.17E+00 & 3.86E+00 & 0.00E+00 & 9.95E-01 & 9.97E-01 & 0.00E+00 & 0.00E+00 & 9.95E-01 & 0.00E+00 & 1.99E+00 & 3.98E+00 & 2.91E-02 & 1.02E+00 & 3.10E+00 \\
 & 10 & 2.98E+00 & 5.97E+00 & 8.26E+01 & 1.56E+01 & 1.99E+00 & 1.90E+01 & 2.78E+01 & 6.45E-04 & 1.99E+00 & 2.99E+00 & 1.99E+00 & 3.48E+00 & 5.97E+00 & 2.98E+00 & 1.14E+01 & 2.69E+01 & 3.98E+00 & 7.26E+00 & 1.69E+01 \\
 & 20 & 1.39E+01 & 3.28E+01 & 1.58E+02 & 4.93E+01 & 5.97E+00 & 1.88E+01 & 9.25E+01 & 9.43E-04 & 3.49E+00 & 4.98E+00 & 1.99E+00 & 3.98E+00 & 7.96E+00 & 9.95E+00 & 2.84E+01 & 7.55E+01 & 1.51E+01 & 4.13E+01 & 6.73E+01 \\ \hline
\multirow{3}{*}{F16} & 5  & 0.00E+00 & 1.64E+00 & 1.50E-02 & 2.73E-05 & 4.59E-03 & 7.16E-01 & 2.40E+00 & 1.54E-03 & 6.20E-02 & 2.83E-01 & 0.00E+00 & 7.62E-02 & 1.09E+00 & 2.42E-07 & 6.45E-03 & 5.55E-01 & 3.21E-02 & 1.81E-01 & 7.64E-01 \\
 & 10 & 6.97E-03 & 5.92E+00 & 5.92E+00 & 2.72E-02 & 3.23E-01 & 5.90E+00 & 8.66E+00 & 2.18E-01 & 9.29E-01 & 1.52E+00 & 1.40E-03 & 1.82E-01 & 1.45E+00 & 2.40E-02 & 6.11E-01 & 1.94E+00 & 4.75E-01 & 9.39E-01 & 2.10E+00 \\
 & 20 & 7.91E-02 & 7.77E+00 & 1.05E+01 & 6.16E-02 & 1.72E-02 & 1.40E+01 & 1.72E+01 & 1.83E+00 & 3.46E+00 & 4.46E+00 & 9.44E-03 & 3.83E-01 & 1.11E+00 & 1.07E+00 & 3.76E+00 & 7.95E+00 & 1.98E+00 & 3.11E+00 & 4.86E+00 \\ \hline
\multirow{3}{*}{F17} & 5  & 1.26E-04 & 1.24E-02 & 6.38E-01 & 2.63E-06 & 0.00E+00 & 0.00E+00 & 0.00E+00 & 0.00E+00 & 0.00E+00 & 0.00E+00 & 0.00E+00 & 0.00E+00 & 0.00E+00 & 1.67E-05 & 2.38E-03 & 4.68E-02 & 0.00E+00 & 2.15E-06 & 5.02E-04 \\
 & 10 & 2.35E-03 & 1.23E+00 & 1.36E+00 & 1.86E-02 & 0.00E+00 & 0.00E+00 & 0.00E+00 & 0.00E+00 & 0.00E+00 & 0.00E+00 & 0.00E+00 & 0.00E+00 & 0.00E+00 & 5.67E-03 & 4.37E-02 & 4.36E-01 & 7.58E-05 & 1.04E-03 & 1.37E-02 \\
 & 20 & 4.79E-02 & 1.53E+00 & 2.18E+00 & 1.17E-01 & 0.00E+00 & 0.00E+00 & 7.99E-08 & 0.00E+00 & 0.00E+00 & 0.00E+00 & 0.00E+00 & 1.28E-04 & 2.07E-03 & 3.53E-02 & 1.99E-01 & 9.95E-01 & 9.17E-04 & 6.39E-03 & 2.13E-02 \\ \hline
\multirow{3}{*}{F18} & 5  & 5.60E-05 & 3.22E-01 & 1.37E+00 & 1.57E-04 & 0.00E+00 & 0.00E+00 & 0.00E+00 & 0.00E+00 & 0.00E+00 & 0.00E+00 & 0.00E+00 & 1.54E-02 & 6.45E-02 & 7.48E-04 & 2.52E-02 & 4.09E-01 & 2.06E-06 & 1.01E-04 & 2.79E-03 \\
 & 10 & 5.86E-02 & 2.46E+00 & 4.95E+00 & 3.45E-02 & 0.00E+00 & 0.00E+00 & 0.00E+00 & 0.00E+00 & 0.00E+00 & 0.00E+00 & 0.00E+00 & 2.51E-04 & 4.49E-03 & 1.56E-02 & 2.37E-01 & 5.93E-01 & 2.57E-04 & 7.20E-03 & 1.02E-01 \\
 & 20 & 2.35E-01 & 4.09E+00 & 1.71E+01 & 1.76E-01 & 0.00E+00 & 4.61E-05 & 1.05E-03 & 0.00E+00 & 1.46E-05 & 1.46E-03 & 2.06E-04 & 1.46E-03 & 2.17E-02 & 3.46E-01 & 1.18E+00 & 4.99E+00 & 1.15E-02 & 6.37E-02 & 4.46E-01 \\ \hline
\multirow{3}{*}{F19} & 5  & 1.57E-02 & 1.57E-02 & 1.57E-02 & 5.41E-02 & 8.86E-02 & 3.49E-01 & 7.07E-01 & 7.85E-03 & 6.02E-02 & 2.02E-01 & 0.00E+00 & 2.03E-02 & 1.08E+00 & 6.64E-02 & 2.72E-01 & 6.62E-01 & 9.54E-02 & 3.05E-01 & 6.69E-01 \\
 & 10 & 1.57E-02 & 1.57E-02 & 1.57E-02 & 1.31E-01 & 1.13E+00 & 1.79E+00 & 2.48E+00 & 7.11E-02 & 1.94E-01 & 3.71E-01 & 1.08E-02 & 5.73E-02 & 3.70E+00 & 8.56E-01 & 1.54E+00 & 2.12E+00 & 6.27E-01 & 1.03E+00 & 1.91E+00 \\
 & 20 & 2.59E-02 & 1.57E-02 & 1.72E-01 & 2.13E-01 & 2.85E+00 & 3.64E+00 & 4.11E+00 & 2.08E-01 & 4.17E-01 & 5.84E-01 & 2.56E-02 & 8.33E-02 & 5.33E+00 & 2.36E+00 & 3.78E+00 & 4.70E+00 & 1.71E+00 & 2.49E+00 & 3.24E+00 \\ \hline
\multirow{3}{*}{F20} & 5  & 2.37E-01 & 4.74E-01 & 9.08E-01 & 2.37E-01 & 0.00E+00 & 0.00E+00 & 0.00E+00 & 0.00E+00 & 0.00E+00 & 2.37E-01 & 0.00E+00 & 3.36E-01 & 9.08E-01 & 0.00E+00 & 2.37E-01 & 4.74E-01 & 0.00E+00 & 0.00E+00 & 2.41E-01 \\
 & 10 & 5.53E-01 & 7.11E-01 & 2.11E+00 & 7.70E-01 & 0.00E+00 & 0.00E+00 & 9.96E-01 & 0.00E+00 & 0.00E+00 & 2.17E-01 & 4.34E-01 & 1.02E+00 & 1.56E+00 & 0.00E+00 & 3.55E-01 & 7.70E-01 & 0.00E+00 & 1.71E-01 & 3.65E-01 \\
 & 20 & 1.14E+00 & 1.34E+00 & 1.80E+00 & 1.81E+00 & 5.92E-02 & 1.78E-01 & 6.91E-01 & 5.92E-02 & 1.18E-01 & 2.76E-01 & 1.24E+00 & 1.74E+00 & 2.56E+00 & 2.96E-01 & 6.02E-01 & 8.49E-01 & 1.87E-01 & 2.90E-01 & 4.40E-01 \\ \hline
\multirow{3}{*}{F21} & 5  & 0.00E+00 & 0.00E+00 & 2.00E+00 & 0.00E+00 & 0.00E+00 & 0.00E+00 & 1.25E+00 & 0.00E+00 & 0.00E+00 & 1.25E+00 & 0.00E+00 & 1.25E+00 & 1.25E+00 & 0.00E+00 & 3.46E-01 & 3.17E+00 & 0.00E+00 & 0.00E+00 & 0.00E+00 \\
 & 10 & 0.00E+00 & 1.75E-08 & 2.26E+00 & 4.24E-05 & 0.00E+00 & 9.30E-01 & 1.25E+00 & 0.00E+00 & 1.25E+00 & 1.80E+00 & 2.26E+00 & 2.26E+00 & 2.26E+00 & 0.00E+00 & 1.80E+00 & 7.32E+00 & 0.00E+00 & 0.00E+00 & 0.00E+00 \\
 & 20 & 0.00E+00 & 1.74E-06 & 1.41E+01 & 1.34E-02 & 0.00E+00 & 2.12E+00 & 1.71E+01 & 0.00E+00 & 0.00E+00 & 2.47E+00 & 0.00E+00 & 0.00E+00 & 1.71E+01 & 0.00E+00 & 2.36E+00 & 2.70E+01 & 0.00E+00 & 0.00E+00 & 0.00E+00 \\ \hline
\multirow{3}{*}{F22} & 5  & 0.00E+00 & 0.00E+00 & 0.00E+00 & 2.53E-05 & 0.00E+00 & 0.00E+00 & 0.00E+00 & 0.00E+00 & 0.00E+00 & 6.92E-01 & 0.00E+00 & 0.00E+00 & 0.00E+00 & 0.00E+00 & 5.29E-08 & 6.92E-01 & 0.00E+00 & 0.00E+00 & 0.00E+00 \\
 & 10 & 3.87E-07 & 7.25E-08 & 1.95E+00 & 4.54E-02 & 1.95E+00 & 1.95E+00 & 1.95E+00 & 1.95E+00 & 1.95E+00 & 1.95E+00 & 1.95E+00 & 1.95E+00 & 1.95E+00 & 5.54E-07 & 1.95E+00 & 5.06E+00 & 0.00E+00 & 0.00E+00 & 0.00E+00 \\
 & 20 & 6.92E-01 & 6.92E-01 & 1.96E+00 & 1.42E-01 & 6.92E-01 & 1.95E+00 & 1.95E+00 & 1.95E+00 & 1.95E+00 & 1.95E+00 & 1.95E+00 & 1.95E+00 & 1.95E+00 & 6.92E-01 & 1.95E+00 & 3.01E+01 & 0.00E+00 & 6.92E-01 & 1.95E+00 \\ \hline
\multirow{3}{*}{F23} & 5  & 2.81E-05 & 1.51E-01 & 1.54E-01 & 8.44E-01 & 1.42E-01 & 2.72E-01 & 3.50E-01 & 0.00E+00 & 0.00E+00 & 3.67E-01 & 0.00E+00 & 0.00E+00 & 3.20E-01 & 3.22E-01 & 6.46E-01 & 1.16E+00 & 1.16E-01 & 5.52E-01 & 9.09E-01 \\
 & 10 & 1.08E-02 & 2.22E-01 & 1.19E-01 & 2.02E-01 & 5.03E-01 & 8.33E-01 & 1.18E+00 & 0.00E+00 & 2.65E-01 & 4.13E-01 & 0.00E+00 & 1.51E-02 & 3.05E-01 & 2.98E-01 & 9.19E-01 & 1.26E+00 & 8.98E-02 & 6.00E-01 & 8.93E-01 \\
 & 20 & 2.39E-02 & 1.35E-01 & 6.14E-01 & 1.18E+00 & 1.14E+00 & 1.61E+00 & 1.93E+00 & 2.45E-01 & 3.67E-01 & 4.90E-01 & 0.00E+00 & 2.12E-02 & 1.54E-01 & 8.89E-01 & 1.40E+00 & 1.73E+00 & 6.69E-01 & 9.43E-01 & 1.32E+00 \\ \hline
\multirow{3}{*}{F24} & 5  & 8.00E-01 & 8.00E+00 & 6.87E+00 & 7.05E+00 & 3.53E+00 & 7.30E+00 & 1.11E+01 & 1.67E+00 & 5.95E+00 & 7.31E+00 & 5.08E+00 & 5.38E+00 & 7.78E+00 & 1.65E+00 & 6.90E+00 & 1.09E+01 & 2.51E+00 & 5.82E+00 & 8.29E+00 \\
 & 10 & 2.65E+00 & 3.76E+01 & 6.16E+01 & 2.26E+01 & 1.70E+01 & 3.26E+01 & 3.92E+01 & 1.16E+01 & 1.30E+01 & 1.42E+01 & 1.03E+01 & 1.10E+01 & 1.54E+01 & 8.29E+00 & 3.01E+01 & 4.32E+01 & 1.41E+01 & 2.10E+01 & 2.75E+01 \\
 & 20 & 4.02E+01 & 9.17E+01 & 1.36E+02 & 1.53E+02 & 6.00E+01 & 9.90E+01 & 1.16E+02 & 2.33E+01 & 2.57E+01 & 2.79E+01 & 2.22E+01 & 2.43E+01 & 3.60E+01 & 2.21E+01 & 1.05E+02 & 1.38E+02 & 3.59E+01 & 5.52E+01 & 7.33E+01
\end{tabular} \label{bbob_res2}
\egroup
}
\end{table}
\end{landscape}

% \begin{landscape}
% \begin{table}[!h]
% \caption{Detailed statistics of the 30 runs of the selected algorithms on the BBOB benchmark set, F13-24.}
% \centering
% %\scriptsize	
% \includegraphics[width=\linewidth]{tab5.png}
% \label{bbob_res2}
% \end{table}
% \end{landscape}

\subsection{CEC 2022 Results}
\begin{figure}[!h]
    \centering
    \begin{tabular}{cc}
        \includegraphics[width = 0.4\linewidth]{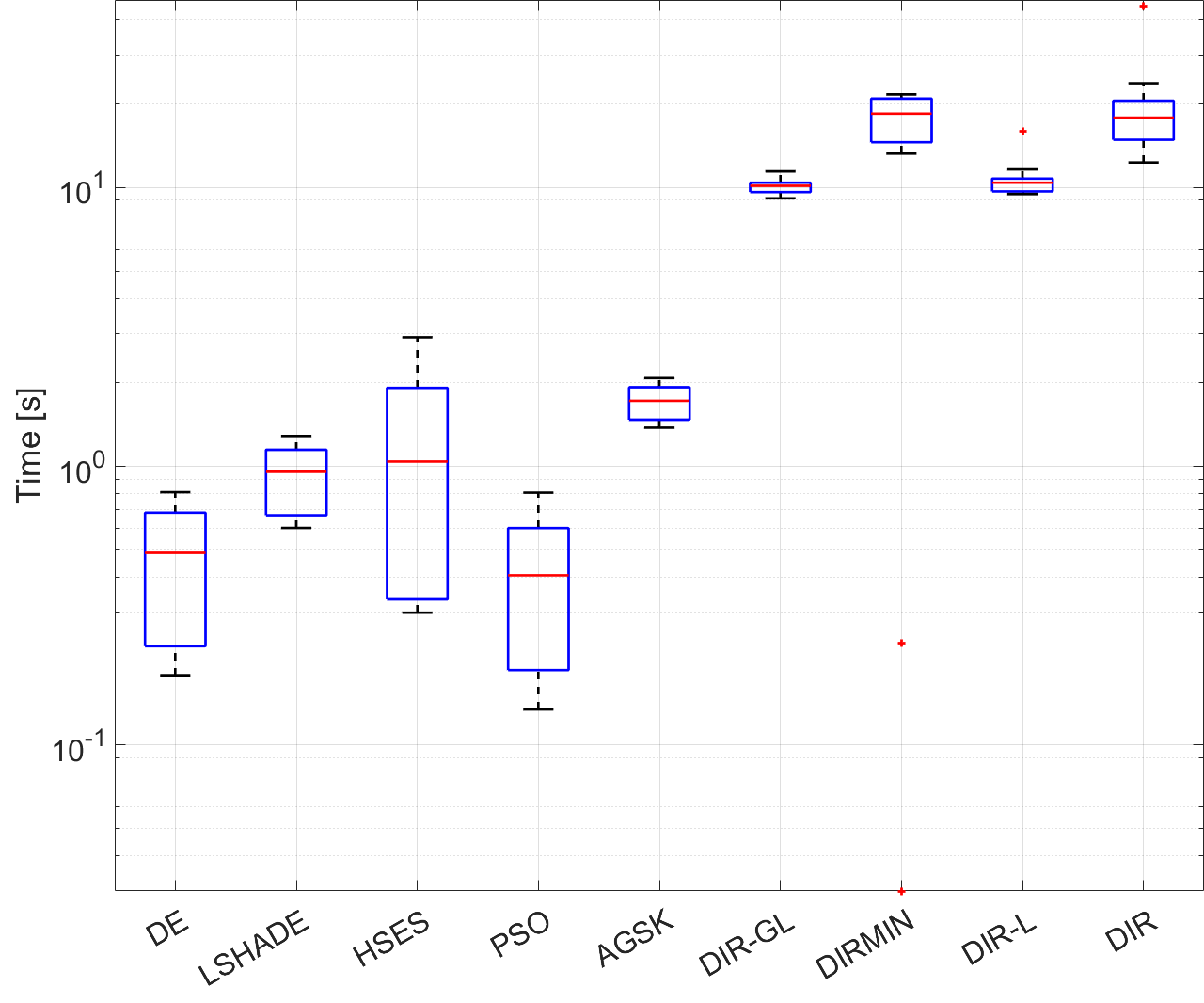} & \includegraphics[width = 0.4\linewidth]{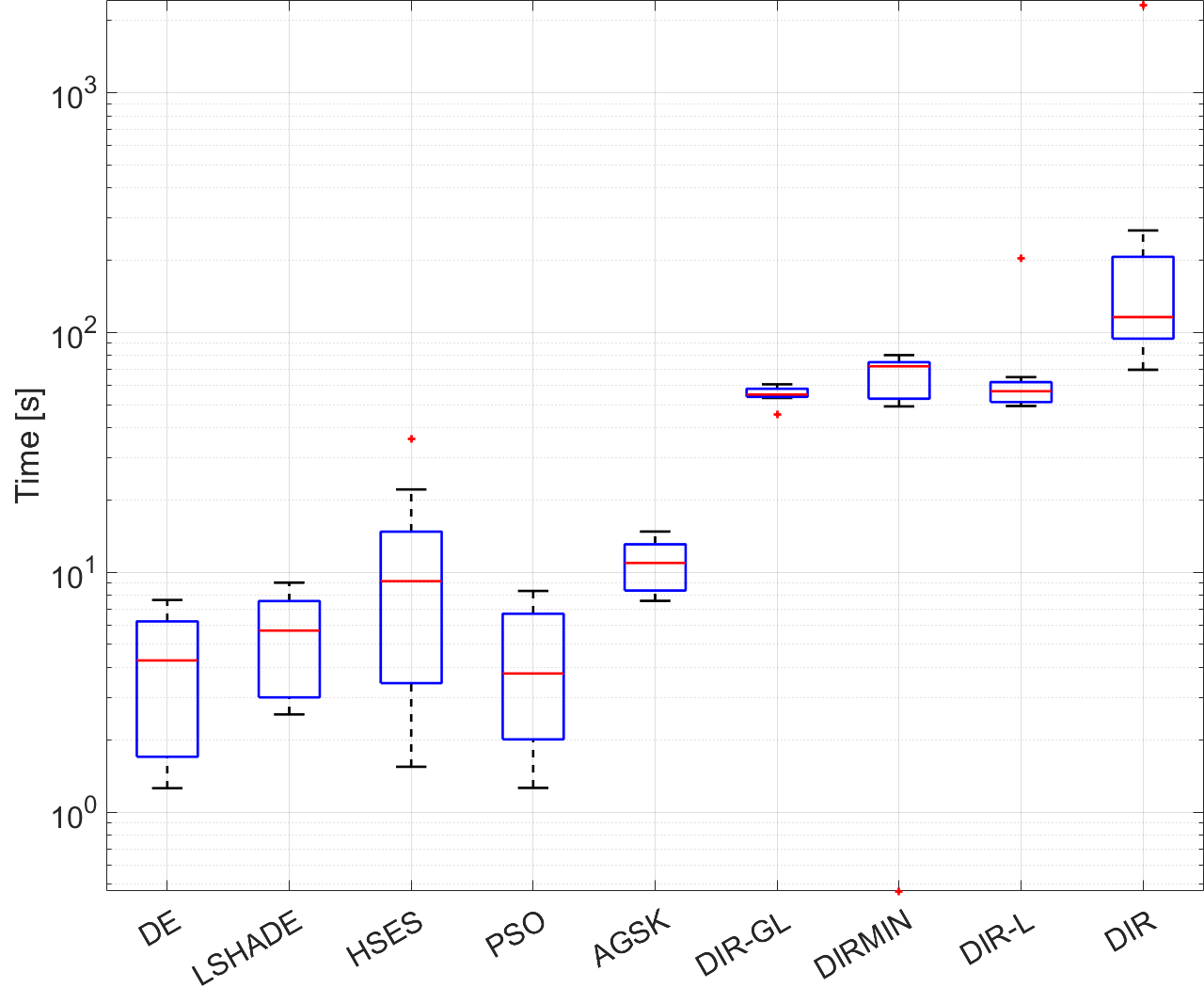} \\
        a) & b)
    \end{tabular}
    \caption{Time complexity of the different algorithms on the CEC 2022 benchmark set, a) $D=10$, b) $D=20$.} \label{fig_cec2022_time}
\end{figure}

Next, we focus on the results on the CEC 2022 benchmark set. In Figure \ref{fig_cec2022_time} we see roughly an order of magnitude difference in time complexity between the DIRECT-type and nature-inspired methods.

\begin{figure}[!h]
    \centering
    \begin{tabular}{cc}
        \includegraphics[width = 0.45\linewidth]{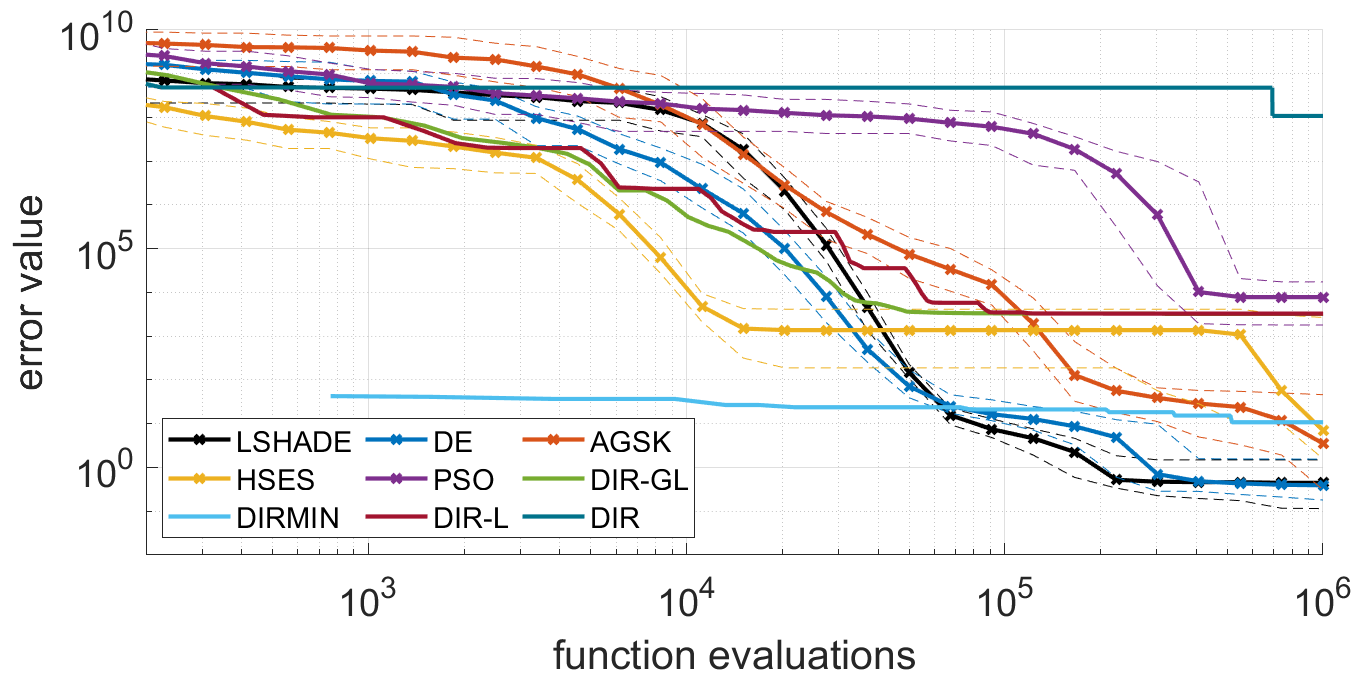} & \includegraphics[width = 0.45\linewidth]{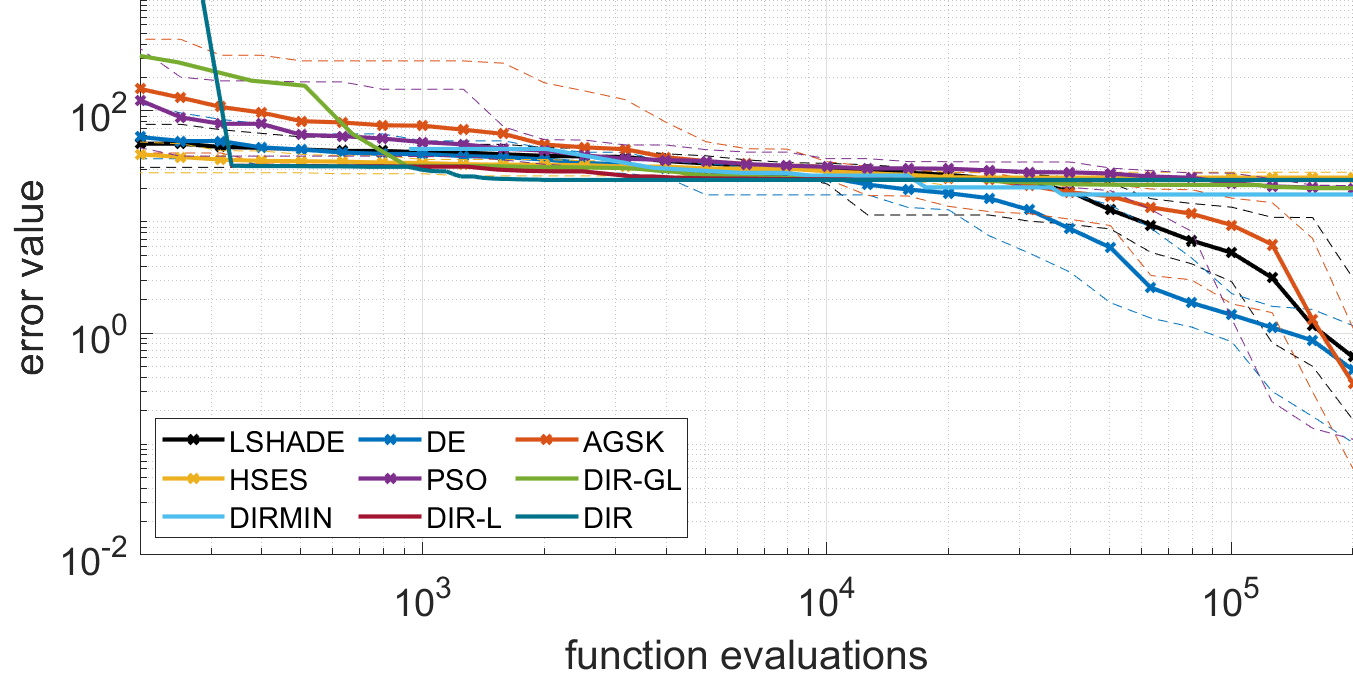} \\
        a) & b)
    \end{tabular}
    \caption{Convergence plots from the CEC 2022 benchmark set, a) F6 in $D=20$, b) F8 in $D=10$.}
    \label{conv_cec2022}
\end{figure}

In Table \ref{cec22_res} are reported detailed statistics of the results of the computations on the CEC 2022 benchmark set, and Figure \ref{conv_cec2022} shows representative convergence plots. When comparing the results of the DIRECT-type methods against the best results of the nature-inspired ones over the 30 runs, we once again see that on most problems, the nature-inspired methods (in particular LSHADE, AGSK, and DE) outperform the DIRECT-type ones by a large margin. The only instances where a DIRECT-type method was better that the nature-inspired ones was in F2 in $D = 10$, and in F11 in $D=20$, both times using the DIRMIN algorithm. When comparing the median of the runs, the conclusions remain roughly the same, with LSHADE being the best method (apart from F2 in both dimensions, F8 in $D= 20$, and F11 in $D= 10$), and the nature-inspired methods being on the whole better than the DIRECT-type ones. However, this situation changes substantially when comparing against the worst results. Here, the nature-inspired methods outperform the DIRECT-type ones only on 10 of the 24 instances, while the reverse is true on 4 instances, and the rest are draws (i.e. the best of the DIRECT-type methods is as good as the best of the nature-inspired ones).

\begin{figure}[!h]
    \centering
    \begin{tabular}{cc}
        \includegraphics[width = 0.4\linewidth]{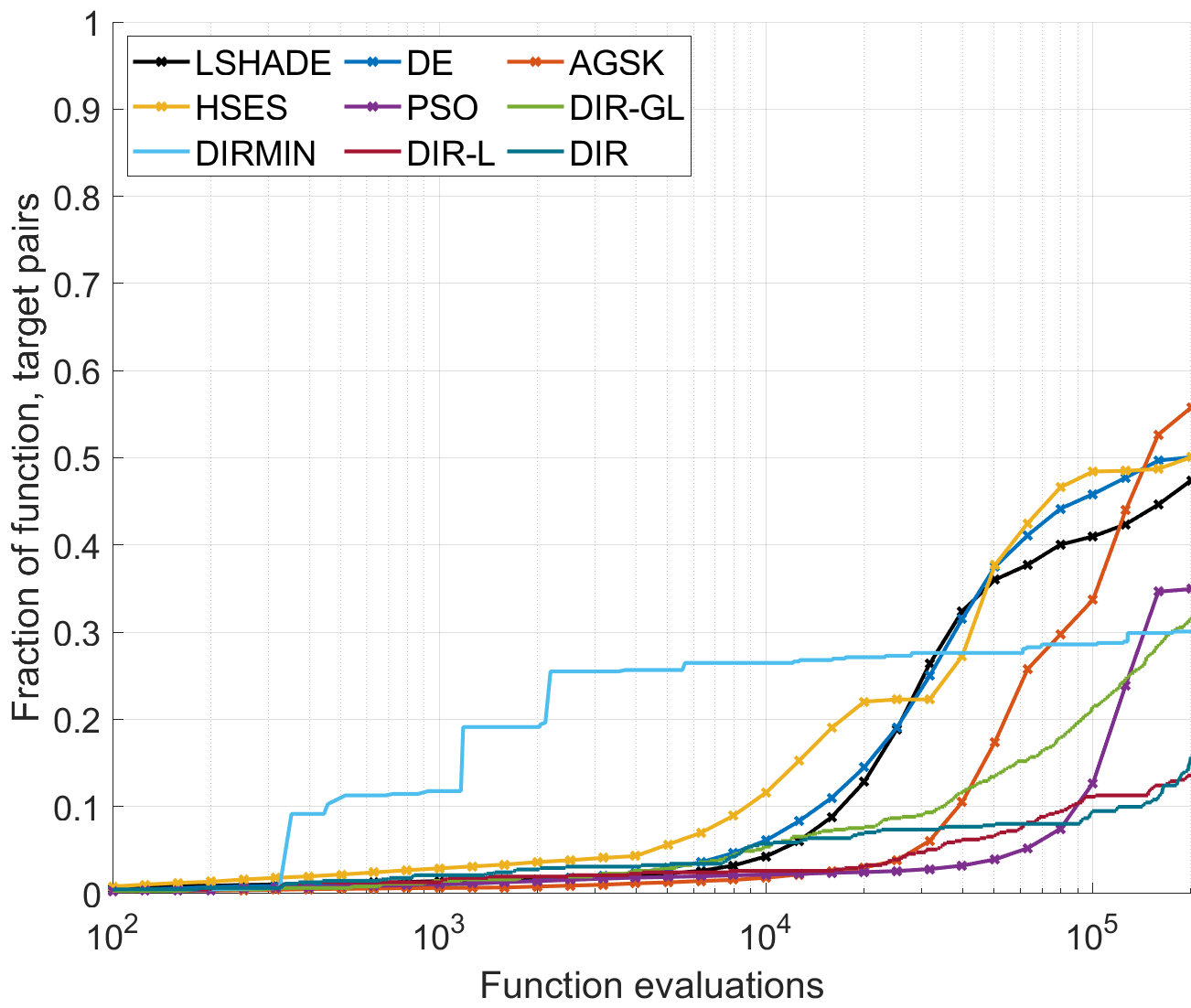} & \includegraphics[width = 0.4\linewidth]{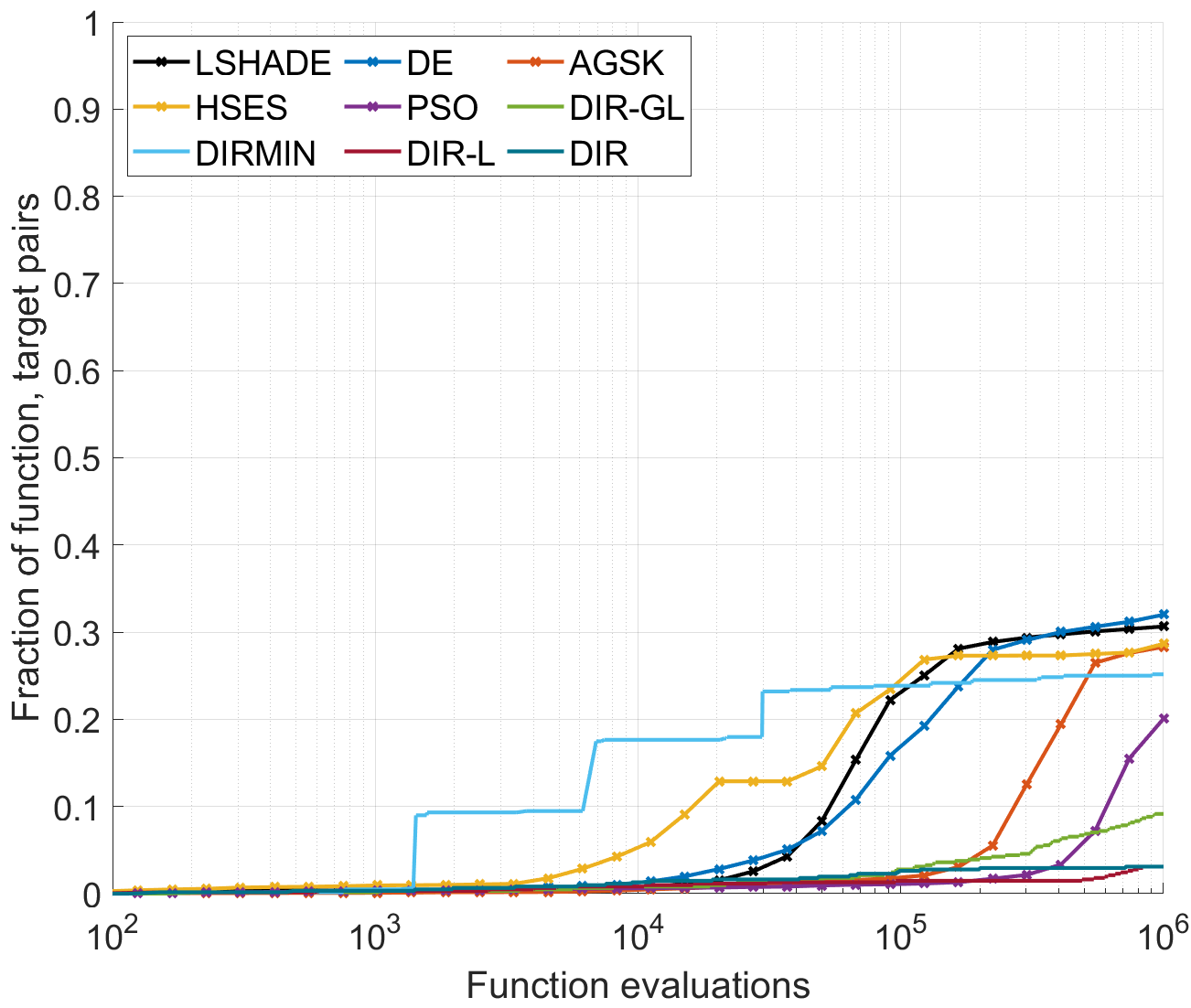} \\
        a) & b)
    \end{tabular}
    \caption{ECD of the number of objective function evaluations for different target precisions on the CEC 2022 benchmark set, a) $D=10$, b) $D=20$.} \label{fig_cec2022_ecd}
\end{figure}

The results of rank test are reported in Table \ref{cec2022_ranks}, while the ECD plots are shown in Figure \ref{fig_cec2022_ecd}. In the early stages of the search, HSES and some the of the DIRECT-type methods (in this case DIR and DIRMIN) are able to find good solutions quickly. However, as the iterations progress, they get dominated by DE and LSHADE. Overall, the best of the DIRECT-type methods is the DIR-GL, but its performance was worse than that of all the nature-inspired methods with the exception of PSO.

\begin{table}[!h]
\caption{Mean ranks from Friedman tests at different stages of the search, CEC 2022 benchmark set.}
\centering
\resizebox{0.6\columnwidth}{!}{%
\setlength{\tabcolsep}{0.3em}
\bgroup
\def\arraystretch{1.1}
\begin{tabular}{l|ccccccccc}
FES            & \multicolumn{1}{c}{DE} & \multicolumn{1}{c}{LSHADE} & \multicolumn{1}{c}{HSES} & \multicolumn{1}{c}{PSO} &\multicolumn{1}{c}{AGSK} & \multicolumn{1}{c}{DIR-GL} & \multicolumn{1}{c}{DIRMIN} & \multicolumn{1}{c}{DIR-L} & \multicolumn{1}{c}{DIR} \\ \hline 
(1/256)*MaxFES & 5.4                    & 4.9                        & \textbf{2.4}                     & 5.9                     & 8.1                      & 5.4                        & \textbf{4.1}                        & 4.9                       & \textbf{3.9}                     \\
(1/128)*MaxFES & 5.1                    & 5.6                        & \textbf{2.6}                      & 6.2                     & 8.0                      & 4.8                        & \textbf{3.9}                        & 5.0                       & \textbf{3.7}                     \\
(1/32)*MaxFES  & \textbf{3.4}                    & {4.8}                        & \textbf{3.1}                      & 7.1                     & 7.6                      & {4.8}                        & \textbf{3.5}                        & 5.7                       & 5.1                     \\
(1/8)*MaxFES   & \textbf{3.1}                    & \textbf{2.9}                        & \textbf{3.8}                      & 7.8                     & 5.9                      & 4.3                        & 3.9                        & 6.9                       & 6.4                     \\
(1/4)*MaxFES   & \textbf{3.1}                    & \textbf{2.3}                        & 4.5                      & 7.8                     & 4.5                      & 4.6                        & \textbf{4.3}                        & 7.1                       & 6.9                     \\
(1/2)*MaxFES   & \textbf{2.9}                    & \textbf{2.3}                        & 4.5                      & 7.3                     & \textbf{3.7}                      & 4.7                        & 5.0                        & 7.4                       & 7.1                     \\
(3/4)*MaxFES   & \textbf{3.0}                    & \textbf{2.4}                        & 4.8                      & 6.1                     & \textbf{3.2}                     & 5.1                        & 5.3                        & 7.8                       & 7.4                     \\
MaxFES         & \textbf{3.1}                    & \textbf{2.6}                        & 4.7                      & 5.5                     & \textbf{3.0}                      & 5.2                        & 5.5                        & 7.8                       & 7.5                    
              
\end{tabular} \label{cec2022_ranks}
\egroup
}
\end{table}

\begin{landscape}
\begin{table}[!h]
\caption{Detailed statistics of the 30 runs of the selected algorithms on the CEC 2022 benchmark set.}
\centering
%\scriptsize	
\resizebox{1.00\columnwidth}{!}{%
\setlength{\tabcolsep}{0.3em}
\bgroup
\def\arraystretch{1.3}
\begin{tabular}{cc|cccc|ccc|ccc|ccc|ccc|ccc}
\multicolumn{2}{l}{}& \multicolumn{4}{c}{Direct methods} & \multicolumn{3}{c}{DE} & \multicolumn{3}{c}{LSHADE} & \multicolumn{3}{c}{HSES} & \multicolumn{3}{c}{PSO} & \multicolumn{3}{c}{AGSK} \\
\multicolumn{1}{l}{} & D                    & DIR-GL               & DIRMIN               & DIR-L                & DIR                  & min                  & median   & max                            & min                  & median   & max                                  & min                  & median   & max                               & min                  & median   & max     & min                  & median   & max                               \\ \hline
\multirow{2}{*}{F1}  & 10                    & 1.32E-06                   & 0.00E+00                   & 1.83E-07                  & 2.69E+03                & 0.00E+00                & 0.00E+00                   & 0.00E+00                & 0.00E+00                & 0.00E+00                   & 0.00E+00                & 0.00E+00                & 0.00E+00                   & 0.00E+00                & 0.00E+00                & 0.00E+00                   & 0.00E+00                & 0.00E+00                & 0.00E+00                   & 0.00E+00                \\
  & 20                    & 1.23E+00                   & 0.00E+00                   & 1.00E+00                  & 1.04E+04                & 0.00E+00                & 0.00E+00                   & 0.00E+00                & 0.00E+00                & 0.00E+00                   & 0.00E+00                & 0.00E+00                & 0.00E+00                   & 0.00E+00                & 0.00E+00                & 0.00E+00                   & 0.00E+00                & 0.00E+00                & 0.00E+00                   & 0.00E+00                \\ \hline
\multirow{2}{*}{F2}  & 10                    & 1.04E-03                   & 0.00E+00                   & 4.82E+00                  & 3.77E-02                & 0.00E+00                & 3.99E+00                   & 8.92E+00                & 3.99E+00                & 3.99E+00                   & 8.92E+00                & 0.00E+00                & 0.00E+00                   & 0.00E+00                & 6.31E-02                & 4.58E+00                   & 6.83E+01                & 0.00E+00                & 0.00E+00                   & 0.00E+00                \\
  & 20                    & 3.08E+00                   & 0.00E+00                   & 1.63E+02                  & 5.53E+01                & 4.49E+01                & 4.49E+01                   & 4.91E+01                & 4.49E+01                & 4.91E+01                   & 4.91E+01                & 4.92E+01                & 5.00E+01                   & 6.91E+01                & 4.77E+01                & 5.19E+01                   & 9.15E+01                & 0.00E+00                & 4.49E+01                   & 4.91E+01                \\ \hline
\multirow{2}{*}{F3}  & 10                    & 5.59E-04                   & 1.40E+01                   & 4.25E+00                  & 3.95E-02                & 0.00E+00                & 0.00E+00                   & 0.00E+00                & 0.00E+00                & 0.00E+00                   & 0.00E+00                & 0.00E+00                & 0.00E+00                   & 0.00E+00                & 0.00E+00                & 0.00E+00                   & 0.00E+00                & 0.00E+00                & 0.00E+00                   & 1.42E-06                \\
  & 20                    & 1.31E-02                   & 1.73E+01                   & 3.84E+00                  & 3.87E+00                & 0.00E+00                & 0.00E+00                   & 0.00E+00                & 0.00E+00                & 0.00E+00                   & 0.00E+00                & 0.00E+00                & 0.00E+00                   & 0.00E+00                & 0.00E+00                & 0.00E+00                   & 0.00E+00                & 0.00E+00                & 0.00E+00                   & 0.00E+00                \\ \hline
\multirow{2}{*}{F4}  & 10                    & 3.98E+00                   & 1.29E+01                   & 1.99E+01                  & 1.99E+01                & 3.98E+00                & 2.00E+01                   & 2.98E+01                & 9.95E-01                & 1.99E+00                   & 5.97E+00                & 0.00E+00                & 9.95E-01                   & 2.98E+00                & 5.97E+00                & 1.24E+01                   & 2.69E+01                & 2.98E+00                & 8.10E+00                   & 1.34E+01                \\
  & 20                    & 1.59E+01                   & 2.79E+01                   & 7.66E+01                  & 7.66E+01                & 7.96E+00                & 2.23E+01                   & 9.49E+01                & 2.98E+00                & 3.98E+00                   & 5.97E+00                & 9.95E-01                & 3.98E+00                   & 6.96E+00                & 1.99E+01                & 3.48E+01                   & 6.52E+01                & 1.19E+01                & 4.34E+01                   & 7.58E+01                \\ \hline
\multirow{2}{*}{F5}  & 10                    & 0.00E+00                   & 5.44E-01                   & 9.07E-02                  & 1.04E-08                & 0.00E+00                & 0.00E+00                   & 0.00E+00                & 0.00E+00                & 0.00E+00                   & 0.00E+00                & 0.00E+00                & 0.00E+00                   & 0.00E+00                & 0.00E+00                & 0.00E+00                   & 5.44E-01                & 0.00E+00                & 0.00E+00                   & 0.00E+00                \\
  & 20                    & 2.21E+00                   & 1.01E+02                   & 1.57E+03                  & 3.20E+00                & 0.00E+00                & 0.00E+00                   & 0.00E+00                & 0.00E+00                & 0.00E+00                   & 0.00E+00                & 0.00E+00                & 0.00E+00                   & 0.00E+00                & 8.95E-02                & 3.59E+00                   & 4.77E+01                & 0.00E+00                & 8.95E-02                   & 6.33E-01                \\ \hline
\multirow{2}{*}{F6}  & 10                    & 1.38E+02                   & 1.21E-01                   & 2.99E+02                  & 8.34E+01                & 2.23E-02                & 2.25E-01                   & 5.00E-01                & 3.88E-02                & 3.45E-01                   & 5.00E-01                & 1.02E-02                & 9.30E+00                   & 1.02E+03                & 1.08E+01                & 4.78E+02                   & 3.00E+03                & 1.83E-02                & 1.51E-01                   & 4.22E-01                \\
  & 20                    & 3.22E+03                   & 1.08E+01                   & 3.27E+03                  & 1.08E+08                & 1.80E-01                & 3.91E-01                   & 1.55E+00                & 1.15E-01                & 4.44E-01                   & 1.49E+00                & 1.58E+00                & 6.96E+00                   & 2.67E+03                & 1.80E+03                & 7.78E+03                   & 1.74E+04                & 3.10E-01                & 3.53E+00                   & 4.60E+01                \\ \hline
\multirow{2}{*}{F7}  & 10                    & 2.50E+01                   & 2.92E+01                   & 7.56E+01                  & 4.02E+02                & 0.00E+00                & 0.00E+00                   & 6.24E-01                & 0.00E+00                & 6.93E-08                   & 2.97E-05                & 0.00E+00                & 9.95E-01                   & 3.69E+00                & 2.22E-07                & 2.00E+01                   & 2.16E+01                & 0.00E+00                & 0.00E+00                   & 0.00E+00                \\
  & 20                    & 7.12E+01                   & 7.95E+01                   & 1.78E+02                  & 4.76E+02                & 0.00E+00                & 4.68E-01                   & 2.04E+01                & 1.41E+00                & 2.85E+00                   & 2.01E+01                & 4.97E+00                & 2.91E+01                   & 3.62E+01                & 3.17E-01                & 2.23E+01                   & 4.27E+01                & 7.61E-04                & 1.09E+01                   & 2.10E+01                \\ \hline
\multirow{2}{*}{F8}  & 10                    & 2.02E+01                   & 1.77E+01                   & 2.39E+01                  & 2.39E+01                & 1.03E-01                & 4.64E-01                   & 1.17E+00                & 1.65E-01                & 6.08E-01                   & 3.11E+00                & 2.11E+01                & 2.58E+01                   & 2.99E+01                & 1.10E-01                & 2.02E+01                   & 2.13E+01                & 5.99E-02                & 3.51E-01                   & 1.11E+00                \\
  & 20                    & 2.06E+01                   & 2.30E+01                   & 1.41E+02                  & 1.40E+02                & 2.27E-03                & 1.14E+00                   & 2.36E+01                & 7.53E+00                & 1.97E+01                   & 2.05E+01                & 2.51E+01                & 2.97E+01                   & 3.40E+01                & 2.05E+01                & 2.11E+01                   & 2.22E+01                & 1.73E+01                & 2.18E+01                   & 2.23E+01                \\ \hline
\multirow{2}{*}{F9}  & 10                    & 2.29E+02                   & 2.29E+02                   & 2.30E+02                  & 2.29E+02                & 2.29E+02                & 2.29E+02                   & 2.29E+02                & 2.29E+02                & 2.29E+02                   & 2.29E+02                & 2.40E+02                & 2.49E+02                   & 2.63E+02                & 2.31E+02                & 2.32E+02                   & 2.33E+02                & 0.00E+00                & 2.29E+02                   & 2.29E+02                \\
  & 20                    & 1.81E+02                   & 1.81E+02                   & 1.89E+02                  & 2.00E+02                & 1.81E+02                & 1.81E+02                   & 1.81E+02                & 1.81E+02                & 1.81E+02                   & 1.81E+02                & 1.85E+02                & 1.88E+02                   & 1.91E+02                & 1.82E+02                & 1.83E+02                   & 1.84E+02                & 1.81E+02                & 1.81E+02                   & 1.81E+02                \\ \hline
\multirow{2}{*}{F10} & 10                    & 1.00E+02                   & 1.00E+02                   & 1.00E+02                  & 1.00E+02                & 1.00E+02                & 1.00E+02                   & 1.00E+02                & 1.00E+02                & 1.00E+02                   & 1.00E+02                & 1.00E+02                & 2.05E+02                   & 2.08E+02                & 1.00E+02                & 1.00E+02                   & 2.18E+02                & 1.00E+02                & 1.00E+02                   & 1.00E+02                \\
 & 20                    & 1.00E+02                   & 1.00E+02                   & 1.00E+02                  & 1.01E+02                & 1.00E+02                & 1.00E+02                   & 1.00E+02                & 1.00E+02                & 1.00E+02                   & 1.00E+02                & 6.92E+00                & 1.00E+02                   & 2.25E+02                & 2.60E+01                & 1.62E+02                   & 4.89E+02                & 0.00E+00                & 1.00E+02                   & 1.00E+02                \\ \hline
\multirow{2}{*}{F11} & 10                    & 2.20E-05                   & 3.83E-06                   & 1.50E+02                  & 9.92E+01                & 0.00E+00                & 0.00E+00                   & 0.00E+00                & 0.00E+00                & 0.00E+00                   & 0.00E+00                & 0.00E+00                & 0.00E+00                   & 3.00E+02                & 0.00E+00                & 0.00E+00                   & 4.00E+02                & 0.00E+00                & 0.00E+00                   & 0.00E+00                \\
 & 20                    & 4.00E+02                   & 1.02E-05                   & 3.00E+02                  & 4.86E+02                & 3.00E+02                & 4.00E+02                   & 4.00E+02                & 3.00E+02                & 3.00E+02                   & 3.00E+02                & 3.00E+02                & 3.00E+02                   & 3.00E+02                & 3.00E+02                & 3.00E+02                   & 4.00E+02                & 0.00E+00                & 3.00E+02                   & 4.00E+02                \\ \hline
\multirow{2}{*}{F12} & 10                    & 1.63E+02                   & 1.64E+02                   & 1.92E+02                  & 1.63E+02                & 1.63E+02                & 1.64E+02                   & 1.65E+02                & 1.59E+02                & 1.61E+02                   & 1.65E+02                & 1.66E+02                & 1.67E+02                   & 1.79E+02                & 1.65E+02                & 1.66E+02                   & 1.67E+02                & 1.59E+02                & 1.59E+02                   & 1.60E+02                \\
 & 20                    & 2.36E+02                   & 2.63E+02                   & 2.83E+02                  & 2.53E+02                & 2.32E+02                & 2.32E+02                   & 2.44E+02                & 2.32E+02                & 2.32E+02                   & 2.35E+02                & 2.46E+02                & 2.54E+02                   & 2.68E+02                & 2.44E+02                & 2.61E+02                   & 3.30E+02                & 2.31E+02                & 2.36E+02                   & 2.41E+02               

\end{tabular} \label{cec22_res}
\egroup
}
\end{table}
\end{landscape}

% \begin{landscape}
% \begin{table}[!h]
% \caption{Detailed statistics of the 30 runs of the selected algorithms on the CEC 2022 benchmark set.}
% \centering
% %\scriptsize	
% \includegraphics[width=\linewidth]{tab6.png}
% \label{cec22_res}
% \end{table}
% \end{landscape}

\subsection{Ambiguous Benchmark Set Results}
\begin{figure}[!h]
    \centering
    \begin{tabular}{cc}
        \includegraphics[width = 0.4\linewidth]{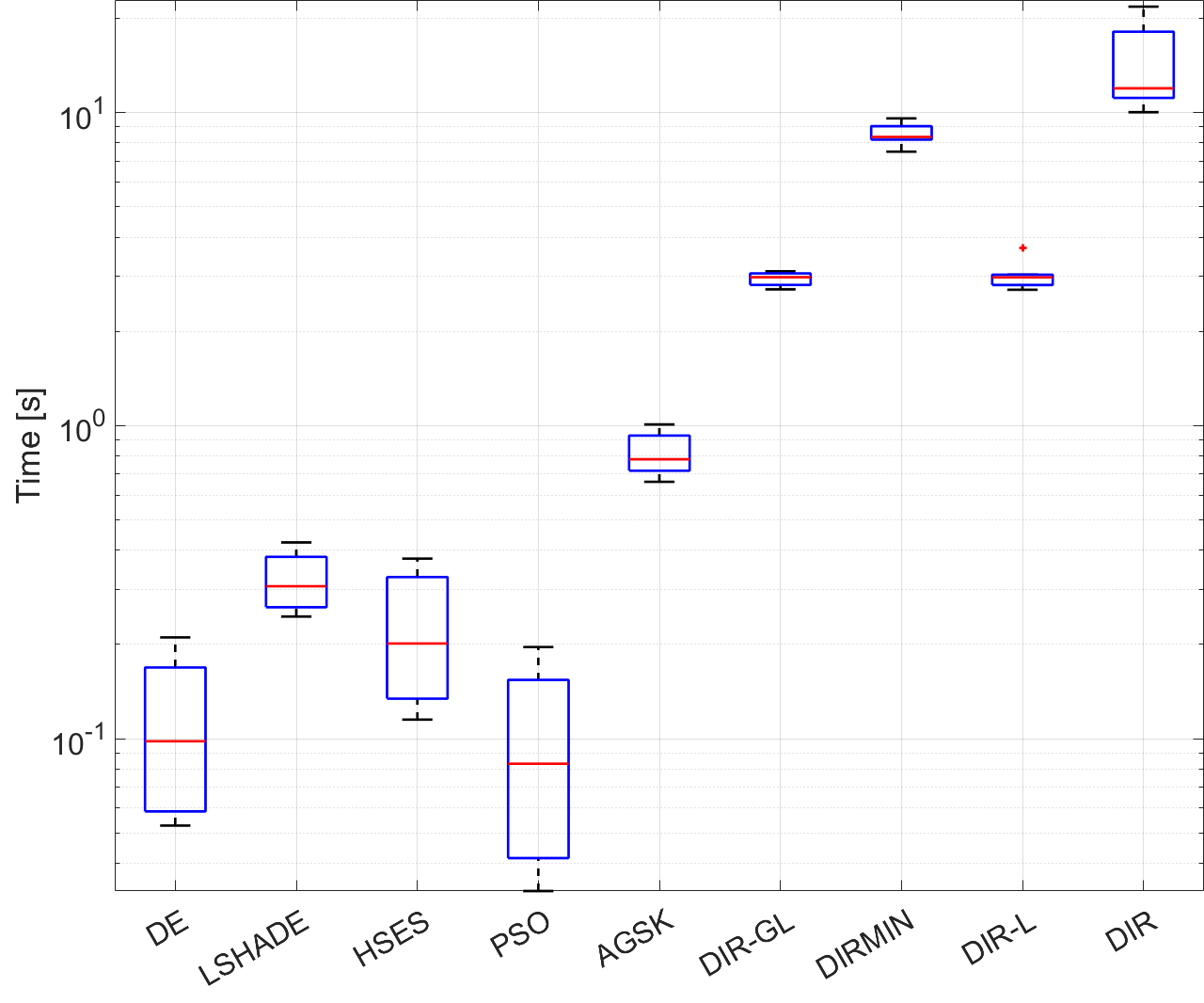} & \includegraphics[width = 0.4\linewidth]{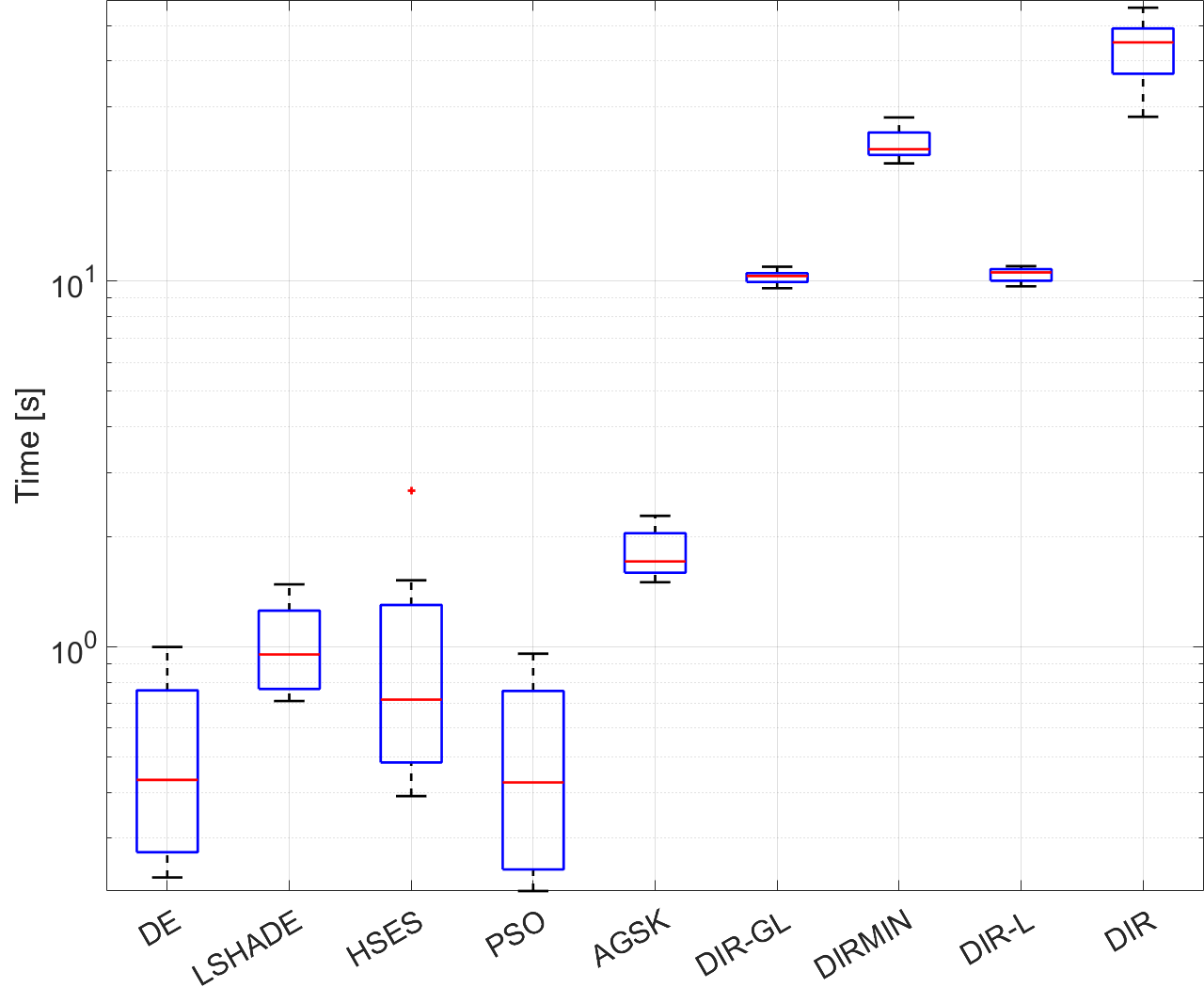} \\
        a) & b) \vspace{3mm}\\
        \includegraphics[width = 0.4\linewidth]{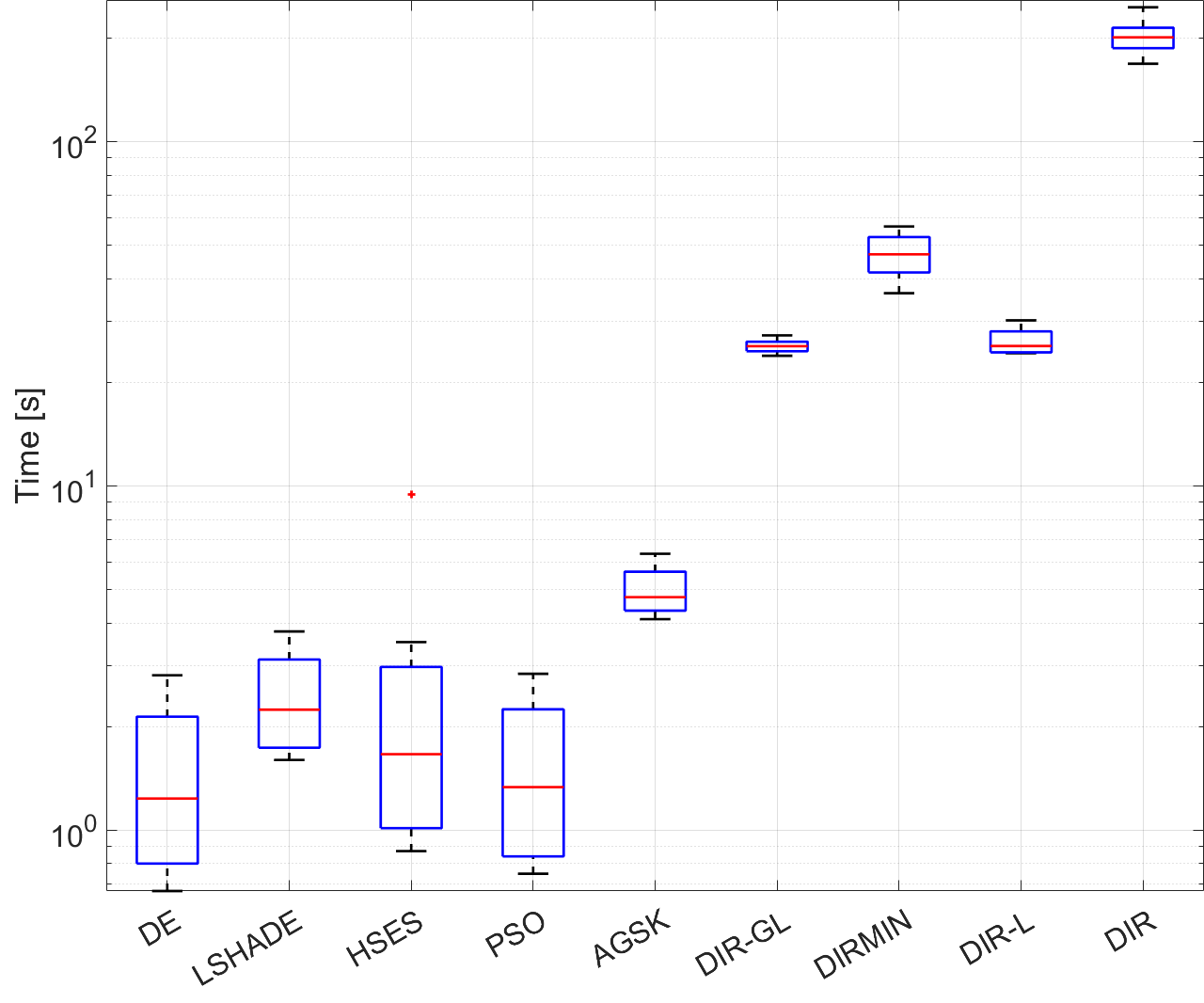} & \includegraphics[width = 0.4\linewidth]{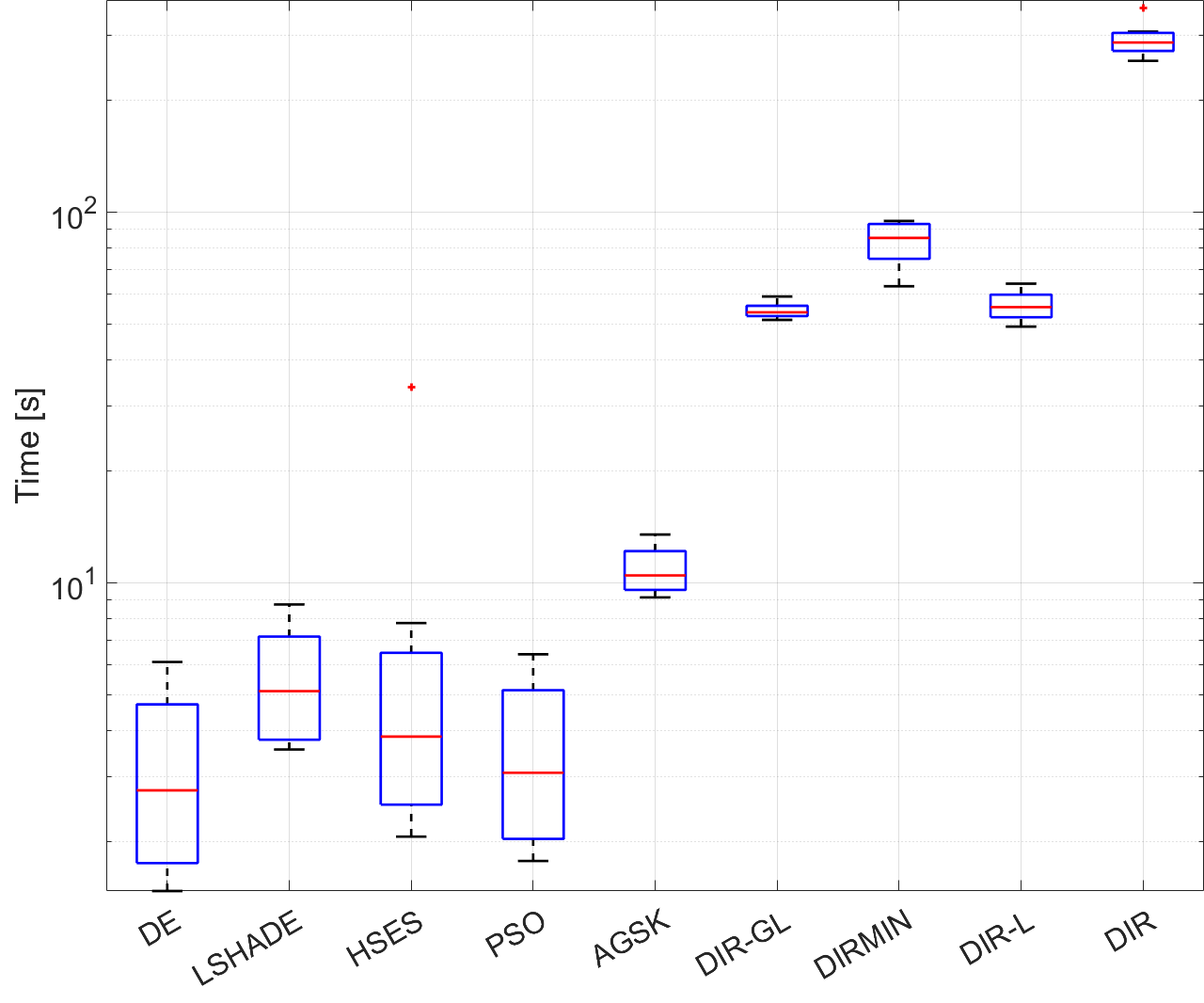} \\
        c) & d)
    \end{tabular}
    \caption{Time complexity of the different algorithms on the ambiguous benchmark set, a) $D=5$, b) $D=10$, c) $D=15$, d) $D=20$.} \label{fig_amb_time}
\end{figure}

Finally, we perform the same analyses for the results on the ambiguous benchmark set. We can see that, similarly as on the previous benchmark sets, the time complexity (shown in Figure \ref{fig_amb_time}) of the DIRECT-type methods is roughly an order of magnitude higher than that of the nature-inspired ones. A slight difference can be seen in the time complexity of the original DIR method, that seems to be growing a bit faster than the other methods when the dimension increases.

\begin{figure}[!h]
    \centering
    \begin{tabular}{cc}
        \includegraphics[width = 0.45\linewidth]{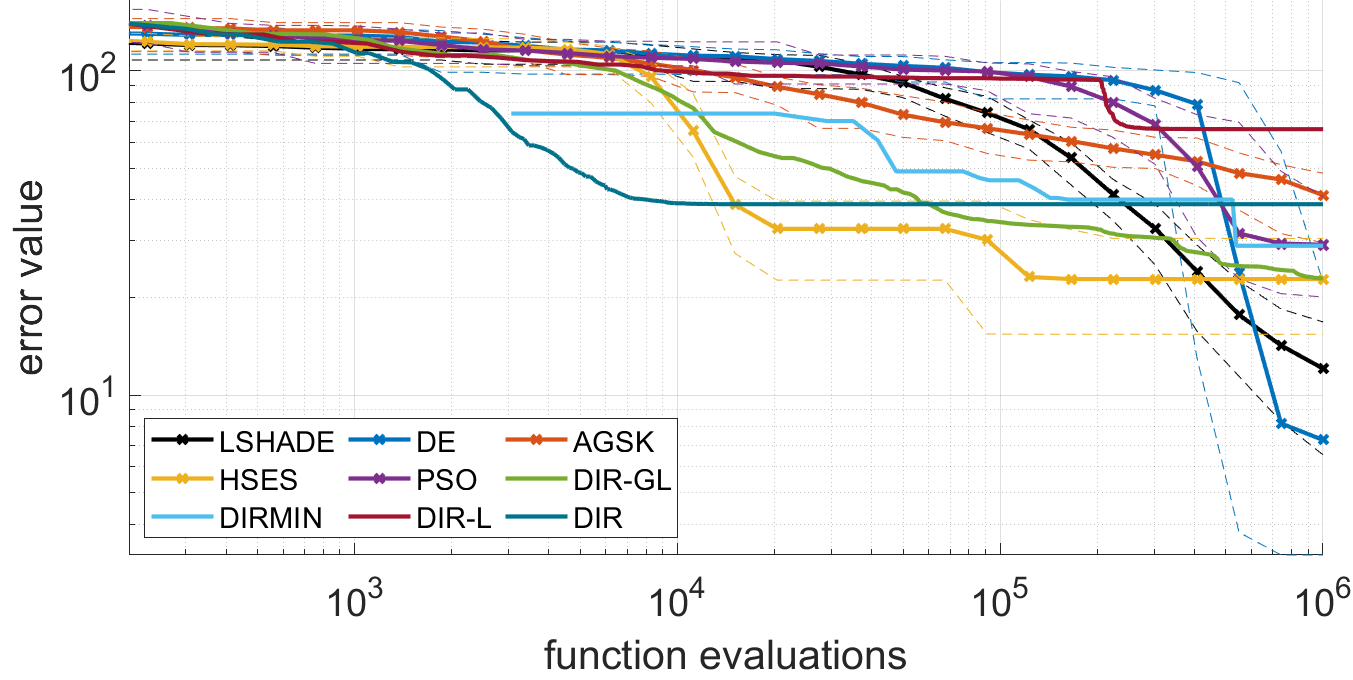} & \includegraphics[width = 0.45\linewidth]{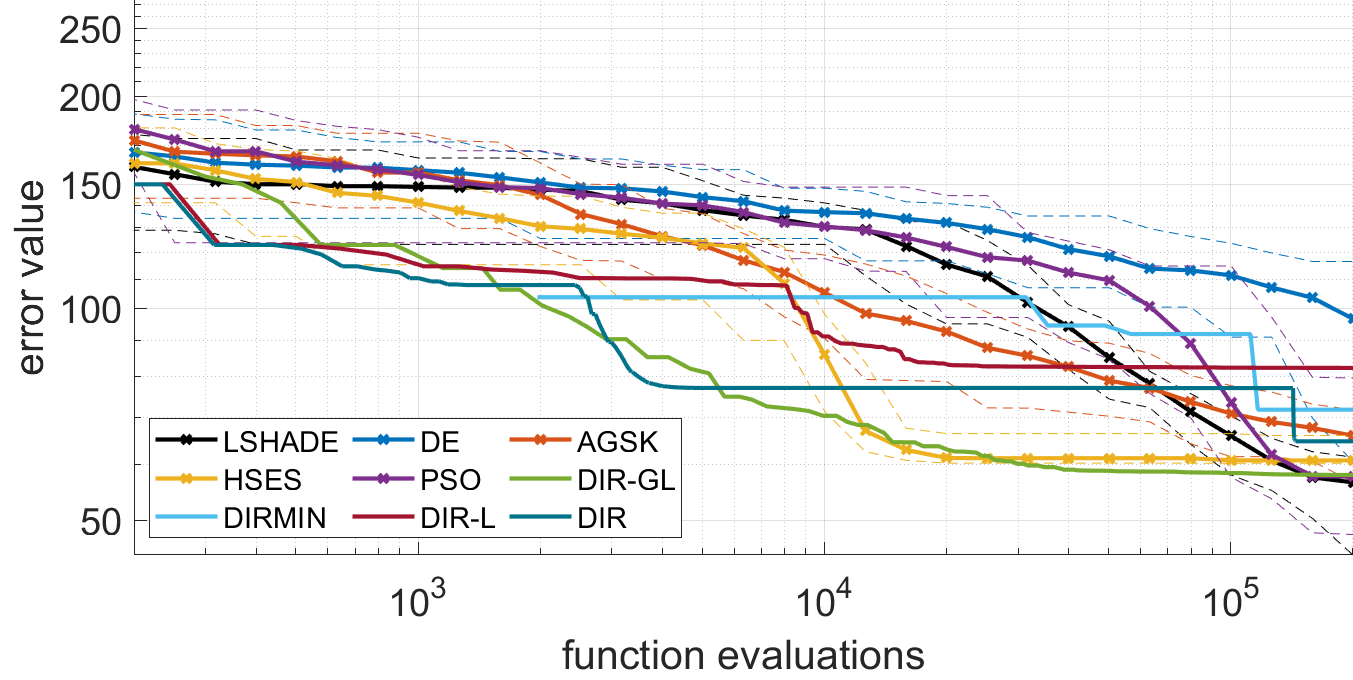} \\
        a) & b)
    \end{tabular}
    \caption{Convergence plots from the ambiguous benchmark set, a) F3 in $D=20$, b) F5 in $D=10$.}
    \label{conv_amb}
\end{figure}

In Table \ref{amb_res} are reported detailed statistics of the results of the computations, while representative convergence plots are shown in Figure \ref{conv_amb}. In the comparison of the best values achieved over the 30 runs of the nature-inspired methods against the DIRECT-type ones, the nature-inspired methods are a clear winner. However, the ranking between the nature-inspired methods is not exactly clear - DE, LSHADE and HSES all performed very well on all instances. The performance of PSO was a bit lacking, but more interestingly, AGSK, which performed very well on the previous benchmark sets, did extremely poorly on this one. This situation changes quite a bit when looking at the median values. Here, the nature-inspired methods still dominate, but not so strongly. Both LSHADE and HSES still come out as the overall best methods, but they get outperformed by DIR-GL on the F8 problem in all dimensions. DE still did very well on most instances, but was also the weakest of all methods on F1 in $D= 5$, F4 in $D=10$, F5 in $D = [5, 10]$, and F8 in $ D=20$. The best-performing of the DIRECT-type methods, DIR-GL, did almost as well as HSES. Lastly, when comparing against the worst over the 30 runs, the best-performing method remains LSHADE, but is closely followed by DIR-GL. Following these two are HSES, DIRMIN and DIR. The worst methods in this comparison are DE, DIR-L, PSO, and AGSK.

\begin{figure}[!h]
    \centering
    \begin{tabular}{cc}
        \includegraphics[width = 0.4\linewidth]{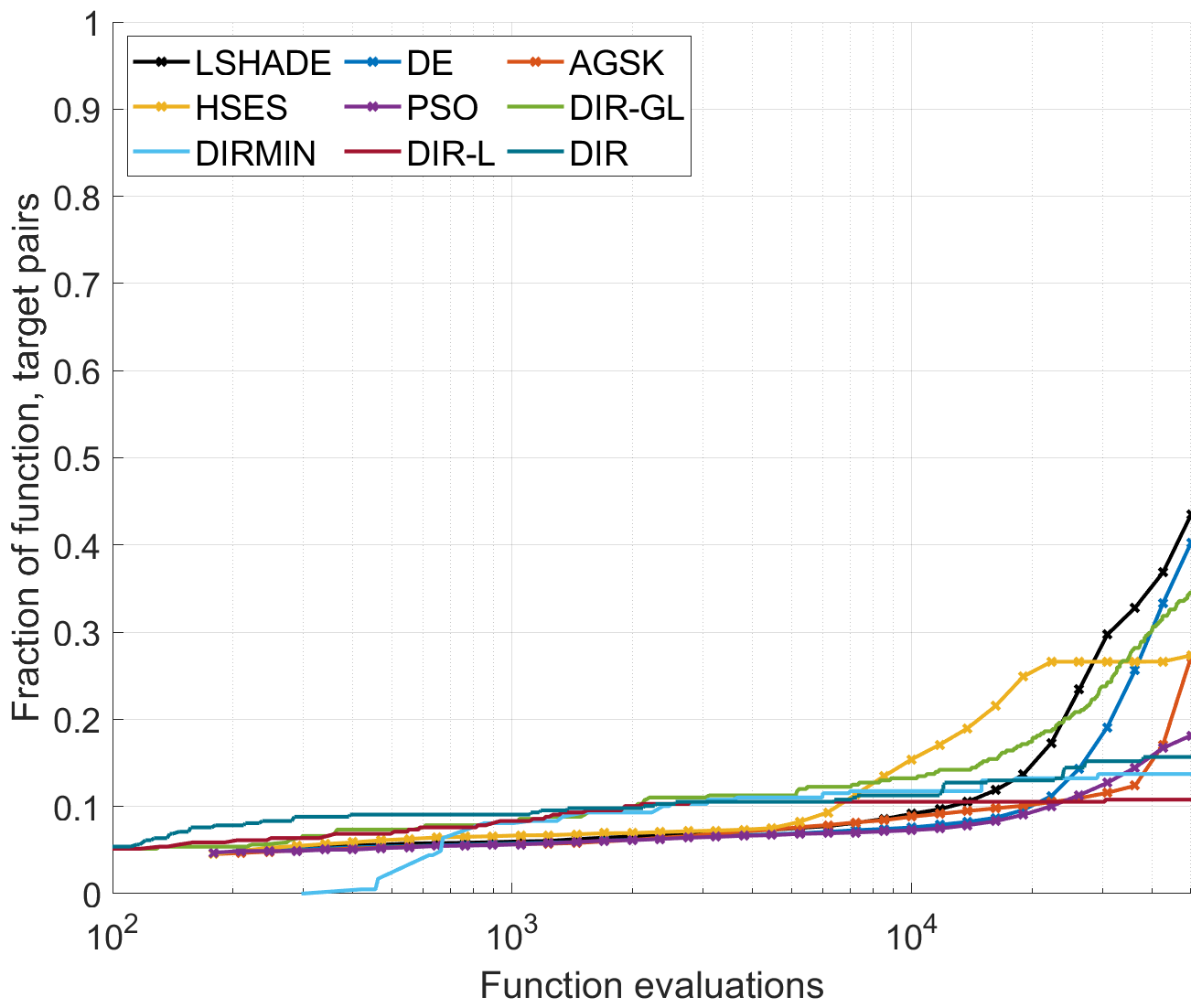} & \includegraphics[width = 0.4\linewidth]{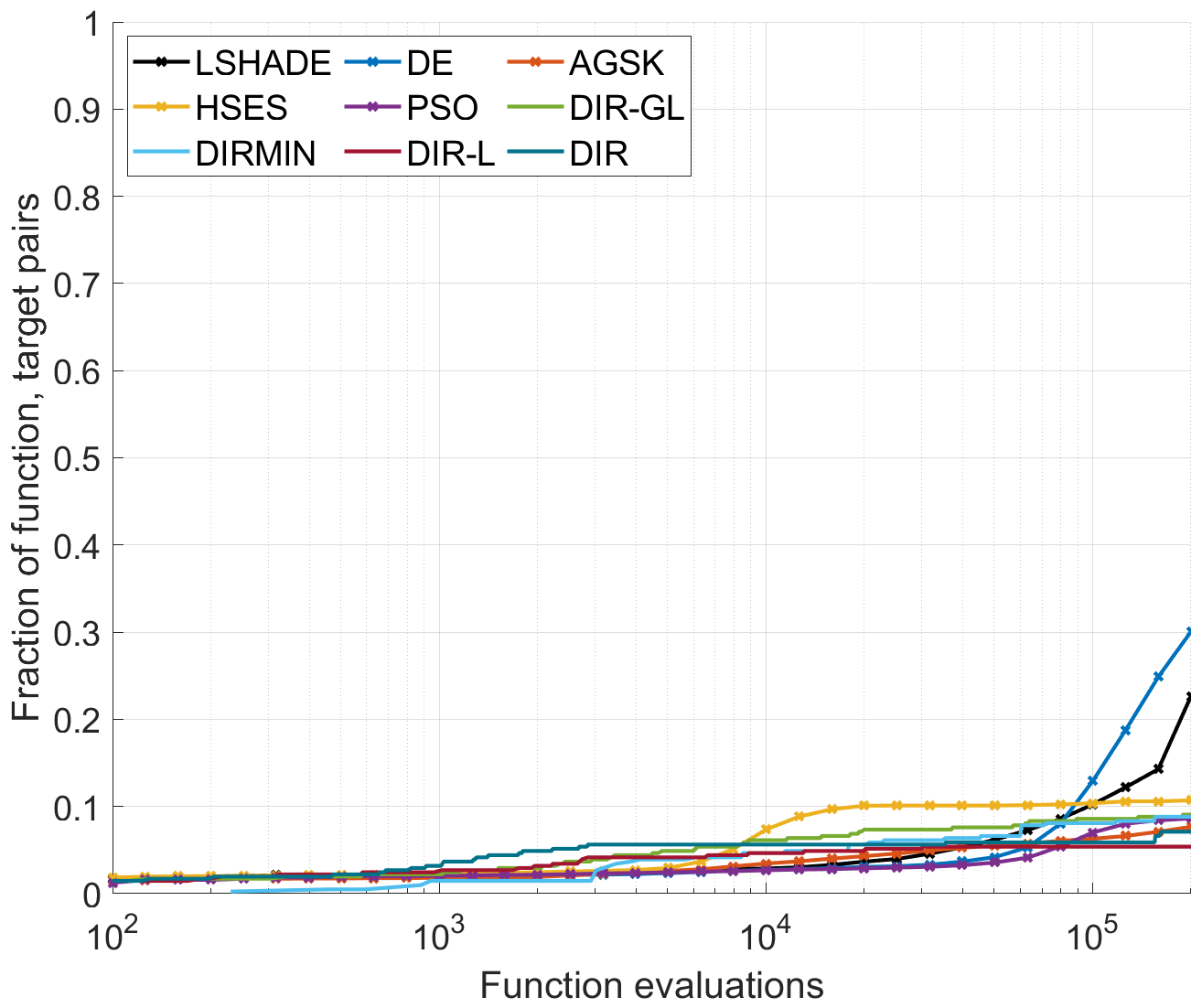} \\
        a) & b) \vspace{3mm}\\
        \includegraphics[width = 0.4\linewidth]{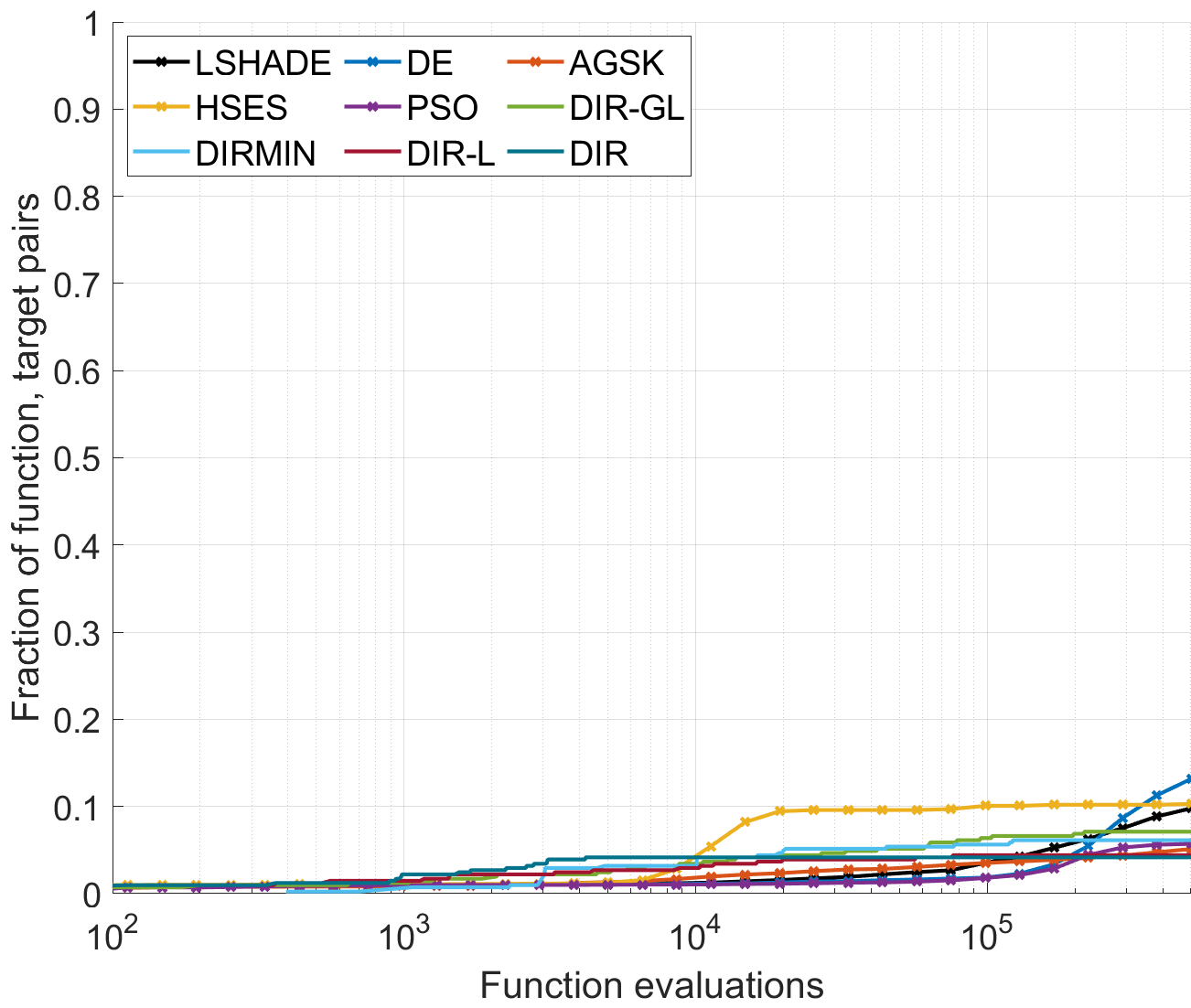} & \includegraphics[width = 0.4\linewidth]{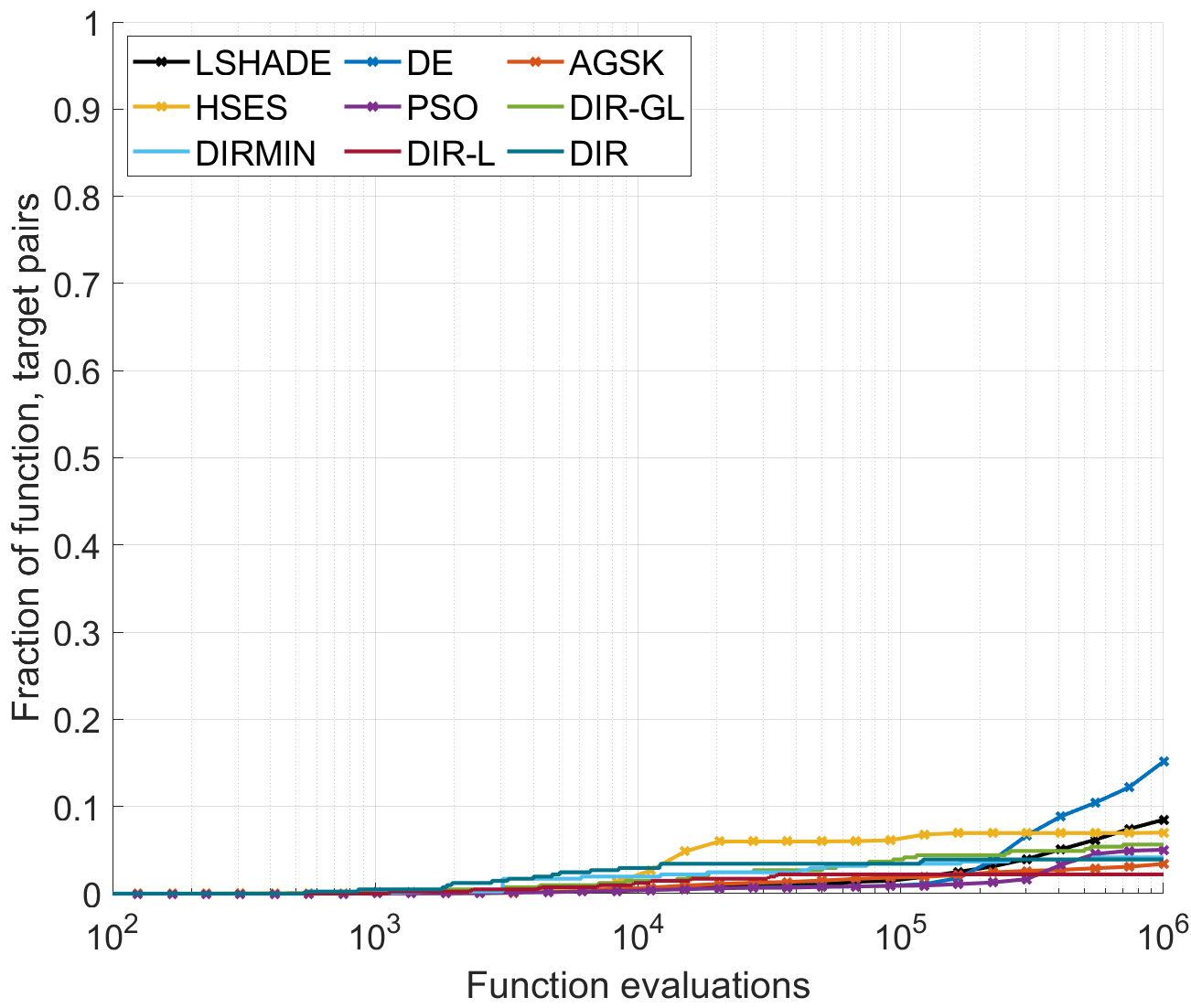} \\
        c) & d)
    \end{tabular}
    \caption{ECD of the number of objective function evaluations for different target precisions on the ambiguous benchmark set, a) $D=5$, b) $D=10$, c) $D=15$, d) $D=20$.} \label{fig_amb_ecd}
\end{figure}

The results of the ranking analysis are reported in Table \ref{abm_ranks} and the EDC plots are shown in Figure \ref{fig_amb_ecd}. In the early stages of the search, the DIRECT-type methods, and especially the original DIR, are clearly outperforming all the nature-inspired ones. In the medium stages (between 1/8 and 1/4 of MaxFES), both HSES and the DIRECT-type methods with some exploitation capabilities (DIR-GL and DIRMIN) performed the best. In the late stages of the search, LSHADE became the best method, followed by HSES and DIR-GL. On the whole, the ambiguous benchmark set seems to be the most difficult one of the five, with the lowest fraction of reached targets, even by the most successful methods.

\begin{table}[!h]
\caption{Mean ranks from Friedman tests at different stages of the search, ambiguous benchmark set.}
\centering
\resizebox{0.6\columnwidth}{!}{%
\setlength{\tabcolsep}{0.3em}
\bgroup
\def\arraystretch{1.1}
\begin{tabular}{l|ccccccccc}
FES            & \multicolumn{1}{c}{DE} & \multicolumn{1}{c}{LSHADE} & \multicolumn{1}{c}{HSES} & \multicolumn{1}{c}{PSO} &\multicolumn{1}{c}{AGSK} & \multicolumn{1}{c}{DIR-GL} & \multicolumn{1}{c}{DIRMIN} & \multicolumn{1}{c}{DIR-L} & \multicolumn{1}{c}{DIR} \\ \hline 
(1/256)*MaxFES & 7.1 & 6.0 & 5.2 & 6.7 & 7.3 & 4.5 & \textbf{3.1} & \textbf{3.5} & \textbf{1.7} \\
(1/128)*MaxFES & 7.2 & 6.7 & 4.9 & 7.0 & 7.0 & \textbf{3.6} & \textbf{3.6} & \textbf{3.5} & \textbf{1.4} \\
(1/32)*MaxFES  & 7.3 & 7.2 & \textbf{3.4} & 7.7 & 6.7 & \textbf{2.8} & 3.5 & 4.0 &\textbf{2.4} \\
(1/8)*MaxFES   & 7.4 & 6.2 & \textbf{2.3} & 8.4 & 6.2 & \textbf{2.1} & \textbf{3.5} & 5.1 & 3.8 \\
(1/4)*MaxFES   & 7.2 & 5.0 & \textbf{1.8} & 8.3 & 6.4 & \textbf{2.5} & \textbf{3.8} & 5.8 & 4.3 \\
(1/2)*MaxFES   & 6.2 & \textbf{3.2} & \textbf{2.3} & 7.0 & 6.8 & \textbf{2.6} & 4.4 & 7.1 & 5.5 \\
(3/4)*MaxFES   & 5.1 & \textbf{2.4} & \textbf{2.7} & 5.6 & 6.9 & \textbf{3.2} & 5.1 & 7.7 & 6.1 \\
MaxFES         & 3.8 & \textbf{2.3} & \textbf{3.0} & 5.5 & 6.4 & \textbf{3.7} & 5.7 & 8.0 & 6.7
\end{tabular} \label{abm_ranks}
\egroup
}
\end{table}

\begin{landscape}
\begin{table}[!h]
\caption{Detailed statistics of the 30 runs of the selected algorithms on the ambiguous benchmark set.}
\centering
%\scriptsize	
\resizebox{1.00\columnwidth}{!}{%
\setlength{\tabcolsep}{0.3em}
\bgroup
\def\arraystretch{1.3}
\begin{tabular}{cc|cccc|ccc|ccc|ccc|ccc|ccc}
\multicolumn{2}{l}{}& \multicolumn{4}{c}{Direct methods} & \multicolumn{3}{c}{DE} & \multicolumn{3}{c}{LSHADE} & \multicolumn{3}{c}{HSES} & \multicolumn{3}{c}{PSO} & \multicolumn{3}{c}{AGSK} \\
\multicolumn{1}{l}{} & D                    & DIR-GL               & DIRMIN               & DIR-L                & DIR                  & min                  & median   & max                            & min                  & median   & max                                  & min                  & median   & max                               & min                  & median   & max     & min                  & median   & max                               \\ \hline
\multirow{4}{*}{F1} & 5  & 1.59E-06 & 2.27E-01 & 1.29E+00 & 1.58E-01 & 3.52E-05 & 2.11E+00 & 3.45E+00 & 0.00E+00 & 7.89E-02 & 2.27E-01 & 0.00E+00 & 1.48E-01 & 2.27E-01 & 0.00E+00 & 4.50E-01 & 1.90E+00 & 8.85E-02 & 3.46E-01 & 5.61E-01 \\
 & 10 & 1.07E+00 & 1.70E+00 & 3.93E+00 & 6.56E+00 & 1.04E-07 & 3.06E-01 & 7.66E+00 & 1.33E-03 & 1.31E-01 & 3.21E-01 & 3.06E-01 & 4.98E-01 & 7.60E-01 & 3.06E-01 & 9.52E-01 & 3.26E+00 & 1.30E+00 & 2.13E+00 & 3.60E+00 \\
 & 15 & 2.58E+00 & 1.59E+00 & 3.57E+00 & 6.72E+00 & 2.27E-01 & 6.91E-01 & 1.94E+00 & 2.49E-01 & 4.22E-01 & 5.96E-01 & 7.89E-02 & 2.37E-01 & 4.64E-01 & 1.29E+00 & 2.77E+00 & 4.87E+00 & 2.53E+00 & 3.46E+00 & 4.50E+00 \\
 & 20 & 4.79E+00 & 3.42E+00 & 1.88E+01 & 5.88E+00 & 2.37E-01 & 8.79E-01 & 3.61E+00 & 2.69E-01 & 4.82E-01 & 6.36E-01 & 9.09E-01 & 1.10E+00 & 1.45E+00 & 7.00E-01 & 2.24E+00 & 3.71E+00 & 6.01E+00 & 8.20E+00 & 1.34E+01 \\ \hline
\multirow{4}{*}{F2} & 5  & 1.71E-07 & 3.44E+00 & 1.16E+01 & 1.07E-01 & 0.00E+00 & 0.00E+00 & 1.15E+00 & 0.00E+00 & 0.00E+00 & 0.00E+00 & 0.00E+00 & 2.16E+00 & 3.31E+00 & 0.00E+00 & 1.65E+00 & 5.29E+00 & 0.00E+00 & 0.00E+00 & 0.00E+00 \\
 & 10 & 4.63E+00 & 6.75E+00 & 1.74E+01 & 5.73E+00 & 0.00E+00 & 0.00E+00 & 2.30E+00 & 0.00E+00 & 0.00E+00 & 3.28E-04 & 4.46E+00 & 5.60E+00 & 6.75E+00 & 2.66E+00 & 5.74E+00 & 1.09E+01 & 1.15E+00 & 4.46E+00 & 7.46E+00 \\
 & 15 & 4.46E+00 & 5.47E+00 & 1.63E+01 & 1.84E+01 & 0.00E+00 & 1.15E+00 & 4.46E+00 & 9.76E-03 & 2.31E+00 & 3.47E+00 & 1.15E+00 & 2.30E+00 & 4.59E+00 & 4.59E+00 & 1.41E+01 & 2.76E+01 & 5.47E+00 & 1.10E+01 & 1.55E+01 \\
 & 20 & 2.45E+01 & 2.87E+01 & 5.26E+01 & 3.84E+01 & 0.00E+00 & 1.31E+00 & 1.12E+01 & 7.96E-03 & 1.75E+00 & 3.46E+00 & 1.21E+01 & 1.44E+01 & 1.65E+01 & 9.17E+00 & 1.74E+01 & 2.76E+01 & 1.42E+01 & 2.47E+01 & 2.98E+01 \\ \hline
\multirow{4}{*}{F3} & 5  & 1.89E+00 & 1.83E+00 & 2.94E+00 & 2.17E+00 & 0.00E+00 & 9.37E-04 & 7.99E+00 & 2.26E-07 & 2.04E-02 & 1.92E+00 & 0.00E+00 & 2.17E+00 & 3.77E+00 & 1.61E+00 & 4.97E+00 & 1.06E+01 & 0.00E+00 & 2.28E+00 & 4.69E+00 \\
 & 10 & 1.22E+01 & 5.22E+00 & 3.18E+01 & 1.63E+01 & 0.00E+00 & 2.05E+00 & 3.00E+01 & 1.33E-02 & 3.60E+00 & 7.73E+00 & 3.78E+00 & 8.25E+00 & 1.17E+01 & 3.21E+00 & 9.00E+00 & 1.86E+01 & 9.36E+00 & 1.29E+01 & 1.73E+01 \\
 & 15 & 1.57E+01 & 1.33E+01 & 3.67E+01 & 2.59E+01 & 0.00E+00 & 8.14E+00 & 6.45E+01 & 3.98E+00 & 8.06E+00 & 1.29E+01 & 6.52E+00 & 9.96E+00 & 1.47E+01 & 8.84E+00 & 2.23E+01 & 3.32E+01 & 1.54E+01 & 2.54E+01 & 2.92E+01 \\
 & 20 & 2.29E+01 & 2.89E+01 & 6.61E+01 & 3.88E+01 & 3.21E+00 & 7.27E+00 & 2.24E+01 & 6.54E+00 & 1.21E+01 & 1.68E+01 & 1.54E+01 & 2.28E+01 & 3.04E+01 & 2.01E+01 & 2.91E+01 & 4.09E+01 & 2.95E+01 & 4.14E+01 & 4.83E+01 \\ \hline
\multirow{4}{*}{F4} & 5  & 6.37E-01 & 4.70E-01 & 6.37E-01 & 6.45E-01 & 2.76E-08 & 1.54E+00 & 5.55E+00 & 0.00E+00 & 9.85E-03 & 8.60E-01 & 0.00E+00 & 6.37E-01 & 1.27E+00 & 0.00E+00 & 1.81E+00 & 9.86E+00 & 1.41E-06 & 1.38E+00 & 3.75E+00 \\
 & 10 & 4.49E+00 & 1.89E+01 & 1.59E+01 & 5.80E+00 & 0.00E+00 & 2.09E+01 & 3.09E+01 & 4.98E-01 & 1.55E+00 & 4.00E+00 & 1.27E+00 & 2.55E+00 & 3.82E+00 & 1.96E+00 & 7.98E+00 & 2.50E+01 & 7.42E+00 & 1.18E+01 & 1.61E+01 \\
 & 15 & 8.76E+00 & 3.44E+01 & 4.36E+01 & 2.89E+01 & 6.37E-01 & 3.12E+01 & 5.56E+01 & 1.60E+00 & 5.40E+00 & 1.24E+01 & 1.91E+00 & 3.46E+00 & 4.66E+00 & 6.60E+00 & 1.87E+01 & 3.60E+01 & 1.52E+01 & 2.61E+01 & 3.66E+01 \\
 & 20 & 1.43E+01 & 4.97E+01 & 5.52E+01 & 3.92E+01 & 1.11E+00 & 5.61E+00 & 8.22E+01 & 2.60E+00 & 1.89E+01 & 3.15E+01 & 4.25E+00 & 6.23E+00 & 9.20E+00 & 8.46E+00 & 2.91E+01 & 6.92E+01 & 3.55E+01 & 4.71E+01 & 5.30E+01 \\ \hline
\multirow{4}{*}{F5} & 5  & 2.79E+01 & 2.80E+01 & 3.03E+01 & 2.70E+01 & 2.57E+01 & 3.50E+01 & 4.12E+01 & 1.55E+01 & 2.41E+01 & 2.65E+01 & 1.70E+01 & 3.04E+01 & 3.30E+01 & 1.98E+01 & 2.92E+01 & 3.72E+01 & 2.18E+01 & 2.82E+01 & 3.12E+01 \\
 & 10 & 5.80E+01 & 7.18E+01 & 8.23E+01 & 6.47E+01 & 6.05E+01 & 9.66E+01 & 1.17E+02 & 4.46E+01 & 5.66E+01 & 6.14E+01 & 6.02E+01 & 6.08E+01 & 6.60E+01 & 4.77E+01 & 5.76E+01 & 7.96E+01 & 5.67E+01 & 6.59E+01 & 7.17E+01 \\
 & 15 & 8.78E+01 & 1.30E+02 & 1.04E+02 & 1.15E+02 & 7.70E+01 & 9.09E+01 & 1.41E+02 & 8.48E+01 & 9.37E+01 & 1.01E+02 & 8.17E+01 & 8.49E+01 & 9.00E+01 & 8.38E+01 & 9.66E+01 & 1.22E+02 & 1.01E+02 & 1.23E+02 & 1.37E+02 \\
 & 20 & 1.22E+02 & 2.00E+02 & 1.94E+02 & 1.69E+02 & 1.09E+02 & 1.21E+02 & 1.38E+02 & 1.21E+02 & 1.33E+02 & 1.45E+02 & 1.02E+02 & 1.12E+02 & 1.22E+02 & 1.26E+02 & 1.42E+02 & 1.74E+02 & 1.55E+02 & 1.74E+02 & 1.84E+02 \\ \hline
\multirow{4}{*}{F6} & 5  & 9.34E-05 & 5.14E+00 & 2.22E+01 & 3.93E+00 & 0.00E+00 & 0.00E+00 & 1.30E-08 & 0.00E+00 & 0.00E+00 & 3.55E+00 & 0.00E+00 & 3.55E+00 & 3.89E+00 & 0.00E+00 & 6.84E+00 & 1.66E+01 & 0.00E+00 & 7.67E+00 & 1.41E+01 \\
 & 10 & 2.15E+01 & 3.67E+00 & 5.87E+01 & 3.58E+01 & 0.00E+00 & 0.00E+00 & 3.55E+00 & 1.54E+00 & 1.23E+01 & 2.10E+01 & 3.55E+00 & 7.10E+00 & 9.03E+00 & 1.35E+01 & 3.18E+01 & 4.37E+01 & 2.70E+01 & 3.89E+01 & 5.10E+01 \\
 & 15 & 4.07E+01 & 6.38E+01 & 8.38E+01 & 8.61E+01 & 0.00E+00 & 1.07E+01 & 1.56E+01 & 1.00E+01 & 3.05E+01 & 4.41E+01 & 1.45E+01 & 1.89E+01 & 2.64E+01 & 2.76E+01 & 5.97E+01 & 1.05E+02 & 5.64E+01 & 9.34E+01 & 1.11E+02 \\
 & 20 & 4.80E+01 & 7.37E+01 & 2.49E+02 & 1.25E+02 & 0.00E+00 & 1.08E+01 & 2.39E+01 & 3.07E+01 & 5.54E+01 & 7.23E+01 & 2.13E+01 & 2.84E+01 & 3.67E+01 & 4.76E+01 & 1.02E+02 & 1.25E+02 & 1.06E+02 & 1.40E+02 & 1.58E+02 \\ \hline
\multirow{4}{*}{F7} & 5  & 5.86E+01 & 1.18E+02 & 1.16E+02 & 1.29E+02 & 4.18E+00 & 3.55E+01 & 7.85E+01 & 0.00E+00 & 5.92E+01 & 7.74E+01 & 5.54E-02 & 2.37E-01 & 7.56E-01 & 5.42E+01 & 9.52E+01 & 1.31E+02 & 8.04E+01 & 1.11E+02 & 1.35E+02 \\
 & 10 & 1.98E+02 & 2.92E+02 & 2.97E+02 & 3.02E+02 & 1.51E+02 & 1.91E+02 & 2.24E+02 & 1.17E+02 & 1.58E+02 & 1.79E+02 & 1.58E+02 & 1.83E+02 & 2.38E+02 & 2.23E+02 & 2.67E+02 & 3.37E+02 & 2.20E+02 & 3.17E+02 & 3.32E+02 \\
 & 15 & 3.55E+02 & 5.00E+02 & 5.20E+02 & 4.95E+02 & 1.28E+02 & 3.14E+02 & 3.78E+02 & 2.06E+02 & 2.52E+02 & 2.80E+02 & 3.55E+02 & 3.92E+02 & 4.07E+02 & 3.52E+02 & 4.29E+02 & 5.04E+02 & 3.32E+02 & 5.08E+02 & 5.35E+02 \\
 & 20 & 4.99E+02 & 6.72E+02 & 7.32E+02 & 7.22E+02 & 0.00E+00 & 2.37E+02 & 5.21E+02 & 2.63E+02 & 3.29E+02 & 3.65E+02 & 1.25E+02 & 5.28E+02 & 5.60E+02 & 5.62E+02 & 6.34E+02 & 6.98E+02 & 4.62E+02 & 7.16E+02 & 7.33E+02 \\ \hline
\multirow{4}{*}{F8} & 5  & 2.00E+01 & 3.08E+01 & 3.77E+01 & 3.34E+01 & 2.77E+01 & 3.45E+01 & 4.11E+01 & 1.78E+01 & 2.07E+01 & 2.61E+01 & 1.98E+01 & 3.58E+01 & 4.93E+01 & 1.86E+01 & 3.63E+01 & 4.56E+01 & 2.17E+01 & 3.28E+01 & 4.17E+01 \\
 & 10 & 5.48E+01 & 7.87E+01 & 1.30E+02 & 9.44E+01 & 1.03E+02 & 1.15E+02 & 1.30E+02 & 4.92E+01 & 5.86E+01 & 6.87E+01 & 4.10E+01 & 5.58E+01 & 6.89E+01 & 5.67E+01 & 1.01E+02 & 1.18E+02 & 7.76E+01 & 9.21E+01 & 1.02E+02 \\
 & 15 & 8.61E+01 & 1.22E+02 & 2.43E+02 & 1.51E+02 & 1.75E+02 & 2.04E+02 & 2.17E+02 & 8.64E+01 & 9.97E+01 & 1.07E+02 & 6.27E+01 & 8.97E+01 & 1.14E+02 & 1.12E+02 & 1.85E+02 & 2.09E+02 & 1.37E+02 & 1.61E+02 & 1.77E+02 \\
 & 20 & 9.82E+01 & 1.72E+02 & 2.32E+02 & 2.19E+02 & 2.56E+02 & 2.85E+02 & 2.99E+02 & 1.24E+02 & 1.52E+02 & 1.73E+02 & 8.62E+01 & 1.22E+02 & 1.50E+02 & 2.00E+02 & 2.74E+02 & 3.07E+02 & 2.15E+02 & 2.47E+02 & 2.73E+02
\end{tabular} \label{amb_res}
\egroup
}
\end{table}
\end{landscape}

% \begin{landscape}
% \begin{table}[!h]
% \caption{Detailed statistics of the 30 runs of the selected algorithms on the ambiguous benchmark set.}
% \centering
% %\scriptsize	
% \includegraphics[width = \linewidth]{tab7.png}
%  \label{amb_res}
% \end{table}
% \end{landscape}

\subsection{Comparison of the Results}
The results of the computational experiments on the five benchmark sets can be read in a number of ways. The first thing one notices is that when the objective function evaluations are cheap, the best strategy for optimizing black-box functions is using some the of the well-performing methods from the CEC competitions, in our case LSHADE, and performing multiple runs of the algorithm. However, betting on a single method may prove to be a poor decision, as for instance AGSK displayed a very bad performance on the ambiguous benchmark set. In this setting, the DIRECT-type methods did very poorly, as their deterministic nature, relatively higher time complexity, and lack of exploitation capabilities hindered their performance. 

On the other hand, when the objective function evaluations are costly, or the maximum number of function evaluations is highly limited, the DIRECT-type methods might offer an advantage over their nature-inspired counterparts. This effect was even more pronounced on the BBOB, DIRECTLib, and ambiguous benchmark sets than on the CEC ones. Another interesting observation can be made regarding the convergence properties of HSES. On all benchmark sets it performed relatively well, but it was especially effective during earlier stages of the search. This behaviour of HSES and the DIRECT-type methods might provide fertile ground for new hybridization \cite{hadi2021single} or composite \cite{kudela2022composite} schemes, which could combine these ``early-stage'' methods with methods that have better exploitation capabilities.

Interestingly, the rank statistics over the two CEC sets are very similar. Both the these benchmark sets seem to favor the DE variants (DE and LSHADE) and the AGSK algorithm over the other nature-inspired methods. This affirms the sentiment that for benchmarking purposes the use of heterogenous benchmark sets from different sources should be preferred \cite{piotrowski2015regarding}.

\section{Conclusions}
To answer the main question: for black-box problems, metaheuristics (at least the nature-inspired ones used in the computational study) are definitely worth it. As shown in the computational experiments, especially in situation when the objective function evaluations are cheap and the number of possible runs of the method is high, well-performing methods from the CEC competitions, such as LSHADE and HSES, consistently outperform the best deterministic DIRECT-type methods on difficult benchmark instances. On the other hand, if the evaluations are costly, or the number of evaluations is limited, the DIRECT-type methods showed better performance and their deterministic nature provided more stable results. 

The analyses of the behaviours of the different methods also uncovered possibilities for new hybrid methods, that could utilize the high exploration capabilities of either the DIRECT-type methods or the HSES algorithm with methods that possess better exploitation routines. Even though the number of considered algorithms and test functions was limited, we believe the performed analyses will carry over to a large portion of real-world optimization problems.

Similar studies could be made for comparing deterministic and nature-inspired methods on problems with general or hidden constraints. This will be the subject of our further research.

\section*{Acknowledgements}
This work was supported by the Grant Agency of the Czech Republic project no. 22-31173S ``Adaptive soft computing framework for inverse heat transfer problems with phase changes'', and IGA Brno University of Technology: No. FSI-S-20-6538.

\bibliography{mybib.bib}

\begin{thebibliography}{10}
\expandafter\ifx\csname url\endcsname\relax
  \def\url#1{\texttt{#1}}\fi
\expandafter\ifx\csname urlprefix\endcsname\relax\def\urlprefix{URL }\fi
\expandafter\ifx\csname href\endcsname\relax
  \def\href#1#2{#2} \def\path#1{#1}\fi

\bibitem{rios2013derivative}
L.~M. Rios, N.~V. Sahinidis, Derivative-free optimization: a review of
  algorithms and comparison of software implementations, Journal of Global
  Optimization 56~(3) (2013) 1247--1293.

\bibitem{hellwig2019benchmarking}
M.~Hellwig, H.-G. Beyer, Benchmarking evolutionary algorithms for single
  objective real-valued constrained optimization--a critical review, Swarm and
  evolutionary computation 44 (2019) 927--944.

\bibitem{rardin2001experimental}
R.~L. Rardin, R.~Uzsoy, Experimental evaluation of heuristic optimization
  algorithms: A tutorial, Journal of Heuristics 7~(3) (2001) 261--304.

\bibitem{pinter2013global}
J.~D. Pint{\'e}r, Global optimization in action: continuous and Lipschitz
  optimization: algorithms, implementations and applications, Vol.~6, Springer
  Science \& Business Media, 2013.

\bibitem{jones1993lipschitzian}
D.~R. Jones, C.~D. Perttunen, B.~E. Stuckman, Lipschitzian optimization without
  the lipschitz constant, Journal of optimization Theory and Applications
  79~(1) (1993) 157--181.

\bibitem{xiao2019evaluation}
Y.~Xiao, H.~Rivaz, M.~Chabanas, M.~Fortin, I.~Machado, Y.~Ou, M.~P. Heinrich,
  J.~A. Schnabel, X.~Zhong, A.~Maier, et~al., Evaluation of mri to ultrasound
  registration methods for brain shift correction: the curious2018 challenge,
  IEEE transactions on medical imaging 39~(3) (2019) 777--786.

\bibitem{ljungberg2004simultaneous}
K.~Ljungberg, S.~Holmgren, {\"O}.~Carlborg, Simultaneous search for multiple
  qtl using the global optimization algorithm direct, Bioinformatics 20~(12)
  (2004) 1887--1895.

\bibitem{campana2016derivative}
E.~F. Campana, M.~Diez, U.~Iemma, G.~Liuzzi, S.~Lucidi, F.~Rinaldi, A.~Serani,
  Derivative-free global ship design optimization using global/local
  hybridization of the direct algorithm, Optimization and Engineering 17~(1)
  (2016) 127--156.

\bibitem{nguyen2010global}
T.~L. Nguyen, K.-S. Low, A global maximum power point tracking scheme employing
  direct search algorithm for photovoltaic systems, IEEE transactions on
  Industrial Electronics 57~(10) (2010) 3456--3467.

\bibitem{sergeyev2006global}
Y.~D. Sergeyev, D.~E. Kvasov, Global search based on efficient diagonal
  partitions and a set of lipschitz constants, SIAM Journal on Optimization
  16~(3) (2006) 910--937.

\bibitem{jones2021direct}
D.~R. Jones, J.~R. Martins, The {DIRECT} algorithm: 25 years later, Journal of
  Global Optimization 79~(3) (2021) 521--566.

\bibitem{molina2020comprehensive}
D.~Molina, J.~Poyatos, J.~D. Ser, S.~Garc{\'\i}a, A.~Hussain, F.~Herrera,
  Comprehensive taxonomies of nature-and bio-inspired optimization: Inspiration
  versus algorithmic behavior, critical analysis recommendations, Cognitive
  Computation 12~(5) (2020) 897--939.

\bibitem{yang2013swarm}
X.-S. Yang, Z.~Cui, R.~Xiao, A.~H. Gandomi, M.~Karamanoglu, Swarm intelligence
  and bio-inspired computation: theory and applications, Newnes, 2013.

\bibitem{young2015optimizing}
S.~R. Young, D.~C. Rose, T.~P. Karnowski, S.-H. Lim, R.~M. Patton, Optimizing
  deep learning hyper-parameters through an evolutionary algorithm, in:
  Proceedings of the workshop on machine learning in high-performance computing
  environments, 2015, pp. 1--5.

\bibitem{vzufan2021advances}
P.~{\v{Z}}ufan, M.~Bidlo, Advances in evolutionary optimization of quantum
  operators, Mendel 27~(2) (2021) 12--22.

\bibitem{bujok2019comparison}
P.~Bujok, J.~Tvrd{\'\i}k, R.~Pol{\'a}kov{\'a}, Comparison of nature-inspired
  population-based algorithms on continuous optimisation problems, Swarm and
  Evolutionary Computation 50 (2019) 100490.

\bibitem{campelo2021sharks}
F.~Campelo, C.~Aranha, Sharks, zombies and volleyball: Lessons from the
  evolutionary computation bestiary, Lifelike Computing Systems Workshop 2021
  (2021).

\bibitem{kudela2022commentary}
J.~Kudela, Commentary on:“stoa: A bio-inspired based optimization algorithm
  for industrial engineering problems”[eaai, 82 (2019), 148--174] and
  “tunicate swarm algorithm: A new bio-inspired based metaheuristic paradigm
  for global optimization”[eaai, 90 (2020), no. 103541], Engineering
  Applications of Artificial Intelligence 113 (2022) 104930.

\bibitem{kudelanature}
J.~Kudela, A critical problem in benchmarking and analysis of evolutionary
  computation methods, Nature Machine Intelligence December (2022).
\newblock \href {https://doi.org/10.1038/s42256-022-00579-0}
  {\path{doi:10.1038/s42256-022-00579-0}}.

\bibitem{cec22}
A.~Kumar, K.~Price, A.~Mohamed, A.~Hadi, P.~Suganthan, Problem definitions and
  evaluation criteria for the {CEC} 2022 special session and competition on
  single objective bound constrained numerical optimization, Technical Report,
  {Indian Institute of Technology}, India (December 2021).

\bibitem{ali2005numerical}
M.~M. Ali, C.~Khompatraporn, Z.~B. Zabinsky, A numerical evaluation of several
  stochastic algorithms on selected continuous global optimization test
  problems, Journal of global optimization 31~(4) (2005) 635--672.

\bibitem{more2009benchmarking}
J.~J. Mor{\'e}, S.~M. Wild, Benchmarking derivative-free optimization
  algorithms, SIAM Journal on Optimization 20~(1) (2009) 172--191.

\bibitem{gould2003cuter}
N.~I. Gould, D.~Orban, P.~L. Toint, Cuter and sifdec: A constrained and
  unconstrained testing environment, revisited, ACM Transactions on
  Mathematical Software (TOMS) 29~(4) (2003) 373--394.

\bibitem{globallib}
Global library, \url{http://www.gamsworld.org/global/globallib.htm} (Accessed
  30th June, 2022).

\bibitem{lukvsan2000test}
L.~Luk{\v{s}}an, J.~Vlcek, Test problems for nonsmooth unconstrained and
  linearly constrained optimization, Tech. rep., Technical report, Institute of
  Computer Science, Academy of Sciences of the Czech Republic (2000).

\bibitem{sergeyev2018efficiency}
Y.~D. Sergeyev, D.~Kvasov, M.~Mukhametzhanov, On the efficiency of
  nature-inspired metaheuristics in expensive global optimization with limited
  budget, Scientific reports 8~(1) (2018) 1--9.

\bibitem{gaviano2003algorithm}
M.~Gaviano, D.~E. Kvasov, D.~Lera, Y.~D. Sergeyev, Algorithm 829: Software for
  generation of classes of test functions with known local and global minima
  for global optimization, ACM Transactions on Mathematical Software (TOMS)
  29~(4) (2003) 469--480.

\bibitem{latorre2021prescription}
A.~LaTorre, D.~Molina, E.~Osaba, J.~Poyatos, J.~Del~Ser, F.~Herrera, A
  prescription of methodological guidelines for comparing bio-inspired
  optimization algorithms, Swarm and Evolutionary Computation 67 (2021) 100973.

\bibitem{camacho2019intelligent}
C.~L. Camacho-Villal{\'o}n, M.~Dorigo, T.~St{\"u}tzle, The intelligent water
  drops algorithm: why it cannot be considered a novel algorithm, Swarm
  Intelligence 13~(3) (2019) 173--192.

\bibitem{camacho2020grey}
C.~L. Camacho~Villal{\'o}n, T.~St{\"u}tzle, M.~Dorigo, Grey wolf, firefly and
  bat algorithms: Three widespread algorithms that do not contain any novelty,
  in: International conference on swarm intelligence, Springer, 2020, pp.
  121--133.

\bibitem{villalon2021cuckoo}
C.~Villal{\'o}n, T.~St{\"u}tzle, M.~Dorigo, Cuckoo search$\equiv$($\mu$+
  $\lambda$)--evolution strategy, in: IRIDIA--Technical Report Series, 2021.

\bibitem{del2021more}
J.~Del~Ser, E.~Osaba, A.~D. Martinez, M.~N. Bilbao, J.~Poyatos, D.~Molina,
  F.~Herrera, More is not always better: Insights from a massive comparison of
  meta-heuristic algorithms over real-parameter optimization problems, in: 2021
  IEEE Symposium Series on Computational Intelligence (SSCI), IEEE, 2021, pp.
  1--7.

\bibitem{kennedy1995particle}
J.~Kennedy, R.~Eberhart, Particle swarm optimization, in: Proceedings of
  ICNN'95-international conference on neural networks, Vol.~4, IEEE, 1995, pp.
  1942--1948.

\bibitem{pradhan2020novel}
A.~Pradhan, S.~K. Bisoy, A novel load balancing technique for cloud computing
  platform based on pso, Journal of King Saud University-Computer and
  Information Sciences (2020).

\bibitem{pradhan2021survey}
A.~Pradhan, S.~K. Bisoy, A.~Das, A survey on pso based meta-heuristic
  scheduling mechanism in cloud computing environment, Journal of King Saud
  University-Computer and Information Sciences (2021).

\bibitem{song2021improved}
B.~Song, Z.~Wang, L.~Zou, An improved pso algorithm for smooth path planning of
  mobile robots using continuous high-degree bezier curve, Applied Soft
  Computing 100 (2021) 106960.

\bibitem{kanagaraj2022meta}
G.~Kanagaraj, S.~S. Masthan, F.~Y. Vincent, Meta-heuristics based inverse
  kinematics of robot manipulator’s path tracking capability under joint
  limits, Mendel 28~(1) (2022) 41--54.

\bibitem{muller2022improving}
J.~Muller, Improving initial aerofoil geometry using aerofoil particle swarm
  optimisation, Mendel 28~(1) (2022) 63--67.

\bibitem{sennan2021t2fl}
S.~Sennan, S.~Ramasubbareddy, S.~Balasubramaniyam, A.~Nayyar, M.~Abouhawwash,
  N.~A. Hikal, T2fl-pso: Type-2 fuzzy logic-based particle swarm optimization
  algorithm used to maximize the lifetime of internet of things, IEEE Access 9
  (2021) 63966--63979.

\bibitem{lukasik2009firefly}
S.~{\L}ukasik, S.~{\.Z}ak, Firefly algorithm for continuous constrained
  optimization tasks, in: International conference on computational collective
  intelligence, Springer, 2009, pp. 97--106.

\bibitem{karaboga2007powerful}
D.~Karaboga, B.~Basturk, A powerful and efficient algorithm for numerical
  function optimization: artificial bee colony (abc) algorithm, Journal of
  global optimization 39~(3) (2007) 459--471.

\bibitem{shi1998modified}
Y.~Shi, R.~Eberhart, A modified particle swarm optimizer, in: 1998 IEEE
  international conference on evolutionary computation proceedings. IEEE world
  congress on computational intelligence (Cat. No. 98TH8360), IEEE, 1998, pp.
  69--73.

\bibitem{storn1997differential}
R.~Storn, K.~Price, Differential evolution--a simple and efficient heuristic
  for global optimization over continuous spaces, Journal of global
  optimization 11~(4) (1997) 341--359.

\bibitem{huynh2021q}
T.~N. Huynh, D.~T. Do, J.~Lee, Q-learning-based parameter control in
  differential evolution for structural optimization, Applied Soft Computing
  107 (2021) 107464.

\bibitem{li2020enhanced}
S.~Li, Q.~Gu, W.~Gong, B.~Ning, An enhanced adaptive differential evolution
  algorithm for parameter extraction of photovoltaic models, Energy Conversion
  and Management 205 (2020) 112443.

\bibitem{deng2020improved}
W.~Deng, H.~Liu, J.~Xu, H.~Zhao, Y.~Song, An improved quantum-inspired
  differential evolution algorithm for deep belief network, IEEE Transactions
  on Instrumentation and Measurement 69~(10) (2020) 7319--7327.

\bibitem{tanabe2013success}
R.~Tanabe, A.~Fukunaga, Success-history based parameter adaptation for
  differential evolution, in: 2013 IEEE congress on evolutionary computation,
  IEEE, 2013, pp. 71--78.

\bibitem{tanabe2014improving}
R.~Tanabe, A.~S. Fukunaga, Improving the search performance of shade using
  linear population size reduction, in: 2014 IEEE congress on evolutionary
  computation (CEC), IEEE, 2014, pp. 1658--1665.

\bibitem{brest2017single}
J.~Brest, M.~S. Mau{\v{c}}ec, B.~Bo{\v{s}}kovi{\'c}, Single objective
  real-parameter optimization: Algorithm j{SO}, in: 2017 IEEE congress on
  evolutionary computation (CEC), IEEE, 2017, pp. 1311--1318.

\bibitem{kochenderfer2019algorithms}
M.~J. Kochenderfer, T.~A. Wheeler, Algorithms for optimization, {MIT} Press,
  2019.

\bibitem{kudela2022new}
J.~Kudela, R.~Matousek, New benchmark functions for single-objective
  optimization based on a zigzag pattern, IEEE Access (2022).

\bibitem{Wagdy2020}
A.~W. Mohamed, A.~A. Hadi, A.~K. Mohamed, N.~H. Awad, Evaluating the
  performance of adaptive gainingsharing knowledge based algorithm on cec 2020
  benchmark problems, in: 2020 IEEE Congress on Evolutionary Computation (CEC),
  2020, pp. 1--8.
\newblock \href {https://doi.org/10.1109/CEC48606.2020.9185901}
  {\path{doi:10.1109/CEC48606.2020.9185901}}.

\bibitem{Wagdy2020b}
A.~W. Mohamed, A.~A. Hadi, A.~K. Mohamed, Gaining-sharing knowledge based
  algorithm for solving optimization problems: a novel nature-inspired
  algorithm, International Journal of Machine Learning and Cybernetics 11
  (2020) 1501--1529.

\bibitem{8477908}
G.~Zhang, Y.~Shi, Hybrid sampling evolution strategy for solving single
  objective bound constrained problems, in: 2018 IEEE Congress on Evolutionary
  Computation (CEC), 2018, pp. 1--7.

\bibitem{hansen2001completely}
N.~Hansen, A.~Ostermeier, Completely derandomized self-adaptation in evolution
  strategies, Evolutionary computation 9~(2) (2001) 159--195.

\bibitem{hauschild2011introduction}
M.~Hauschild, M.~Pelikan, An introduction and survey of estimation of
  distribution algorithms, Swarm and evolutionary computation 1~(3) (2011)
  111--128.

\bibitem{kazikova2020tuning}
A.~Kazikova, M.~Pluhacek, R.~Senkerik, Why tuning the control parameters of
  metaheuristic algorithms is so important for fair comparison?, Mendel 26~(2)
  (2020) 9--16.

\bibitem{matousek2022start}
R.~Matousek, L.~Dobrovsky, J.~Kudela, How to start a heuristic? utilizing lower
  bounds for solving the quadratic assignment problem, International Journal of
  Industrial Engineering Computations 13~(2) (2022) 151--164.

\bibitem{stripinis2021dgo}
L.~Stripinis, R.~Paulavičius, {DIRECTGO}: A new {DIRECT}-type {MATLAB} toolbox
  for derivative-free global optimization (2021).
\newblock \href {http://arxiv.org/abs/2107.02205} {\path{arXiv:2107.02205}}.

\bibitem{gablonsky2001locally}
J.~M. Gablonsky, C.~T. Kelley, A locally-biased form of the direct algorithm,
  Journal of Global Optimization 21~(1) (2001) 27--37.

\bibitem{liuzzi2010direct}
G.~Liuzzi, S.~Lucidi, V.~Piccialli, A direct-based approach exploiting local
  minimizations for the solution of large-scale global optimization problems,
  Computational Optimization and Applications 45~(2) (2010) 353--375.

\bibitem{stripinis2018improved}
L.~Stripinis, R.~Paulavi{\v{c}}ius, J.~{\v{Z}}ilinskas, Improved scheme for
  selection of potentially optimal hyper-rectangles in direct, Optimization
  Letters 12~(7) (2018) 1699--1712.

\bibitem{wu2017problem}
G.~Wu, R.~Mallipeddi, P.~N. Suganthan, Problem definitions and evaluation
  criteria for the cec 2017 competition on constrained real-parameter
  optimization, National University of Defense Technology, Changsha, Hunan, PR
  China and Kyungpook National University, Daegu, South Korea and Nanyang
  Technological University, Singapore, Technical Report (2017).

\bibitem{dirlib}
L.~Stripinis, R.~Paulavičius, Directlib – a library of global optimization
  problems for direct-type methods, v1.2,
  \url{https://doi.org/10.5281/zenodo.3948890} (Accessed 30th June, 2022)
  (2020).

\bibitem{hansen2021coco}
N.~Hansen, A.~Auger, R.~Ros, O.~Mersmann, T.~Tu{\v s}ar, D.~Brockhoff, {COCO}:
  A platform for comparing continuous optimizers in a black-box setting,
  Optimization Methods and Software 36 (2021) 114--144.
\newblock \href {https://doi.org/https://doi.org/10.1080/10556788.2020.1808977}
  {\path{doi:https://doi.org/10.1080/10556788.2020.1808977}}.

\bibitem{9504720}
J.~Kudela, Novel zigzag-based benchmark functions for bound constrained single
  objective optimization, in: 2021 IEEE Congress on Evolutionary Computation
  (CEC), 2021, pp. 857--862.

\bibitem{hadi2021single}
A.~A. Hadi, A.~W. Mohamed, K.~M. Jambi, Single-objective real-parameter
  optimization: Enhanced lshade-spacma algorithm, in: Heuristics for
  optimization and learning, Springer, 2021, pp. 103--121.

\bibitem{kudela2022composite}
J.~Kudela, T.~Nevoral, T.~Holoubek, Composite evolutionary strategy and
  differential evolution method for the icsi’2022 competition, in:
  International Conference on Sensing and Imaging, Springer, 2022, pp.
  432--439.

\bibitem{piotrowski2015regarding}
A.~P. Piotrowski, Regarding the rankings of optimization heuristics based on
  artificially-constructed benchmark functions, Information Sciences 297 (2015)
  191--201.

\end{thebibliography}

\section*{Declarations of Competing Interest}
The authors have no competing interests to declare.

\section*{Availability of Code}
The computer code used to run the experiments is available from the corresponding author on reasonable request.

\end{document}